\documentclass[letterpaper, 10 pt, conference]{IEEEtran}  

\IEEEoverridecommandlockouts                              
\IEEEpubid{\begin{minipage}{\columnwidth}\copyright 2022 IEEE.  Personal use of this material is permitted. Permission from IEEE must be obtained for all other uses, in any current or future media, including reprinting/republishing this material for advertising or promotional purposes, creating new collective works, for resale or redistribution to servers or lists, or reuse of any copyrighted component of this work in other works.
\end{minipage}
\hspace{\columnsep}\makebox[\columnwidth]{ }
}

\usepackage[english]{babel}

\makeatletter
\adddialect\l@ENGLISH\l@english
\makeatother

\usepackage{cite}

\usepackage{epsfig}
\usepackage{pst-all, graphics, graphicx, color}
\usepackage[crop=pdfcrop]{pstool}
\usepackage{psfrag}
\usepackage{subfigure}
\usepackage{amsmath, amssymb} 
\usepackage{epstopdf}
\usepackage{tikz}
\usepackage{multirow}
\usepackage{algorithm, algorithmic}
\usepackage{dsfont}
\usepackage{bm}
\usepackage{booktabs} 
\usepackage{float}

\usepackage[nolist]{acronym}

\newacro{dmp}[DMP]{Dynamic Movement Primitive}
\newacro{seds}[SEDS]{Stable Estimator of Dynamical Systems}
\newacro{gmm}[GMM]{Gaussian Mixture Model}
\newacro{gmr}[GMR]{Gaussian Mixture Regression}
\newacro{gpr}[GPR]{Gaussian Process Regression}
\newacro{tpgmm}[TP-GMM]{Task-Parameterized \ac{gmm}}
\newacro{lfd}[LfD]{Learning from Demonstration}
\newacro{il}[IL]{Imitation Learning}
\newacro{wls}[WLS]{weighted least-squares}
\newacro{vs}[VS]{Visual Servoing}
\newacro{ilvs}[ILVS]{Imitation Learning based Visual Servoing}
\newacro{ds}[DS]{Dynamical System}
\newacro{rdsvs}[RDS-VS]{Reshaped \acl{ds} for \acl{vs}}
\newacro{rds}[RDS]{Reshaped \acl{ds}}
\newacro{clfdm}[CLF-DM]{Control Lyapunov Function-based Dynamic Movements}
\newacro{clf}[CLF]{Control Lyapunov Function}
\newacro{fdm}[FDM]{Fast Diffeomorphic Matching}

\def\tr{^{\top}}

\def\bfe{{\mbox{\boldmath $e$}}}
\def\bff{{\mbox{\boldmath $f$}}}

\def\bfr{{\mbox{\boldmath $r$}}}
\def\bfs{{\mbox{\boldmath $s$}}}

\def\bfu{{\mbox{\boldmath $u$}}}
\def\bfv{{\mbox{\boldmath $v$}}}

\def\bfx{{\mbox{\boldmath $x$}}}

\def\bfJ{{\mbox{\boldmath $J$}}}

\def\bfpsi{{\mbox{\boldmath $\psi$}}}

\definecolor{my_green}{rgb}{0.0,0.49,0.19}

\begin{document}

\title{\LARGE \bf%
Learning Stable Dynamical Systems for Visual Servoing}%
\author{Antonio Paolillo$^{1,\ast}$ and Matteo Saveriano$^{2,\ast}$%
\thanks{This work was granted by the European  Union’s  Horizon  2020  Research and Innovation  Programs under Grant No. 871743 (1-SWARM), the Hasler foundation (EViRCo), 
and the Austrian Research Foundation (Euregio IPN 86-N30, OLIVER).
This work was carried out while M. Saveriano was at the Department of Computer Science, University of Innsbruck, Austria.
}%
\thanks{$^{1}$Dalle Molle Institute for Artificial Intelligence (IDSIA), USI-SUPSI, Lugano, Switzerland {\tt antonio.paolillo@idsia.ch}}%
\thanks{$^{2}$Department of Industrial Engineering (DII), University of Trento, Trento, Italy {\tt matteo.saveriano@unitn.it}.}%
\thanks{$^{*}$All authors contributed equally.}
}%
%
%
\maketitle%
\IEEEpubidadjcol
%
%
\begin{abstract}
This work presents the dual benefit of integrating imitation learning techniques, based on the dynamical systems formalism, with the visual servoing paradigm.
On the one hand, dynamical systems allow to program additional skills 
without explicitly coding them in the visual servoing law, but leveraging few demonstrations of the full desired behavior.
On the other, visual servoing allows to consider 
exteroception 
into the dynamical system architecture and be able to adapt to unexpected environment changes. 
The beneficial combination of the two concepts is proven by applying three existing dynamical systems methods to the visual servoing case. 
Simulations 
validate and compare the methods; experiments with a robot manipulator show the validity of the approach in a real-world scenario.
\end{abstract}
%
%
\IEEEpeerreviewmaketitle
%

\section{Introduction}\label{sec:intro}

Easy-to-use control interfaces are today a must-have for the growing number of robotic platforms employed in both domestic and industrial scenarios.
For instance, in manufacturing processes, the work of human operators is facilitated by dictating the motion to the robot by simply touching it~\cite{lee2011incremental,caccavale2019kinesthetic} or through a gesture~\cite{lee2020gesture}.
To allow also inexperienced users to handily utilize robots, \ac{il} has been introduced to reduce, or even completely avoid, the explicit programming of a robotic task.
To this end, the possibility to learn the desired behavior directly from some demonstrations of the task has been widely investigated~\cite{billard16learning}.
Among the several approaches~\cite{billard16learning,Argall2009survey} a particular class of \ac{il} methods proposes to learn the desired robotic task without forgoing the peculiarity of the classical techniques, such as the stability properties of the controllers.
These approaches, known as \ac{ds}-based \ac{il} methods, describe the robot control behavior, or \emph{skill}, as a non-linear dynamics. 
If needed, the \ac{ds} can be modified to achieve more complex tasks without explicitly coding them within the original controller, but leveraging the information embedded into few demonstrations of the full desired behavior.

\ac{ds}-based \ac{il} methods can effectively 
generate robotic kinematic motion
~\cite{DMP,saveriano2018incremental,saveriano2020energy,SEDS,Clf,Perrin16,urain2020imitationflow}.
Approaches in the literature can be grouped in 
three categories.
%
In the first one, 
a \textit{clock signal} 
is used to suppress possibly unstable dynamics. 
\acp{dmp}~\cite{DMP} 
force linear dynamics 
with a non-linear time-dependant term learned from demonstrations.
\ac{rds}~\cite{saveriano2018incremental} superimposes a stable, in general non-linear dynamics, on 
a possibly unstable state-dependant term 
suppressed by a clock signal. 
The work in~\cite{saveriano2020energy} defines the evolution of the clock signal using the energy tank concept~\cite{Franken11}. 
In the second category, Lyapunov stability constraints are enforced via 
optimization to generate converging trajectories. 
\ac{clfdm}~\cite{Clf} learns a non-linear dynamics and a \ac{clf} from the same 
demonstrations. The resulting stable system is 
more accurate than its predecessor~\cite{SEDS}. 
%
Approaches in the third category learn a diffeomorphic mapping from 
a base path (e.g., linear) to 
a complex demonstration. 
%
%
\ac{fdm}~\cite{Perrin16} uses a data and time-efficient approach to learn the mapping, while~\cite{urain2020imitationflow} uses a deep neural network. 

%

These methods rely
on the system trajectories saved in the demonstrations and barely adapt to 
a dynamic environment.
It would be desirable, instead, to provide \acp{ds} with exteroceptive sensing 
to inform the system of possible changes of the surrounding.
For instance, consider a peg-in-hole robotic task. 
\acp{ds} can be used to learn a stable behaviour from few demonstrations. 
However, learning the joint motion is not enough to cope with 
arbitrary changes of the target.
%
Vision might be used to make the robot aware of unexpected external events (e.g., 
the hole position variation) 
and adapt the \ac{ds} accordingly. 
To this end, 
\ac{vs}~\cite{Chaumette:ram:2006,Chaumette:ram:2007} represents a very sensible solution, since its formulation can be interpreted as a \ac{ds}.
%
Thus, \ac{vs} can be used to adapt the existing \ac{ds}-based approaches to the use 
of a visual feedback.

VS is a 
technique to control robots with vision~\cite{Chaumette:ram:2006}. 
Its basic reactive law 
does not allow to easily include constraints or additional tasks. 
%
Therefore, more advanced approaches~\cite{Chaumette:ram:2007} have been proposed, 
from planning (see, e.g.~\cite{Chesi_tro:2007,Mezouar:tro:2002}) to model predictive control~\cite{Sauvee:cdc:2006,Allibert:tro:2010}. 
These approaches allow to add 
tasks and constraints, e.g., 
actuation limits or obstacle avoidance.
%
However, offline planning does not easily adapt to dynamic scenarios, while
in predictive controllers the anticipatory behavior has to be carefully balanced with the computational cost.
%
%
In~\cite{Paolillo:icra:2020} this balance is relaxed 
by leveraging pre-computed solutions.
%
In humanoid robotics, VS has been integrated in 
quadratic program-based whole-body motion controllers 
with other peculiar tasks and constraints of these platforms, see e.g.~\cite{Agravante:ral:2017,Paolillo:ral:2018,Mingo:icra:2021}.
In any case, adding tasks or constraints to the basic VS, 
needs specific design and implementation. 
%
Applying the \ac{ds} rationale to the \ac{vs} scheme would allow to easily code 
complex visual task 
by simply imitating demonstrations of the desired behavior.

\ac{il} has been exploited in several forms 
to learn visual tasks from demonstrations.
%
\ac{gmr} 
models the pixel-camera motion mapping for a peg-in-hole task in~\cite{pignat2019bayesian},
%
or imitates the classic VS behavior from images with reduced resolution~\cite{castelli2017machine}. 
Human demonstrations video is used to learn the visual task~\cite{jin2019robot} or the visual error~\cite{Jin:iros:2020}.
In a human-robot interaction context, vision-based \ac{il} allows to adapt demonstrated trajectories to different surfaces~\cite{dometios2018vision}.
A deep network is trained to imitate an expert for eye surgical operations~\cite{kim:icra:2020}.   
\ac{il} is also used to learn a vision-based policy for fine operations~\cite{Paradis:icra:2021}, and incrementally 
build the training data for a random forest that 
retrieves commands for clothes manipulation~\cite{Jia:ral:2019}.
%
Differently, 
our approach proposes to imitate a visual strategy by explicitly taking into account the controller stability.
These considerations can be deduced if 
the controller analytical structure is maintained and the visual \ac{il} is used to provide, e.g., the feedback~\cite{Johns:icra:2021} or the reference, as done with 
\acp{dmp} in~\cite{phung2012tool}.
%
Similarly, in~\cite{vakanski2017imagebased} the reference of a classical image-based \ac{vs} is inferred from demonstrations 
and then 
refined by an optimizer. 
This approach, differently from ours, uses 
costly optimization to build 
trajectories. 
%

%
%
%

%
%
%

We propose to achieve complex VS tasks resorting to \acp{ds}. 
The 
classic \ac{vs} 
can be interpreted as \ac{ds} (Sec.~\ref{sec:background}) and 
the adaptations needed to apply three existing \acp{ds} 
are shown in Sec.~\ref{sec:approach}. 
Simulations and experiments (Sec.~\ref{sec:experiments}) compare the different approaches.
%
Final remarks 
are drawn in Sec.~\ref{sec:conclusion}. 

\section{Preliminaries}\label{sec:background}

\subsection{Visual servoing}\label{sec:visual_servoing}
\begin{figure}[t]%
    \centering%
    \includegraphics[width=0.95\columnwidth]{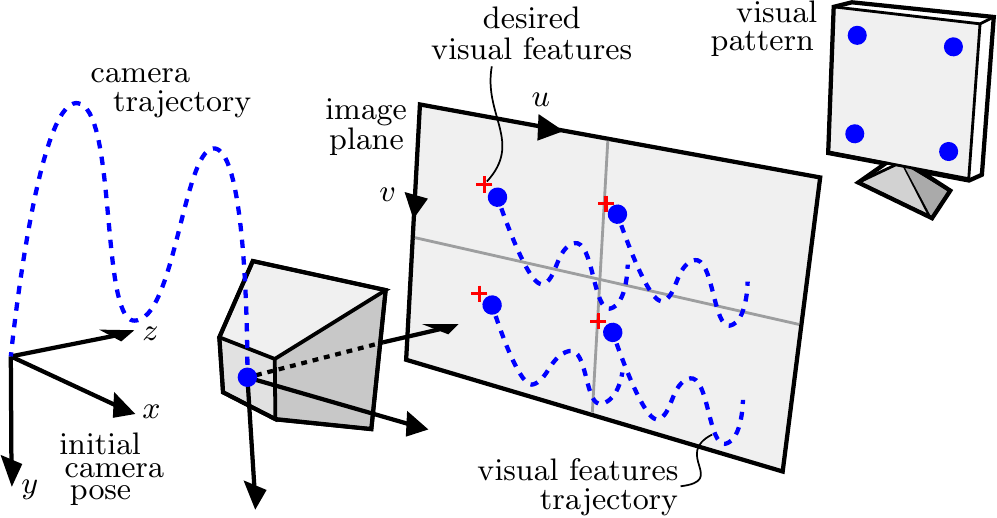}%
    \caption{Schematic representation of a free camera moving in the Cartesian space and observing a visual pattern. VS consists in matching measured visual features with proper desired values, which corresponds to drive the camera to a desired Cartesian pose. Indeed, a visual features trajectory on the image plane corresponds to a camera trajectory in the Cartesian space.}%
    \label{fig:camera_and_traj}%
\end{figure}

%
VS allows to control robots using a visual feedback with
the form of features $\bm{s} \in \mathbb{R}^f$.
The aim is the matching with desired values $\bm{s}^\ast$ to zero the error $\bm{e} = \bm{s} - \bm{s}^\ast$. 
%
In image-based VS, considered in this work, the feedback is directly defined on the image. 
%
With constant $\bm{s}^\ast$ and a fixed target, 
%
%
%
the classical VS computes the 6D camera velocity commands as 
%
\begin{equation}
    \bm{v} = - \lambda \hat{\bm{L}}^{+} \bm{e}
    \label{eq:vs_control}
\end{equation}
where $\lambda$ is a positive control gain and 
$\hat{\bm{L}}$ is an approximation (e.g., 
due to the unknown features depth) 
of the interaction matrix~\cite{Chaumette:ram:2007}. 
A possibility is to use its value at the target, known in advance~\cite{Chaumette:ram:2006}.
%
%
To fully control the camera in the 3D space, avoid 
singularities 
and poses ambiguity, the feedback dimension $f$ has to square the system, at least~\cite{Chaumette:ram:2006}. 
In image-based VS, the stability 
is ensured only close to 
the target, 
if the interaction matrix is full rank and well approximated. 
%
%
In this case, 
\eqref{eq:vs_control} realizes converging trajectories of the features to their desired values. 
For example, using the true interaction matrix and point features, 
such a convergence generates 
a linear motion of $\bfs$ towards $\bfs^\ast$. 
For 
complex trajectories, like that in Fig.~\ref{fig:camera_and_traj}, 
the features should deviate from this 
expected motion, e.g., by adding 
constraints or tasks to the VS law.
%
%
%
\subsection{Skills representation using dynamical systems}
%
%
%
%
%
%
In \ac{ds}-based \ac{il}, it is common to assume that a robotic skill is encoded into a control affine system:
\begin{equation}
	\dot{\bfx} = \bff(\bfx) + \bfu(\bfx,t),
	\label{eq:ds_syst}
\end{equation}
where $\bfx \in \mathbb{R}^{n}$ is the state (e.g., robot position), $\dot{\bm{x}}\in \mathbb{R}^{n}$ is the time derivative of the state, $t \in \mathbb{R}^{+}$ is the time, and $\bfu(\cdot)$ a control input. Both $\bff(\cdot)$ and $\bfu(\cdot)$ are assumed to be smooth.
In order to use \acp{ds} to generate discrete and 
converging-to-a-target trajectories, it is important to guarantee the stability of~\eqref{eq:ds_syst}, 
e.g., using Lyapunov arguments~\cite{Slotine91}. 
The work in~\cite{DMP, saveriano2018incremental} assume that $\bm{f}$ 
is stable and exploits the explicit time dependency of $\bfu$ to ensure $\bfu\rightarrow\bm{0}$ for $t\rightarrow+\infty$, guaranteeing the convergence to a given target. 
Other approaches~\cite{SEDS,Clf,tau-SEDS} 
further restrict the \ac{ds} in~\eqref{eq:ds_syst} to be autonomous 
(without explicit time dependencies). 
More in detail,~\cite{SEDS,tau-SEDS} assume 
$\bfu=\bm{0}$ and focus on learning a stable $\bff$; 
\cite{Clf} computes a stabilizing control input $\bfu$ from a learned Lyapunov function. 

\subsection{Problem definition}

We propose to modify the standard VS law~\eqref{eq:vs_control} using the DS rationale~\eqref{eq:ds_syst} 
to augment the skills of the vision-based robotic behavior.
In this way, additional tasks 
do not need to be explicitly coded. 
Instead, they can be programmed from 
demonstrations of the full desired behavior, which are manually executed and stored offline.
%
The objective is to online infer from data the modification to be applied to the basic original behavior without damaging its 
stability. 

Image-based VS uses a 
feedback directly defined on the image to 
move the camera to a desired pose. 
Indeed, a 3D camera trajectory 
corresponds a visual features trajectory on the 
image plane, see Fig.~\ref{fig:camera_and_traj}.
In practice, the VS skill cannot be directly expressed as in~\eqref{eq:ds_syst}, where the variable of interest $\bm{x}$ and its time derivative are defined in the same state space.

In fact, in the case of robot kinematic control, 
$\bm{x}$ is often associated to configuration vectors or end-effector poses~\cite{SEDS,Clf,saveriano2018incremental,Perrin16} 
%
and actually treated as an error\footnote{Without loss of generality, it is commonly assumed that 
the system has to converge to zero, 
i.e., the target is assumed constant and 
at the origin.
} (in the joint or task space). 
Thus, DS is used to generate a stable behavior zeroing this error.
Joint and task space quantities are also taken into account in the same DS~\cite{shavit2018learning} by first projecting the task space error in the  joint space with the robot Jacobian transpose. 
The projected error is then pre-multiplied by a learned matrix 
forcing the trajectory to follow the demonstrations.
Also in the VS case, DS has 
to move between two spaces
: the image and the Cartesian world.
%
Namely, the DS-based VS goal 
is to describe the desired camera behavior, 
through its velocity $\bm{v}$, 
from visual measurements $\bm{e}$.
%
Thus, it is convenient to define 
\begin{equation}
\bm{\varepsilon} \triangleq \hat{\bm{L}}^{+}\bfe
\label{eq:int_err}
\end{equation}
%
relating the visual error to a Cartesian quantity and allowing to write the DS-based VS as $\bm{v}=\bm{f}(\bm{\varepsilon})+\bm{u}(\bm{\varepsilon},t)$ or $\bm{v}=\bm{f}(\bm{\varepsilon})$ depending on the DS choice. 
%
%
Note that the 
DS-based formulation differs from an end-to-end approach in that the control keeps its 
structure.
%
DS methods learns $\bm{f}$ 
(and/or $\bm{u}$, according with the chosen method) 
respecting
the original controller stability.
%
%
We 
further highlight that 
%
$\hat{\bm{L}}$, 
expressing the image-Cartesian space mapping, is not learned but its analytical structure is 
maintained in our data, see Sec.~\ref{sec:learning}. 
In principle, $\bm{\varepsilon}$ could be measured from images---like in position-based \ac{vs} schemes---and classical DSs could be directly used. However, pose estimation requires a major perception effort; we prefer to use image-based \ac{vs}, for which an adaptation of existing \acp{ds} is required (see Sec.~\ref{sec:approach}).


Solving complex VS tasks is formulated as 
%
learning DSs such that they (i) 
stay 
stable as the original VS; 
(ii) reproduce the demonstrations; 
(iii) 
online adapt to environment changes using visual feedback.
To this end, we apply three existing DSs 
to VS, 
analysing and comparing their performance.

\section{DS-Based Methods for VS}\label{sec:approach}

In this section, we describe more in details the \ac{rds}~\cite{saveriano2018incremental}, \ac{clfdm}~\cite{Clf}, and \ac{fdm}~\cite{Perrin16} methods, and how to exploit and adapt these approaches to learn complex \ac{vs} tasks.

\subsection{Demonstrations dataset}\label{sec:learning}



Let us assume to have $D$ demonstrations of $N$ samples:
\begin{equation}
    {\cal D} = \left\{ \bm{e}_n^d, \, \bm{v}_{n}^d \right\}_{n=1,d=1}^{N,D}
\label{eq:demonstrations}
\end{equation}
where $\bm{v}_{n}^d$ is the camera velocity needed 
to achieve the desired VS task at the $n$-th sample of the $d$-th demonstration; 
$\bm{e}_n^d$ is the related 
visual error.
%
We make the following assumptions:
\begin{enumerate}
    \item the desired visual features $\bm{s}^\ast$ are fixed and known in advance; thus for a given set of observed visual features $\bm{s}$, the error $\bm{e}$ is directly and uniquely defined\label{ass:1};
    \item a sufficient number of visual features, enough to uniquely identify the pose of the camera, is used\label{ass:2};
    \item $\lambda$ is known and remains the same during the demonstrations collection and the methods validation\label{ass:3};
    \item the interaction matrix is approximated with its known value at the target $\forall \, n, d$, i.e., $\hat{\bm{L}}_n^d = \bm{L}_N^D$\label{ass:4}. 
\end{enumerate}
Under assumptions~\ref{ass:1}-\ref{ass:2}, the camera velocity 
can be uniquely inferred from ${\cal D}$ 
only using $\bm{s}$.
Assumptions~\ref{ass:3}-\ref{ass:4} 
allow to compute the variable $\bm{\varepsilon}$
~\eqref{eq:int_err} and the standard VS law~\eqref{eq:vs_control} $\forall \, n, d$.
%
%
The demonstrations~\eqref{eq:demonstrations} are 
the base for building the training datasets 
for the considered \acp{ds} as detailed in what follows.
%

\subsection{\ac{rds} for \ac{vs}}\label{sec:rds}

The RDS idea~\cite{saveriano2018incremental} is to reshape a stable base dynamics with an additive term following 
few demonstrations. 
Convergence 
is ensured 
forcing the reshaping 
to vanish after a certain time. 

Using the the rationale behind \ac{rds}, we propose to handle complex \ac{vs} tasks reshaping the base dynamics~\eqref{eq:vs_control} as follows
%
\begin{equation}
\bm{v} = -\lambda \bm{\varepsilon} + h \bfu_\text{RDS}(\bm{\varepsilon}),
\label{eq:reshaped_control}%
\end{equation}
%
where $\bfu_\text{RDS}(\bm{\varepsilon}) \in \mathbb{R}^{n}$ is a continuous and continuously differentiable function, acting as reshaping term of the original \ac{vs} law.
%
%
It is used to enable, besides classical \ac{vs} tasks, additional skills that are difficult to carry out with~\eqref{eq:vs_control}, leveraging the information stored in demonstrations of the full desired task.
The activation of the reshaping term is triggered by the 
clock variable $h$ that vanishes after a user-defined time~\cite{saveriano2018incremental}.

Our goal is to estimate $\bm{u}_\text{RDS}$ 
such that the camera velocity computed from~\eqref{eq:reshaped_control} accurately reproduces the demonstrations~\eqref{eq:demonstrations}.
Thus, we first need to reconstruct $\bm{u}_{\text{RDS},n}^d$ from $\cal D$. Assuming $h\!=\!1$ in~\eqref{eq:reshaped_control}, it is 
$\bfu_\text{RDS}(\bm{\varepsilon}) = \bm{v} + \lambda \bm{\varepsilon}$. 
The dataset $\cal D$ 
can thus be translated in the actual training dataset
\begin{equation}
    {\cal T}_\text{RDS} = \left\{ \bm{\varepsilon}_n^d, \, \bm{u}_{\text{RDS},n}^d \right\}_{n=1,d=1}^{N,D}.
    \label{eq:training_dataset}
\end{equation}
%
Setting $h=1~\forall~n, d$ implies only minor approximation errors on the overall reproduction accuracy, as remarked in~\cite{saveriano2018incremental} and empirically observed also in this work, see Sec.~\ref{sec:experiments}.
%
%
%
%
%
%
%
%

Given ${\cal T}_\text{RDS}$ and 
the assumptions in Sec.~\ref{sec:learning}, a reshaping term can be uniquely computed from a set of observed visual features.
%
Thus, any regression technique can learn and online retrieve a reshaping term from the training dataset for a novel set of visual feature, i.e. we can estimate the reshaping term as $\hat{\bm{u}}_\text{RDS}(\bm{\varepsilon}) \simeq \bm{r}(\bm{s} | {\cal T}_\text{RDS})$, being $\bm{r}$ a regression function. 
%
%
%
Among the others, we use \ac{gmr}~\cite{GMR} because: its estimates are smooth and respect the assumption on $\bm{u}_\text{RDS}$;
it is fast to compute, thus, suitable to online control loops;
%
%
the only hyper-parameter 
(the number of Gaussians) is easy to tune. 

\subsection{\ac{clfdm} for \ac{vs}}\label{sec:clfdm}
\ac{clfdm} stabilizes at run-time possibly unstable dynamics learned from demonstration~\cite{Clf}. 
The stabilization is achieved with a control input 
computed from a \ac{clf} solving a convex optimization problem. 
%
Also CLF, denoted with $V(\bm{\varepsilon})$, 
is learned from the demonstrations with the objective of minimizing the number of points for which $\dot{V}(\bm{\varepsilon}) \geq 0$.   

With \ac{clfdm}, complex \ac{vs} tasks are handled by the law
\begin{equation}\label{eq:clf_control}
 \bm{v} = \bm{f}(\bm{\varepsilon}) + \bm{u}_\text{CLF}(\bm{\varepsilon}),
\end{equation}
where $\bm{f}(\bm{\varepsilon})$ is a possibly unstable dynamics and $\bm{u}_\text{CLF}(\bm{\varepsilon})$ is a stabilizing controller computed from  $V(\bm{\varepsilon})$. 

In \ac{clfdm}, the VS skill is encoded in $\bm{f}(\bm{\varepsilon})$, while $\bm{u}_\text{CLF}(\bm{\varepsilon})$ is only responsible to correct the dynamics of $\bm{f}(\bm{\varepsilon})$ to preserve stability. Therefore, the learning in \ac{clfdm} is performed in two steps. First, the camera velocity $\bm{f}(\bm{\varepsilon})$ is estimated 
such that 
it accurately reproduces the demonstrations.
Thus, 
we compute $\bm{f}_n^d$ from 
$\cal D$ as $\bm{f}_n^d = \bm{v}_{n}^d$, 
and the demonstrations 
are translated into the actual training dataset 
\begin{equation}
    {\cal T}_\text{CLF} = \left\{ \bm{\varepsilon}_n^d, \, \bm{f}_{n}^d \right\}_{n=1,d=1}^{N,D}.
    \label{eq:training_dataset_clf}
\end{equation}
As for \ac{rds}, any regression technique can learn and online retrieve $\bm{f}(\bm{\varepsilon})$ from 
${\cal T}_\text{CLF}$ for a novel set of visual feature, i.e. 
$\hat{\bm{f}}(\bm{\varepsilon}) \simeq \bfr(\bfs | {\cal T}_\text{CLF})$. 
Also here, 
$\bfr$ is implemented as a \ac{gmr}. 

In the second learning step, an estimate $\hat{V}$ of the CLF is learned solving an optimization problem minimizing the number of points for which $\dot{\hat{V}} = \nabla \hat{V}(\bm{\varepsilon}_n^d) \bm{f}_{n}^d \geq 0$. The expression of $\dot{\hat{V}}$ depends on $\bm{\varepsilon}_n^d$ and $\bm{f}_{n}^d$, which are already contained into the training dataset ${\cal T}_\text{CLF}$ in~\eqref{eq:training_dataset_clf}. 
After the learning, we can retrieve $\hat{V}$ and $\nabla \hat{V}$ for the observed $\bm{\varepsilon}$ and compute the stabilizing control input $\bm{u}_\text{CLF}(\bm{\varepsilon})$ at run time as
 \begin{equation}
 \bfu_\text{CLF}(\bm{\varepsilon}) = 
 \left\{
 \begin{array}{cl}
 \bm{0} & \forall~\bm{\varepsilon} \neq \bm{0}: \dot{\hat{V}} \leq -\rho \\
 -\hat{\bm{f}} & \text{if}~\bm{\varepsilon} = \bm{0} \\
 \frac{\nabla \hat{V} \hat{\bm{f}} + \rho}{\nabla \hat{V} \nabla \hat{V}\tr}  \nabla \hat{V}\tr & \text{otherwise}
 \end{array}
 \right.
 \label{eq:u_clf}
\end{equation}
%
being $\rho(\Vert \bm{\varepsilon} \Vert)\!>\!0$ a class $\mathcal{K}$ function imposing 
the minimum 
change rate of $\hat{V}$; 
the dependence on $\bm{\varepsilon}$ is 
omitted for 
brevity.
The details about the derivation of~\eqref{eq:u_clf} can be found in~\cite{Clf}.

\subsection{\ac{fdm} for \ac{vs}}\label{sec:fdm}

\ac{fdm} is used to compute a diffeomorphism, a continuous and differentiable function $\bfpsi(\cdot)$ with continuous and differentiable inverse, mapping linear trajectories into the demonstrated motions. Convergence to a target is guaranteed by the diffeomorphism properties (see~\cite{Perrin16} for the details).

According to \ac{fdm}, complex \ac{vs} tasks can be handled as:
\begin{equation}\label{eq:fdm_control}
 \bfv = -\gamma(\bm{\varepsilon})\bfJ_{\psi}^{-1}(\bm{\varepsilon}) \bm{\varepsilon},
\end{equation}
%
being $\bfJ_{\psi}$ the Jacobian of the diffeomorphism $\bfpsi(\bm{\varepsilon})$; $\gamma(\bm{\varepsilon})$ is a variable gain used to track the demonstrated velocity.
Both $\bfpsi(\bm{\varepsilon})$ and $\gamma(\bm{\varepsilon})$ are estimated from 
data as described next. 

As already mentioned, $\bfpsi$ maps linear error trajectories $\bm{\varepsilon}_n^l$ into the demonstrated errors $\bm{\varepsilon}_n^d$, i.e., $\bm{\varepsilon}_n^d = \bfpsi(\bm{\varepsilon}_n^l)$. The linear trajectory is obtained as $\{\bm{\varepsilon}_n^l = \left[(N-n)/N\right]\tilde{\bm{\varepsilon}}_1\}_{n=0}^N$, where $\tilde{\bm{\varepsilon}}_1$ is the average of the $D$ initial errors computed from~\eqref{eq:demonstrations}. 
It is easy to verify that $\bm{\varepsilon}_n^l$ contains a set of straight lines connecting each component of $\tilde{\bm{\varepsilon}}_1$ to the origin. Note that the \ac{fdm} training algorithm does not handle multiple demonstrations. 
A solution~\cite{Perrin16} is to average the $D$ demonstrations---assuming they have the same length---obtaining the average errors $\tilde{\bm{\varepsilon}}_n$. This solution is simple, but it generates deviations between reproduced and demonstrated motions (see Sec.~\ref{sec:experiments}).
With these considerations, we translate $\cal D$ into 
\begin{equation}
    {\cal T}_\text{FDM} = \left\{ \tilde{\bm{\varepsilon}}_n, \, \bm{\varepsilon}_{n}^l \right\}_{n=1}^{N}
    \label{eq:training_dataset_fdm}
\end{equation}
and use it to estimate $\bfpsi(\bm{\varepsilon}) \simeq \bfr(\bfs | {\cal T}_\text{FDM})$ using the fast algorithm proposed in~\cite{Perrin16}.
The gain variable $\gamma(\bm{\varepsilon})$ is defined such that the error trajectory, governed by $\dot{\bm{\varepsilon}} = -\gamma(\bm{\varepsilon})\bm{\varepsilon}$, starts from $\tilde{\bm{\varepsilon}}_1$,  passes through the points $\tilde{\bm{\varepsilon}}_n$ and finally converges to zero. Following the formulation in~\cite{Perrin16}, it is:%
%
\begin{equation}\label{eq:gamma_fdm}
\gamma(\bm{\varepsilon}) = 
\begin{cases}
\Vert \tilde{\bm{\varepsilon}}_1 \Vert / (N T \Vert \bm{\varepsilon} \Vert ) &  \text{if}~\Vert \bm{\varepsilon} \Vert \geq \Vert \tilde{\bm{\varepsilon}}_1 \Vert / N \\
\Vert \tilde{\bm{\varepsilon}}_1 \Vert / N & \text{otherwise}
\end{cases}
\end{equation}
where $T$ is the sampling time.

\section{Results}\label{sec:experiments}

\begin{figure}[!t]
    \newcommand{\myscale}{0.25}%
    \centering%
    \includegraphics[trim={0 52 0 0},clip,scale=\myscale]{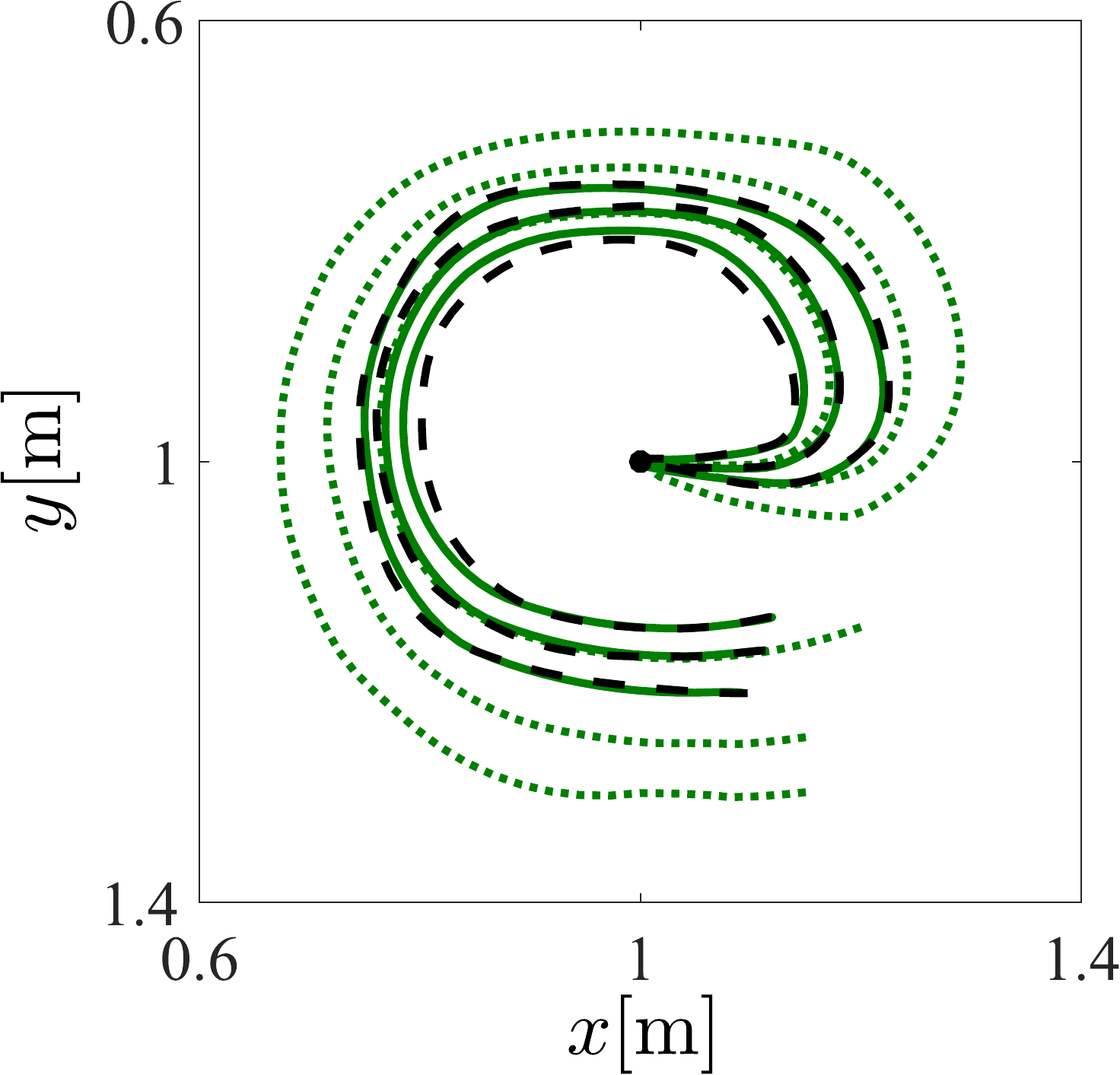}%
    \qquad%
    \includegraphics[trim={0 52 0 0},clip,scale=\myscale]{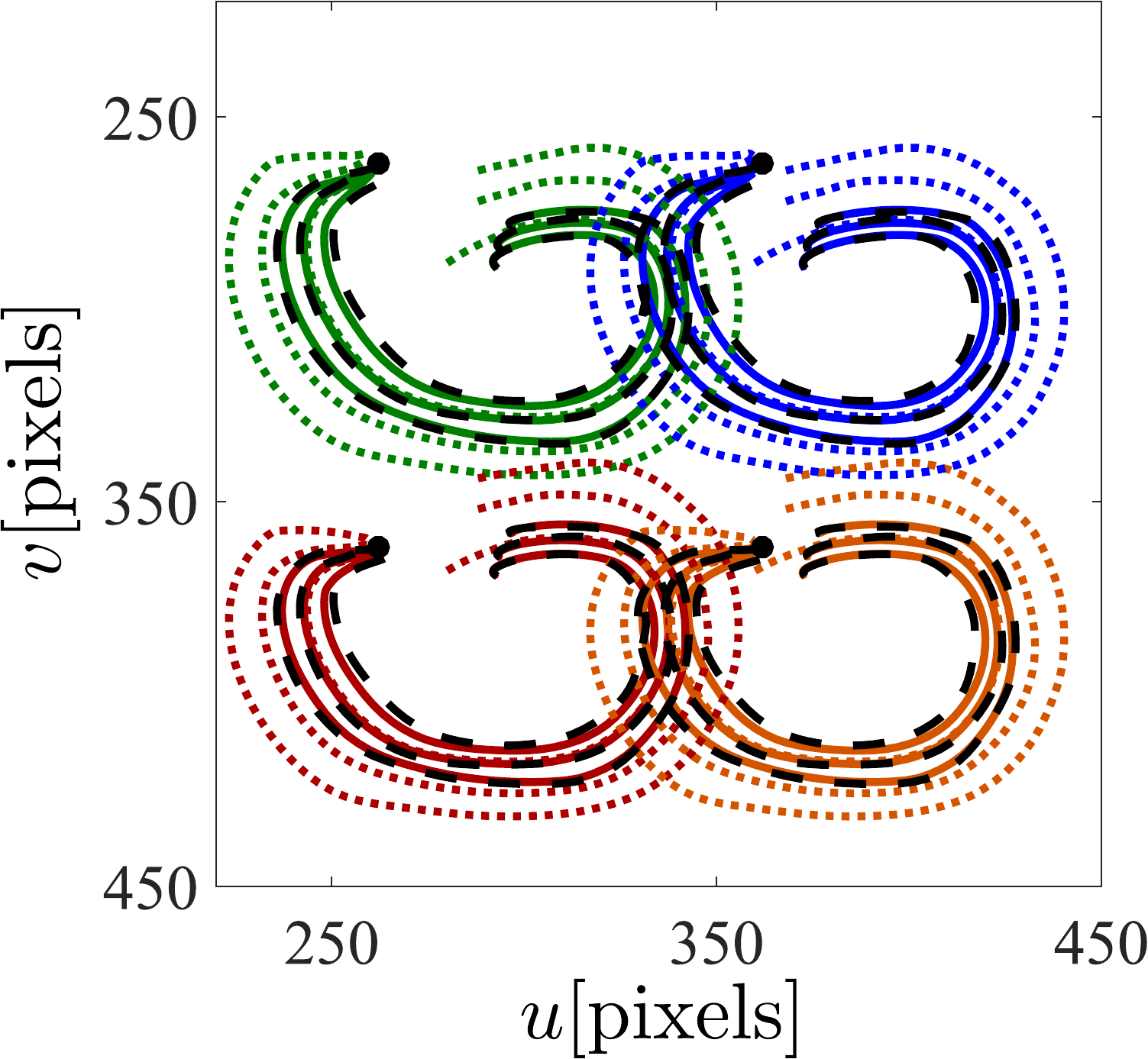}%
    \\[1pt]
    \includegraphics[trim={0 52 0 0},clip,scale=\myscale]{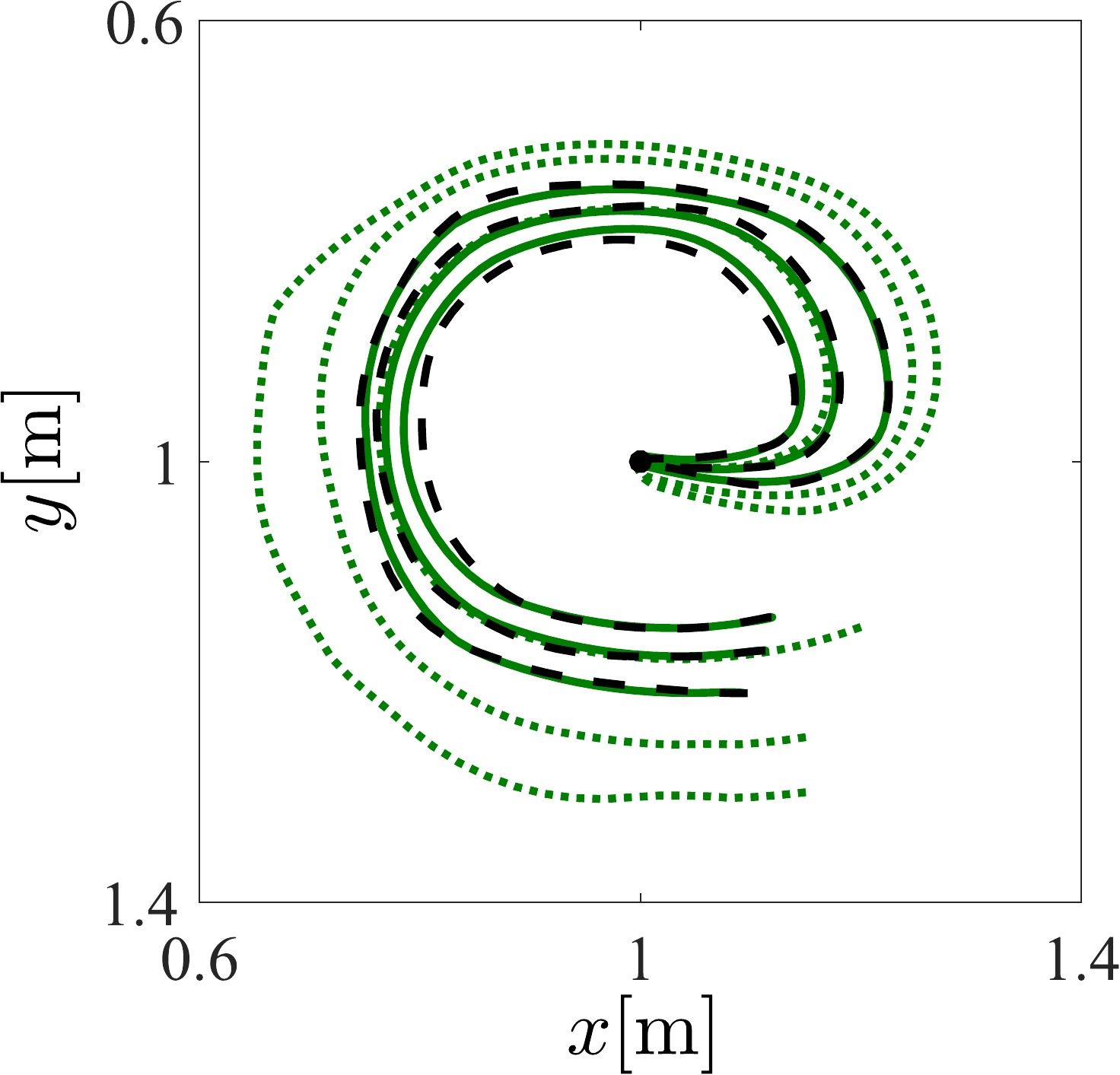}%
    \qquad
    \includegraphics[trim={0 52 0 0},clip,scale=\myscale]{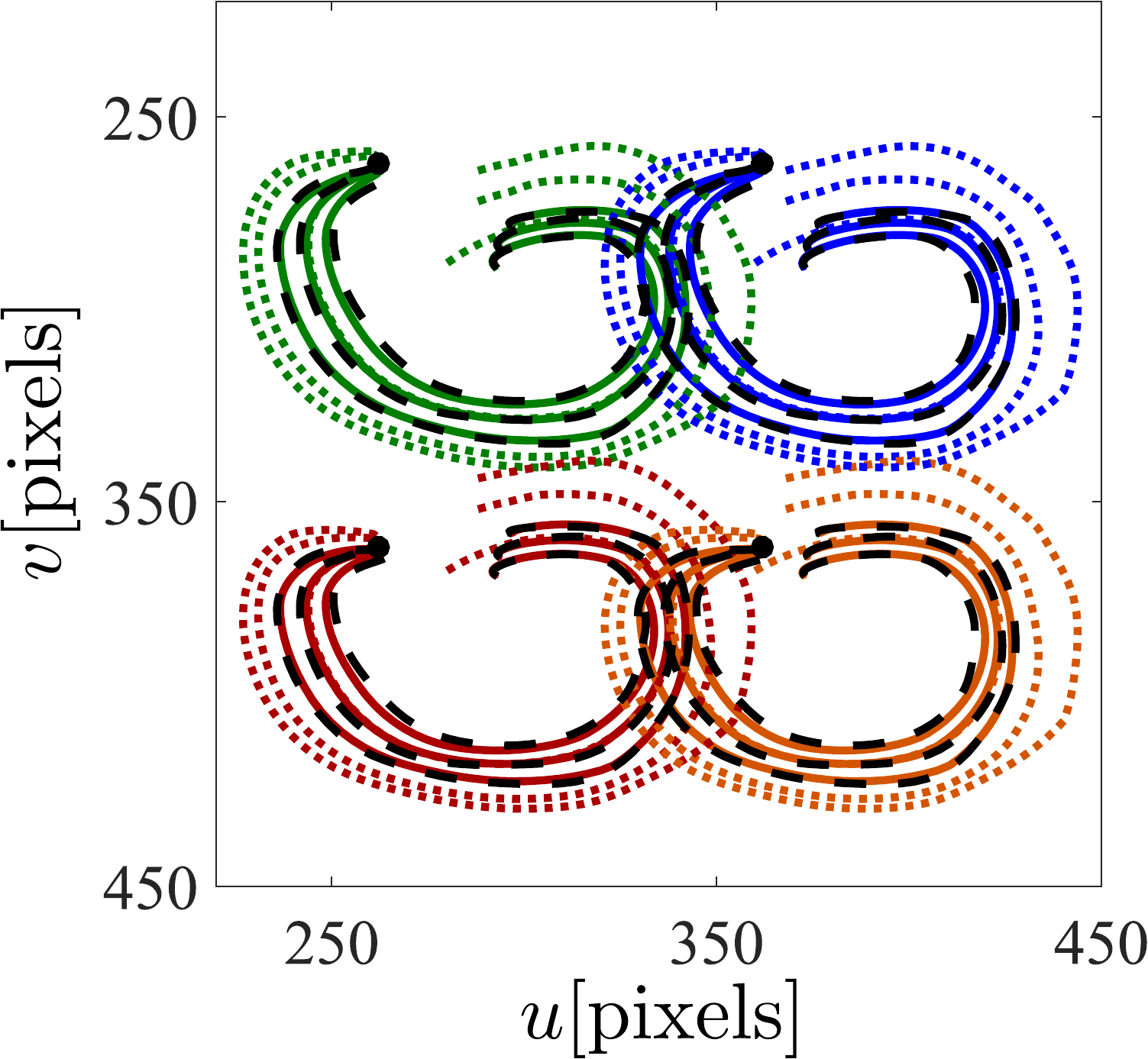}%
    \\[1pt]
    \includegraphics[scale=\myscale]{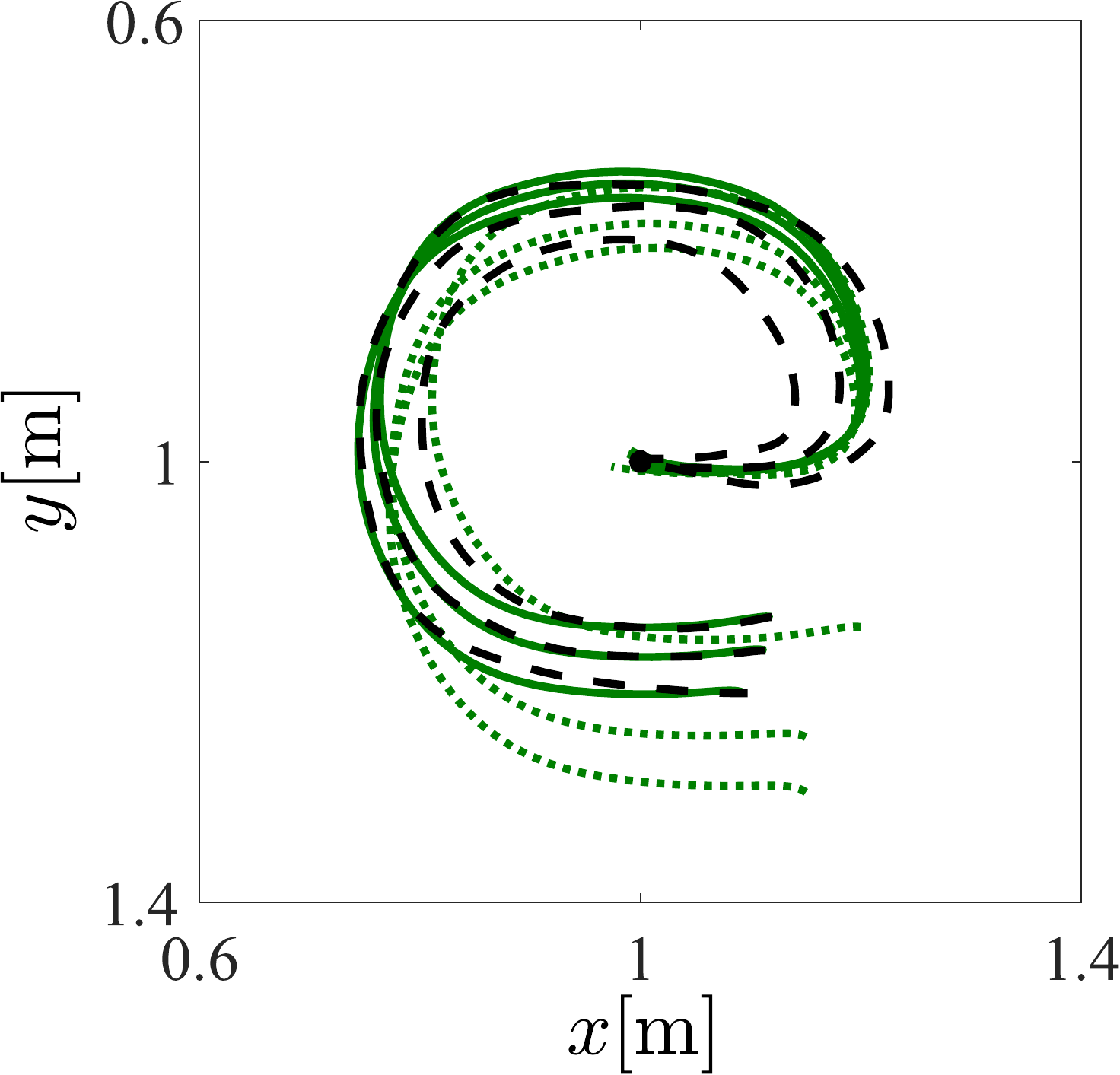}%
    \qquad
    \includegraphics[scale=\myscale]{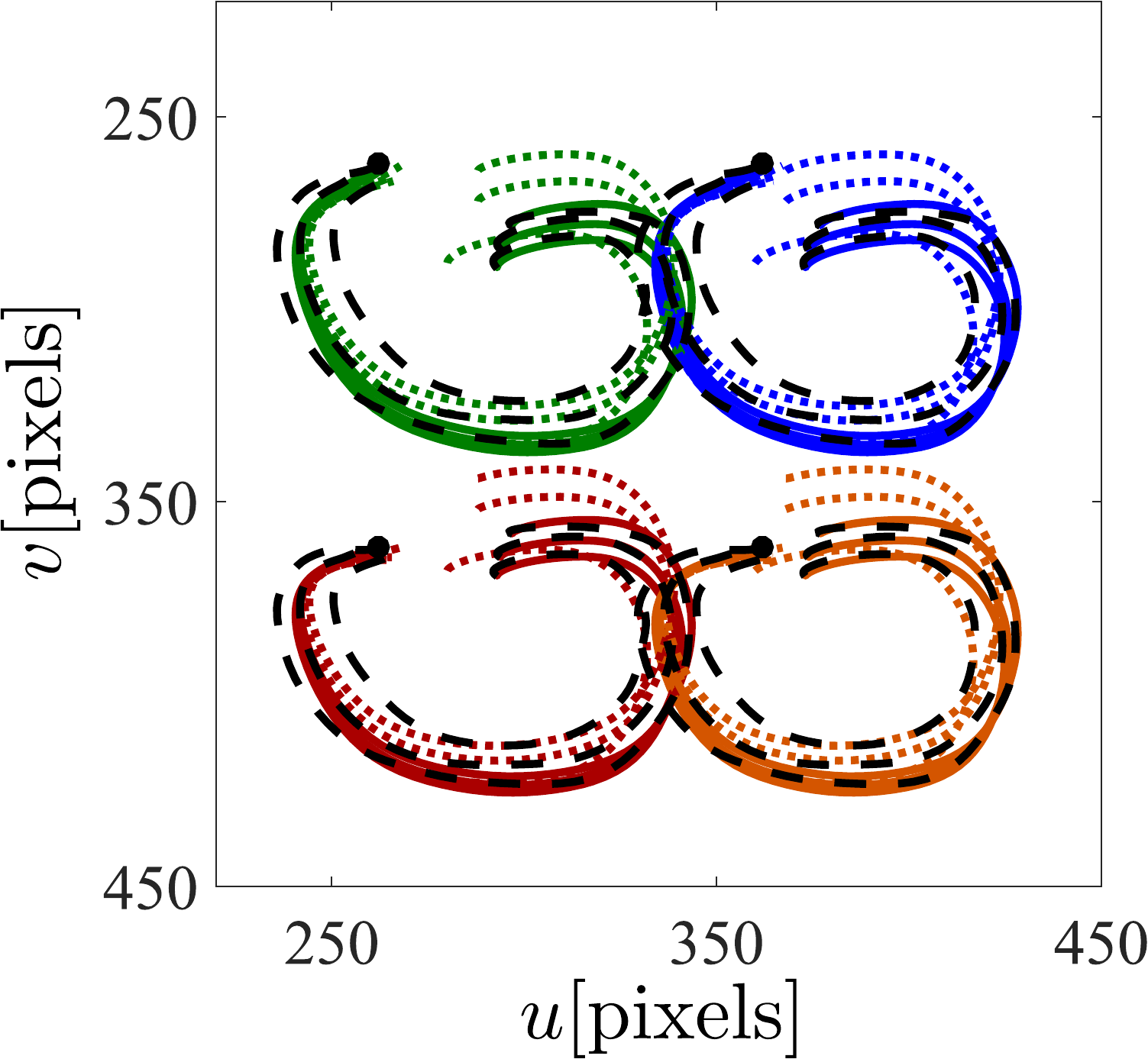}%
    %
    \caption{One motion class of the augmented LASA handwriting dataset and its 
    RDS (top), CLF-DM (center) and FDM (bottom) reproduction in Cartesian space ($x$-$y$ view 
    shown in the left panels) and image plane (right). 
    }%
    \label{fig:dataset_one}
\end{figure}

\begin{figure*}[!th]
    \centering 
    \newcommand{\motionClassSize}{0.078}%
    \hspace{5mm}
    \includegraphics[width=\motionClassSize\textwidth]{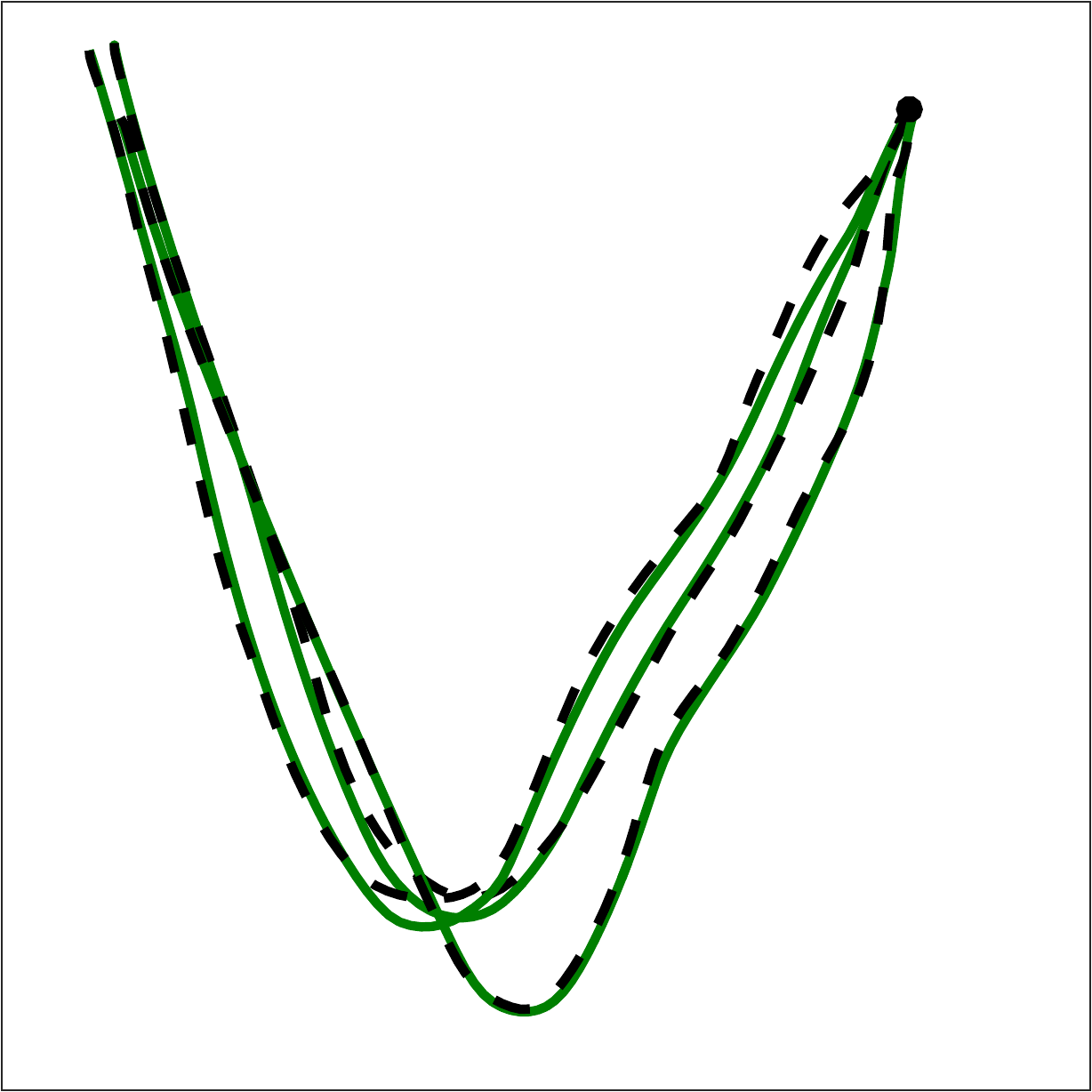}
    \hspace{-0.5mm}
    \includegraphics[width=\motionClassSize\textwidth]{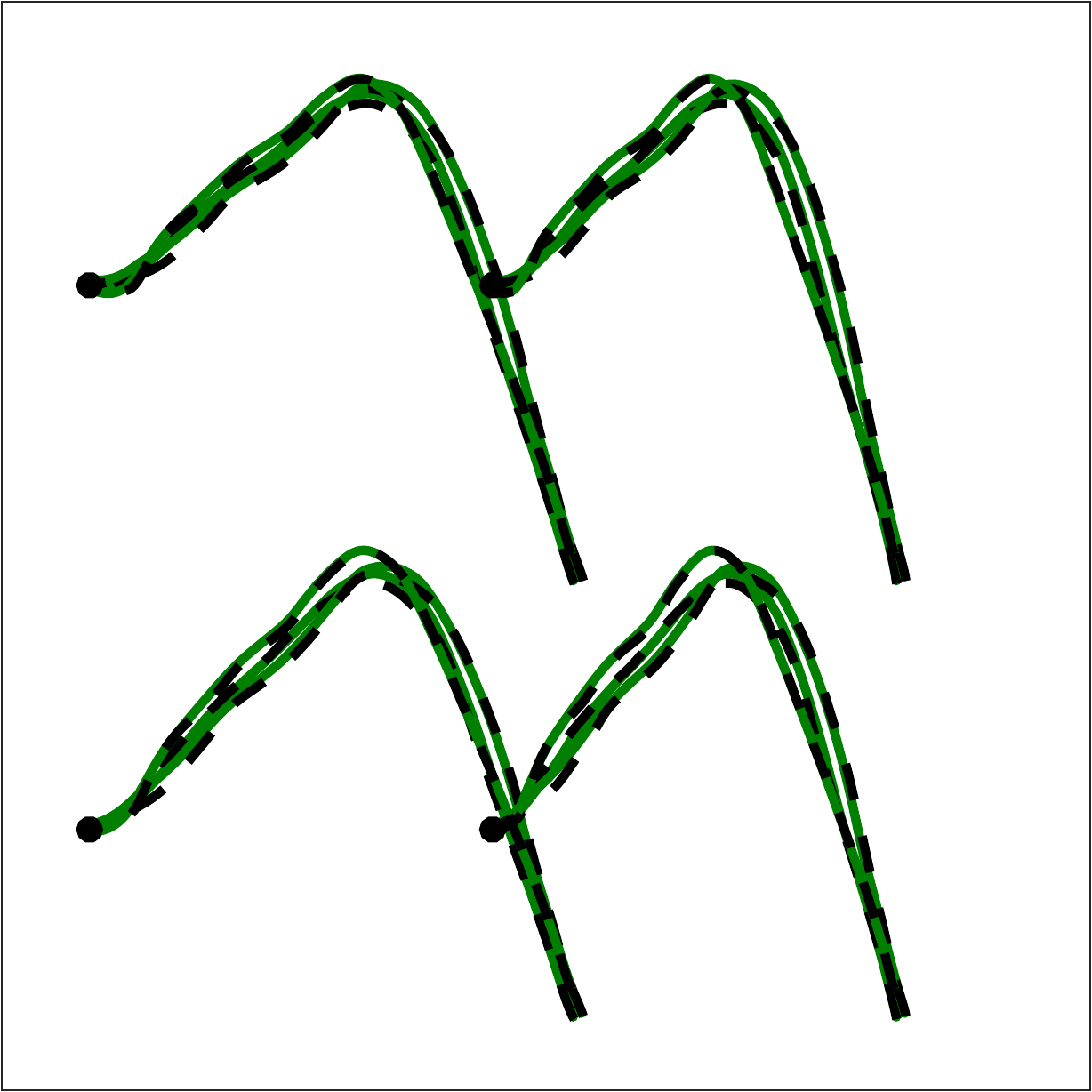}
    \hfill
    \includegraphics[width=\motionClassSize\textwidth]{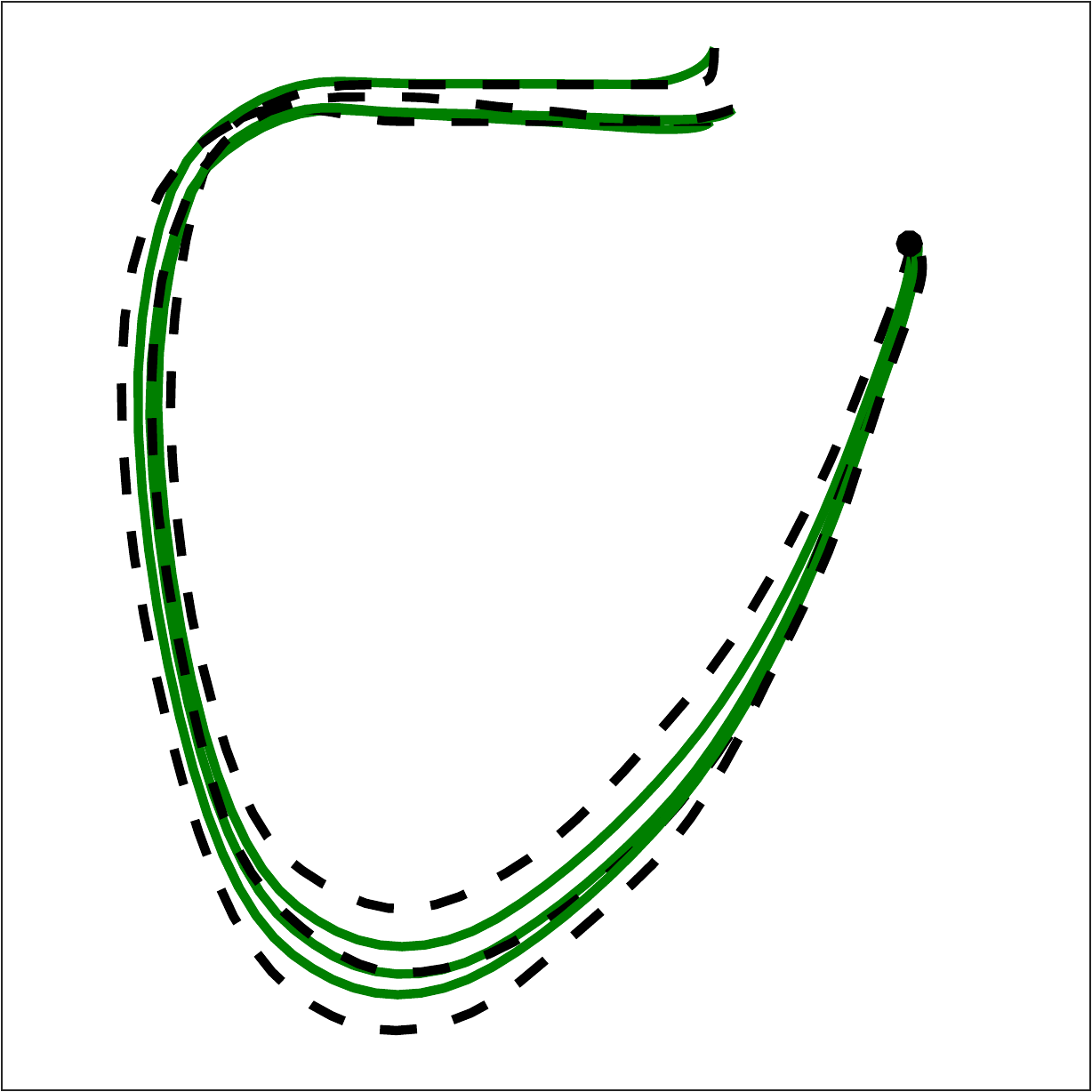}
    \hspace{-0.5mm}
    \includegraphics[width=\motionClassSize\textwidth]{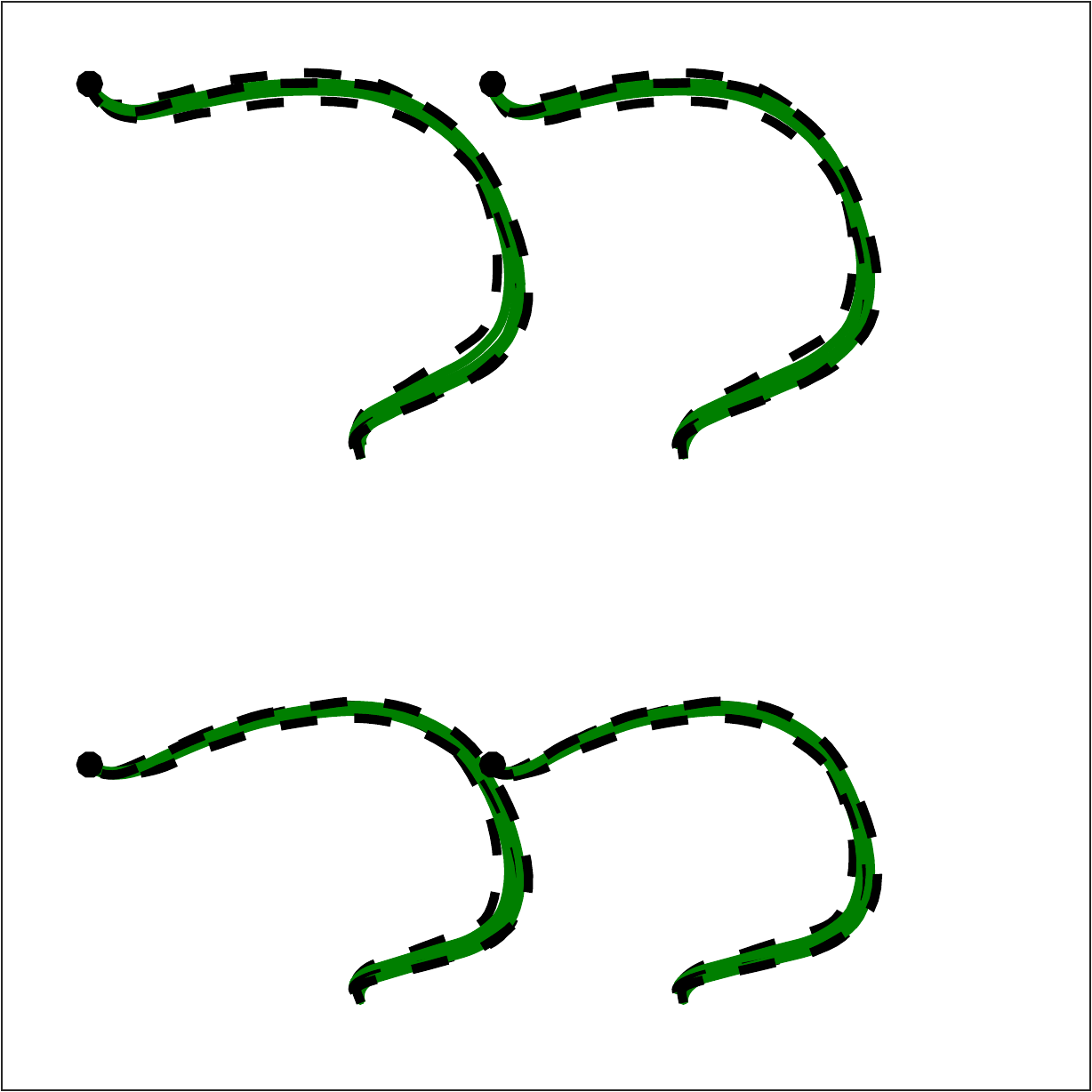}
    \hfill
    \includegraphics[width=\motionClassSize\textwidth]{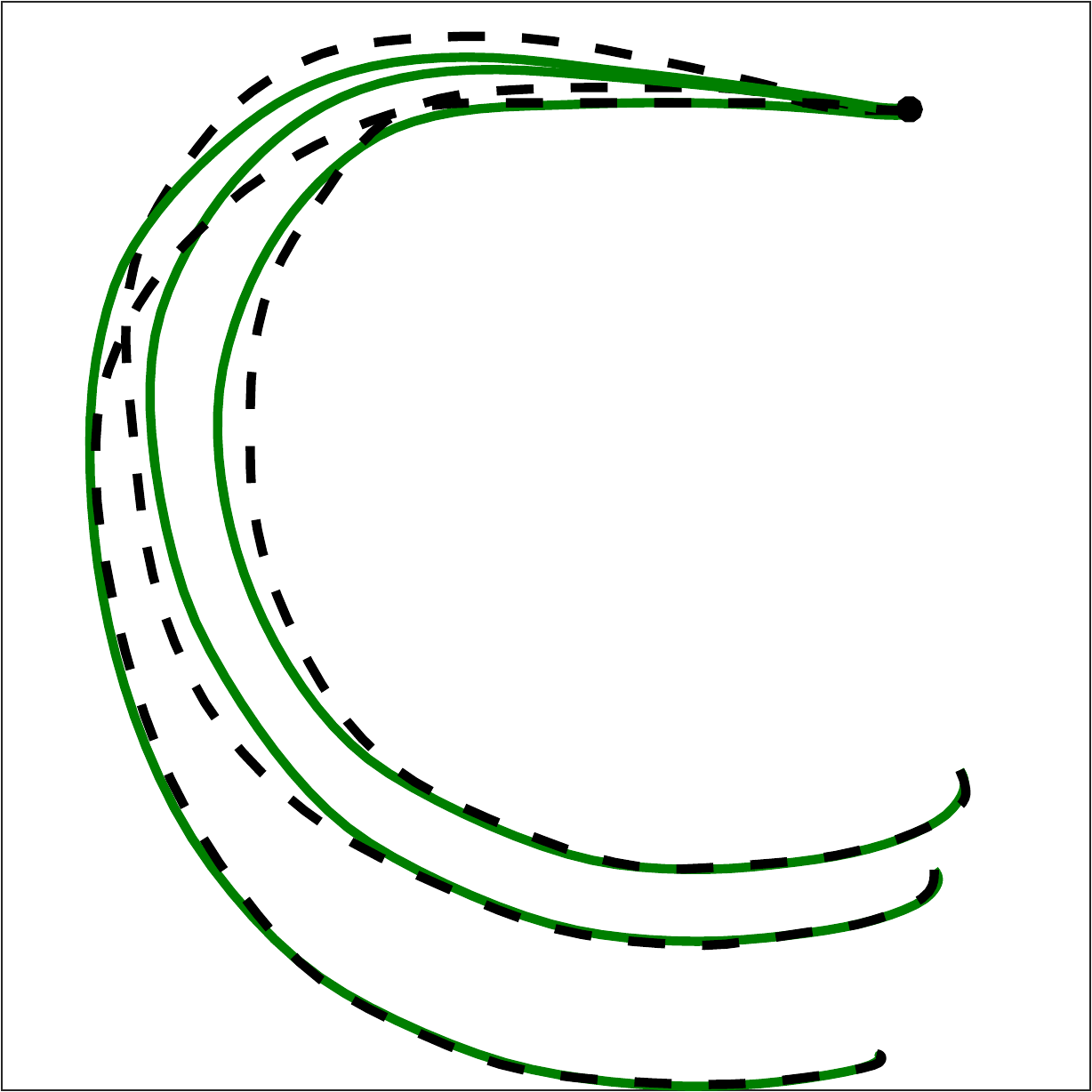}
    \hspace{-0.5mm}
    \includegraphics[width=\motionClassSize\textwidth]{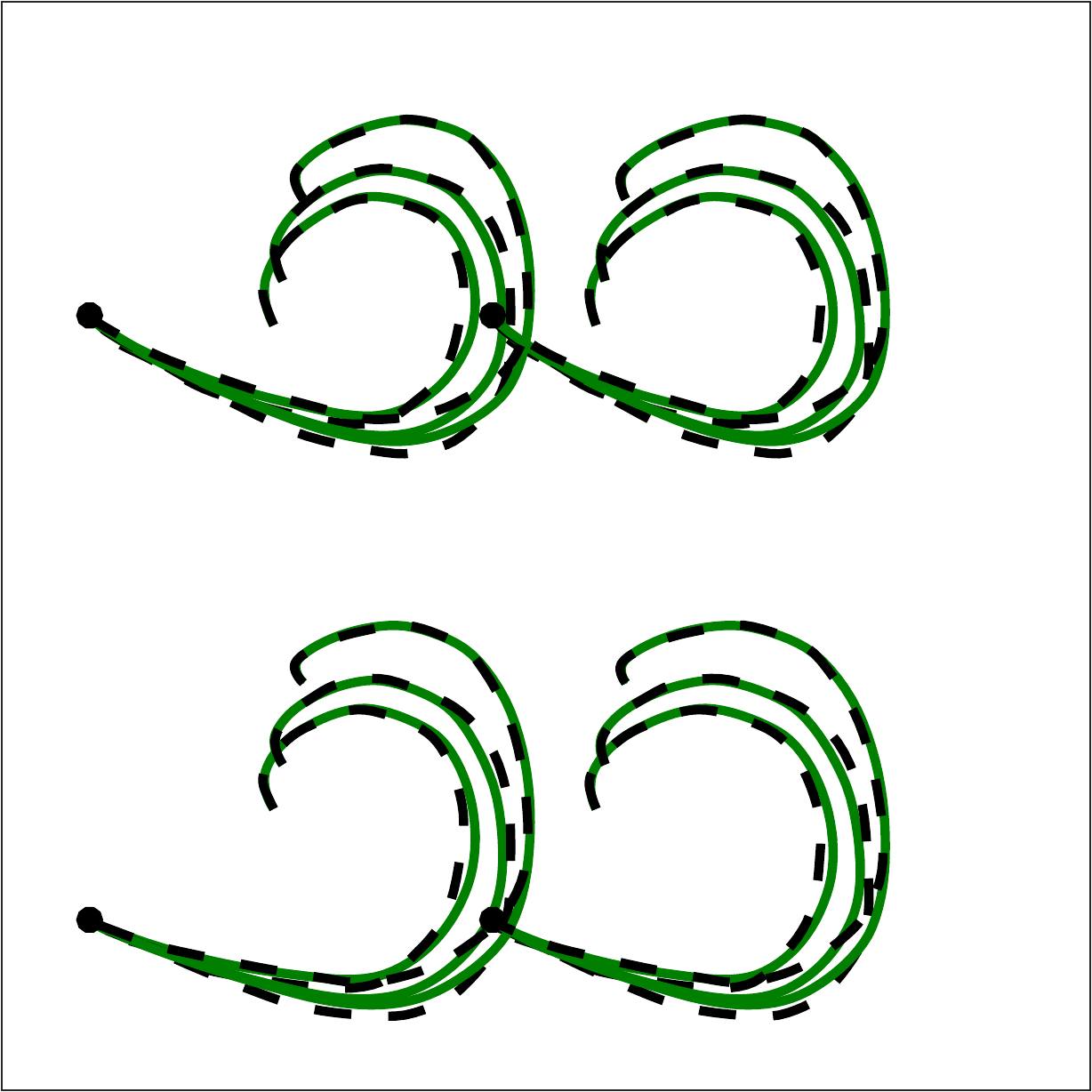}
    \hfill
    \includegraphics[width=\motionClassSize\textwidth]{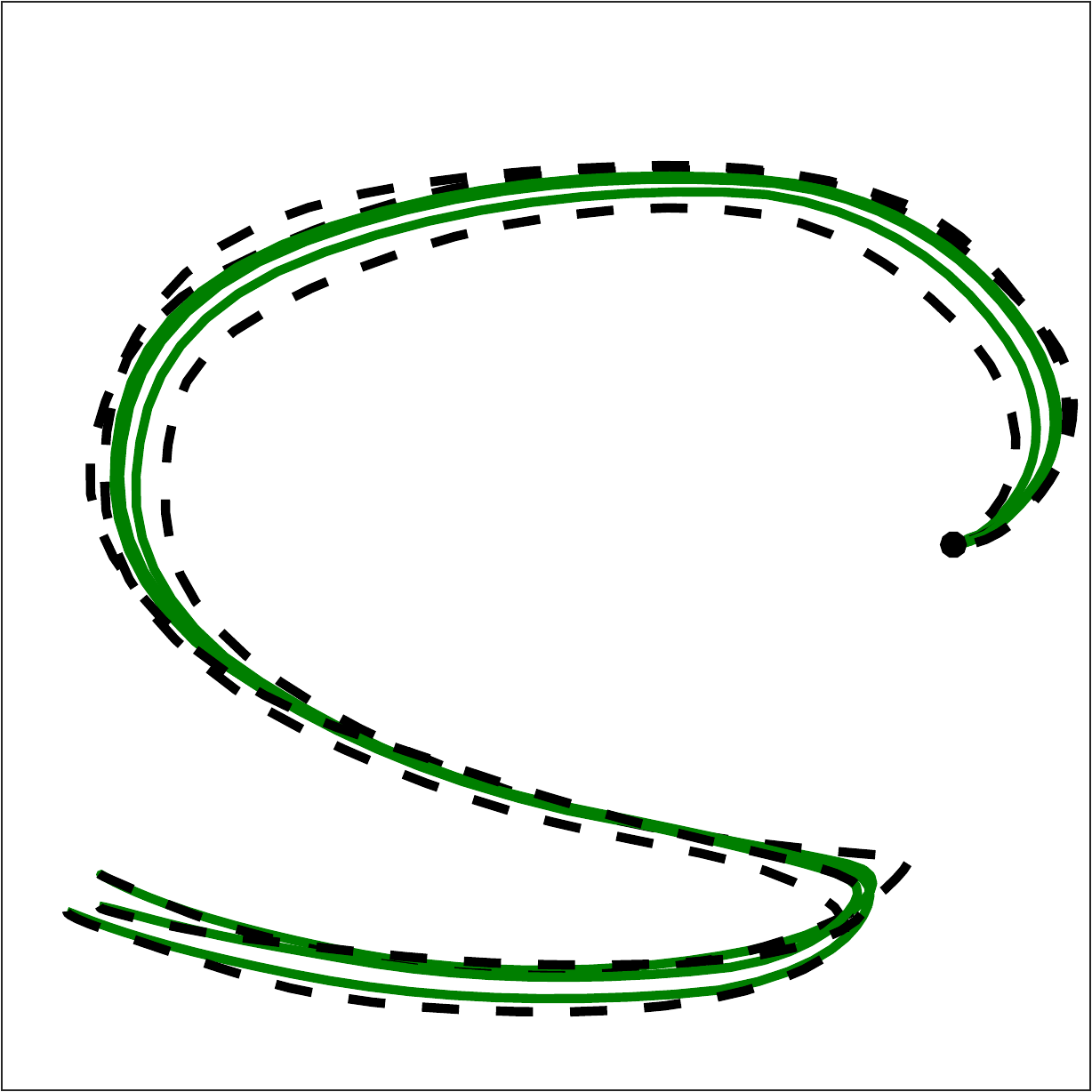}
    \hspace{-0.5mm}
    \includegraphics[width=\motionClassSize\textwidth]{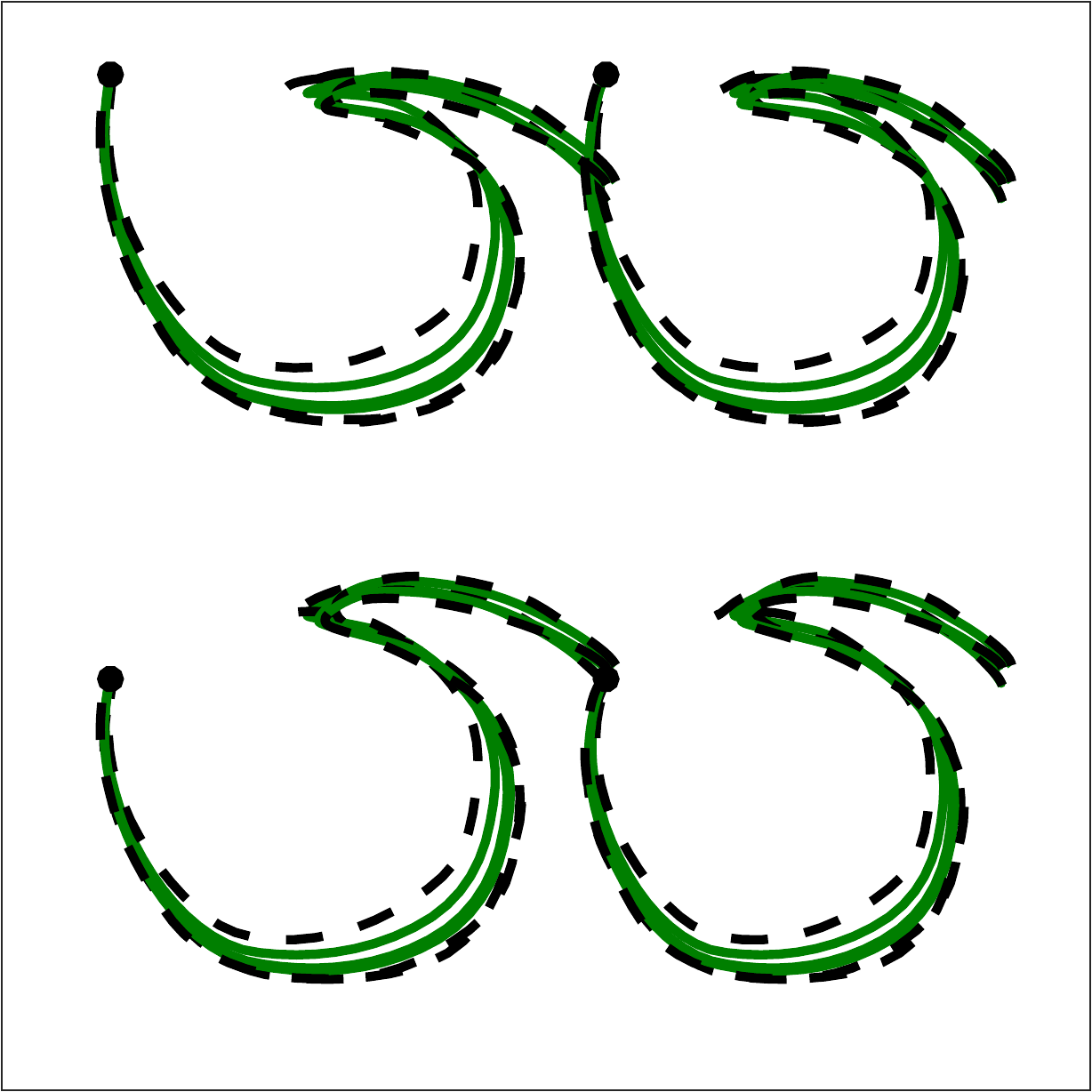}
    \hfill
    \includegraphics[width=\motionClassSize\textwidth]{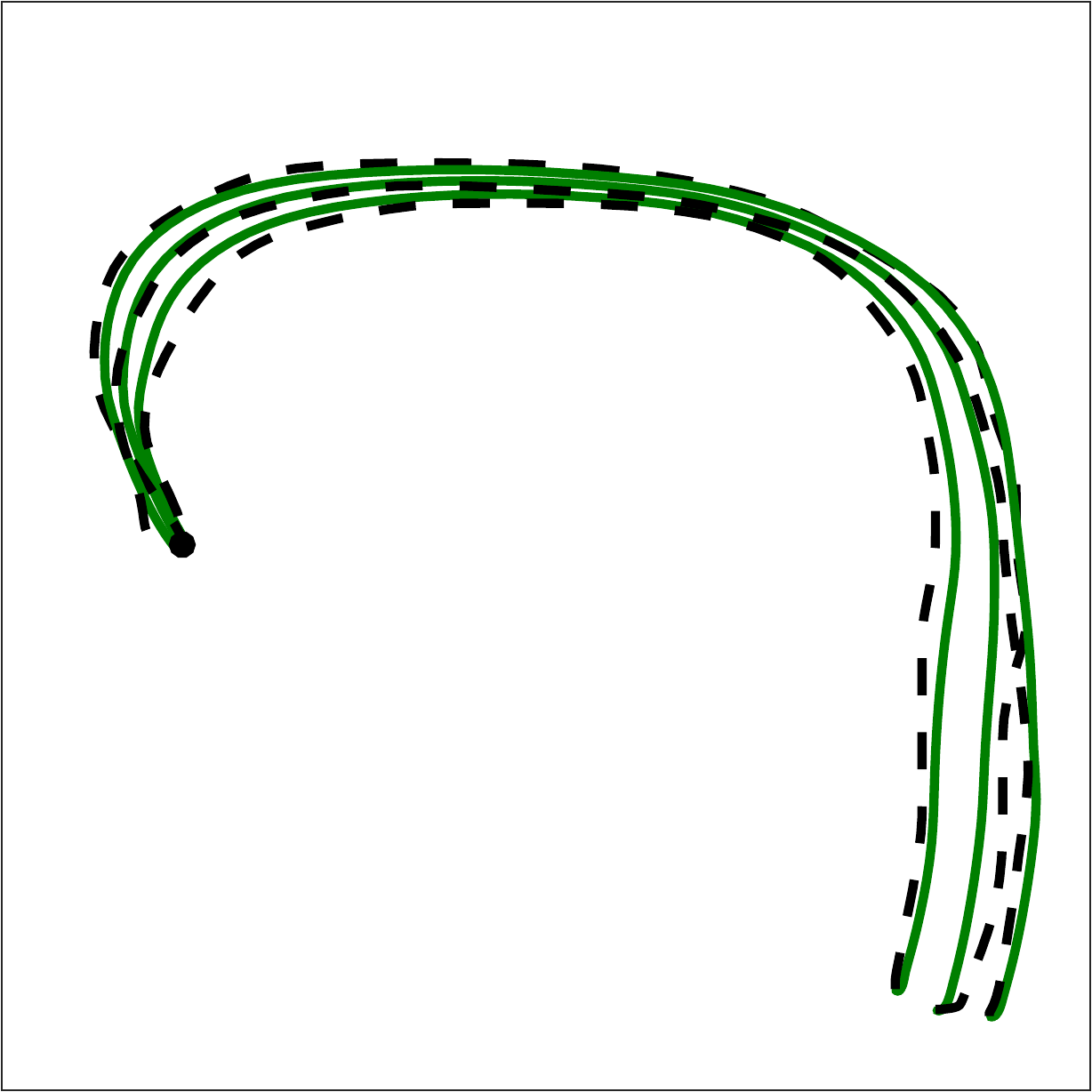}
    \hspace{-0.5mm}
    \includegraphics[width=\motionClassSize\textwidth]{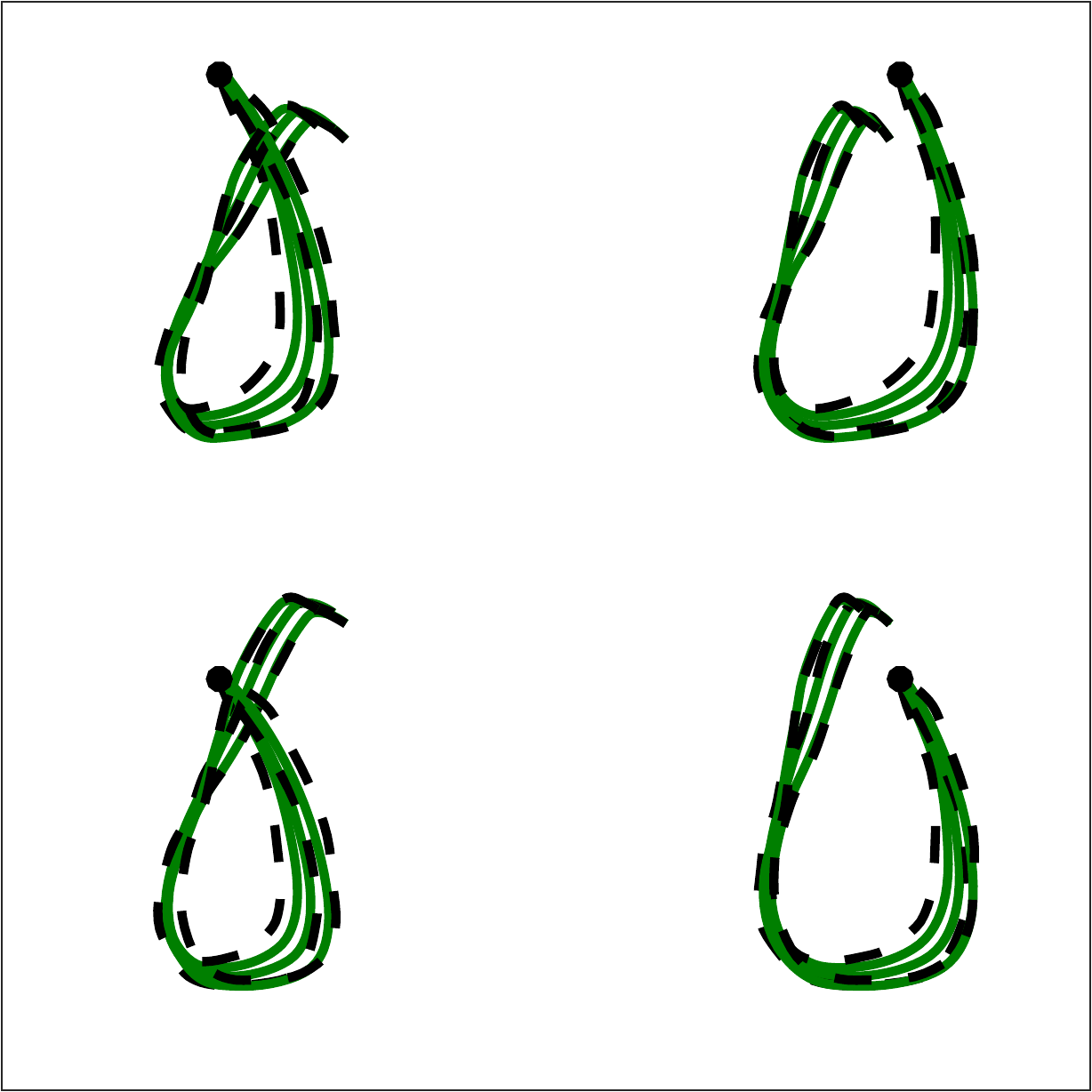}
    \hspace{5mm}~ 
    \\%
    \vspace{2mm}%
    \hspace{5mm}
    \includegraphics[width=\motionClassSize\textwidth]{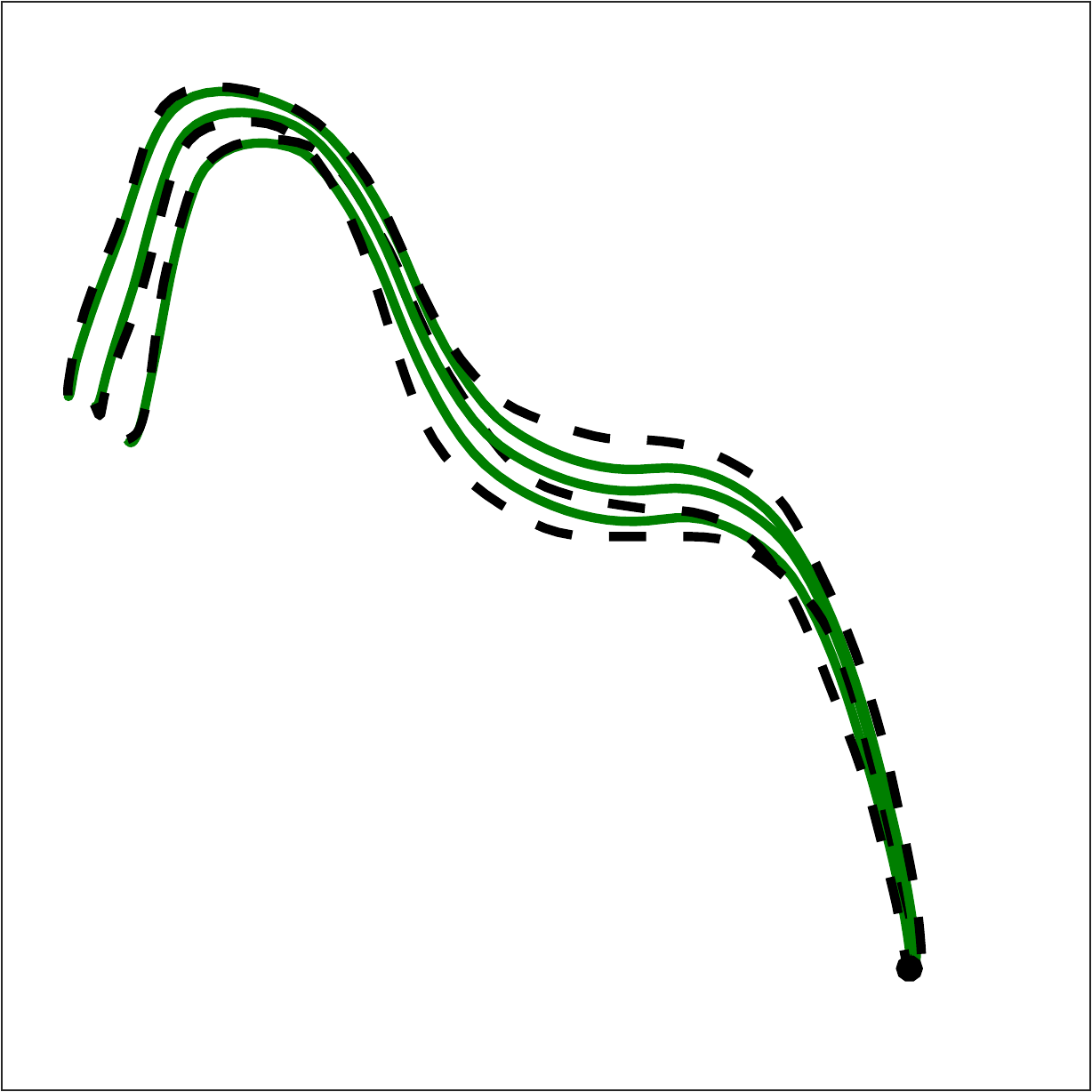}
    \hspace{-0.5mm}
    \includegraphics[width=\motionClassSize\textwidth]{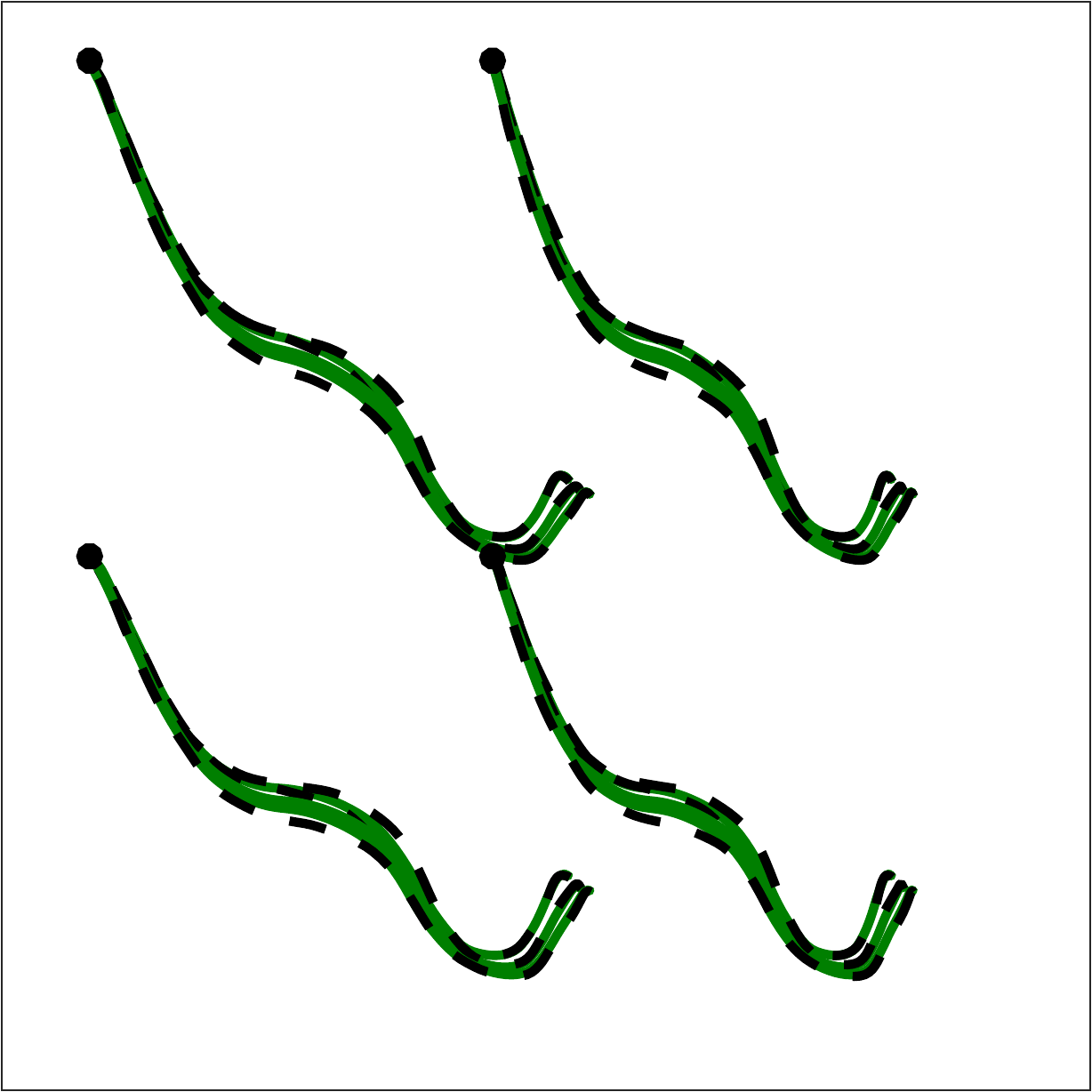}
    \hfill
    \includegraphics[width=\motionClassSize\textwidth]{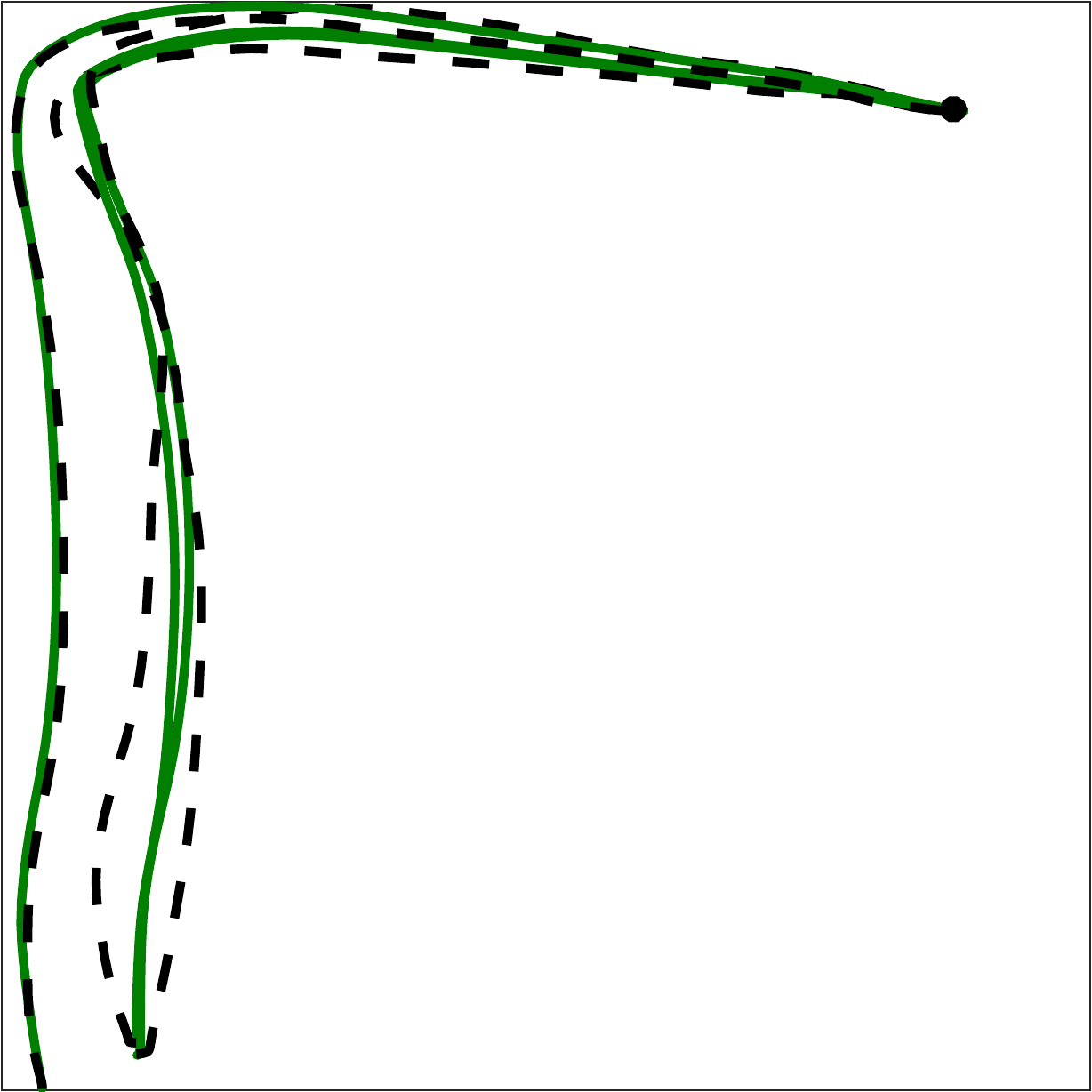}
    \hspace{-0.5mm}
    \includegraphics[width=\motionClassSize\textwidth]{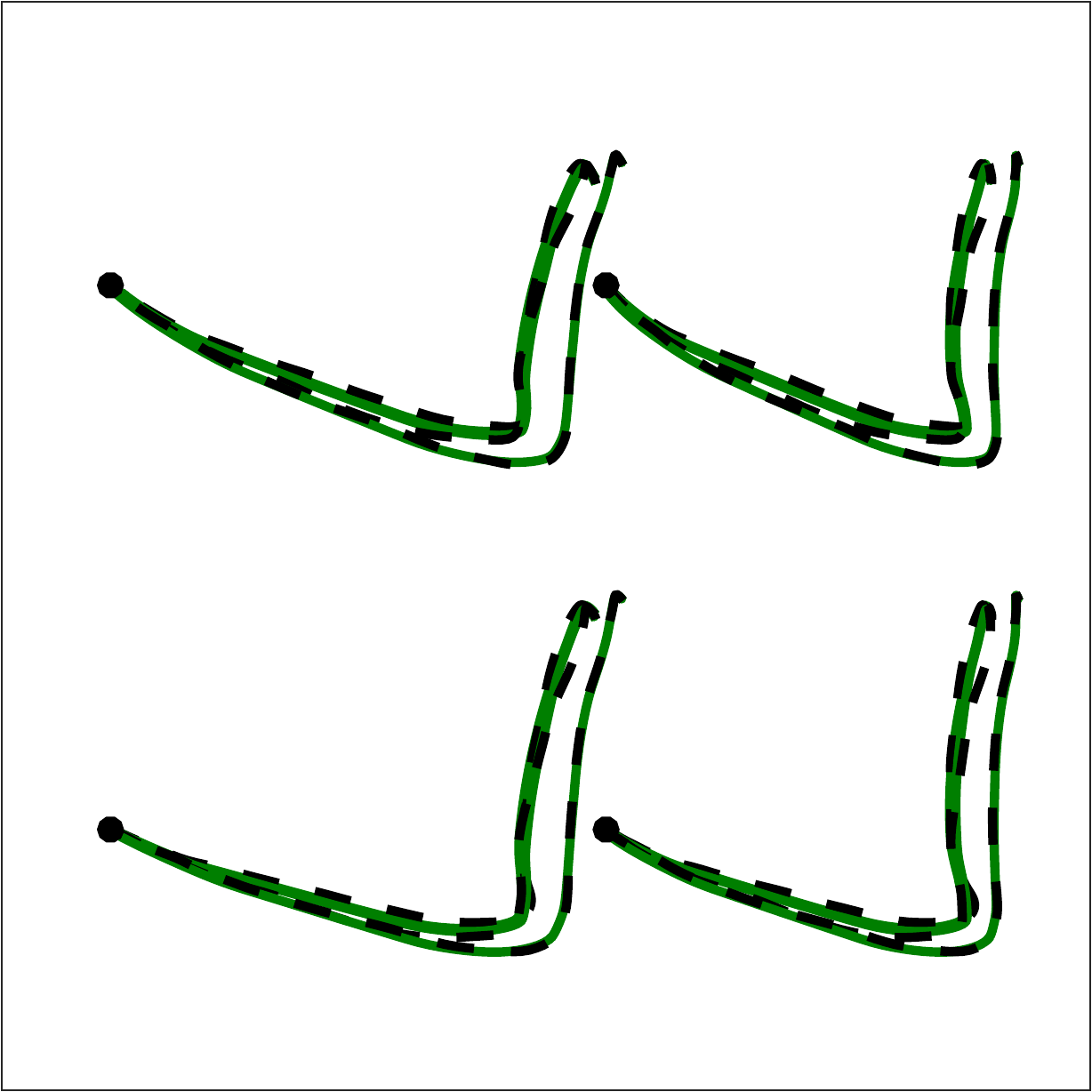}
    \hfill
    \includegraphics[width=\motionClassSize\textwidth]{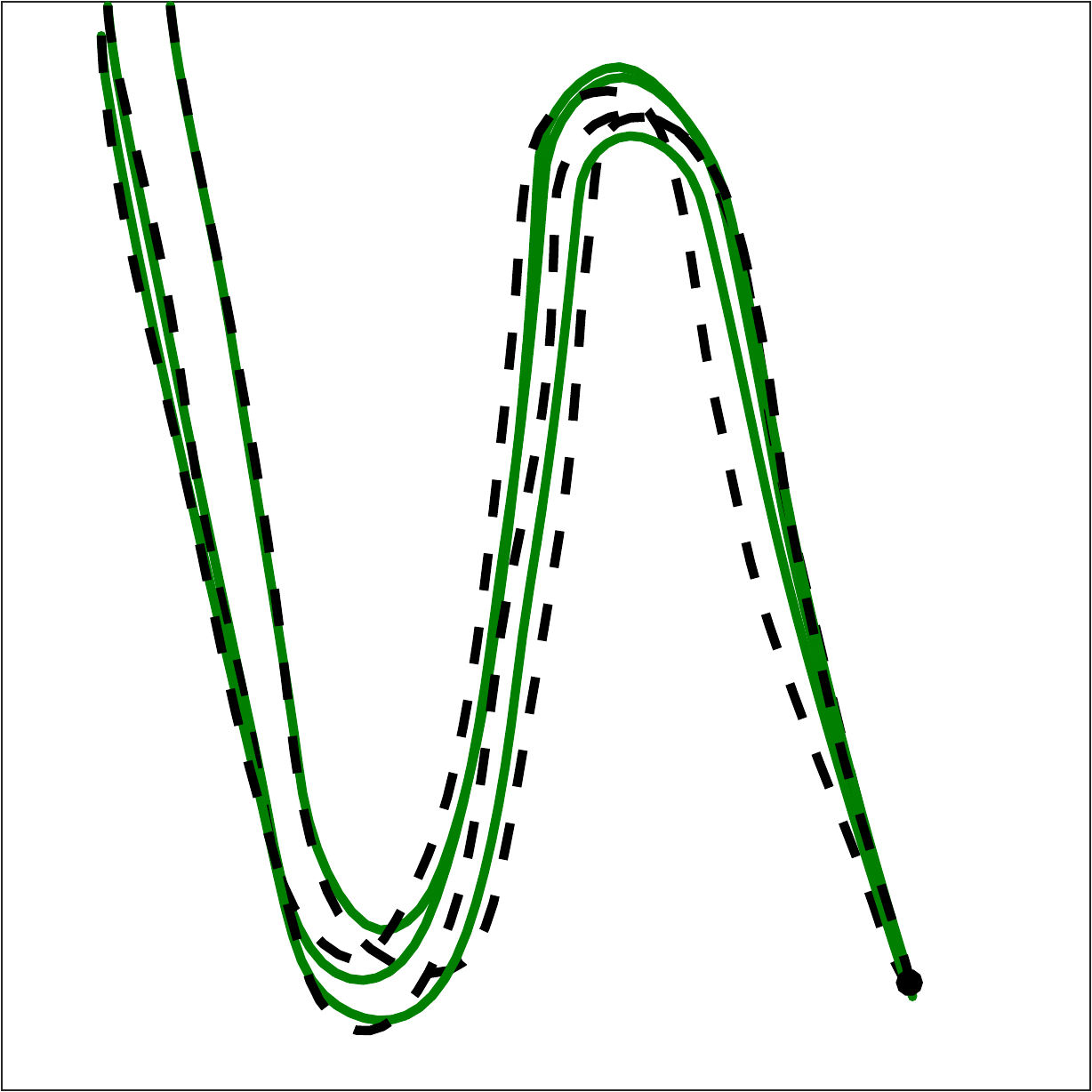}
    \hspace{-0.5mm}
    \includegraphics[width=\motionClassSize\textwidth]{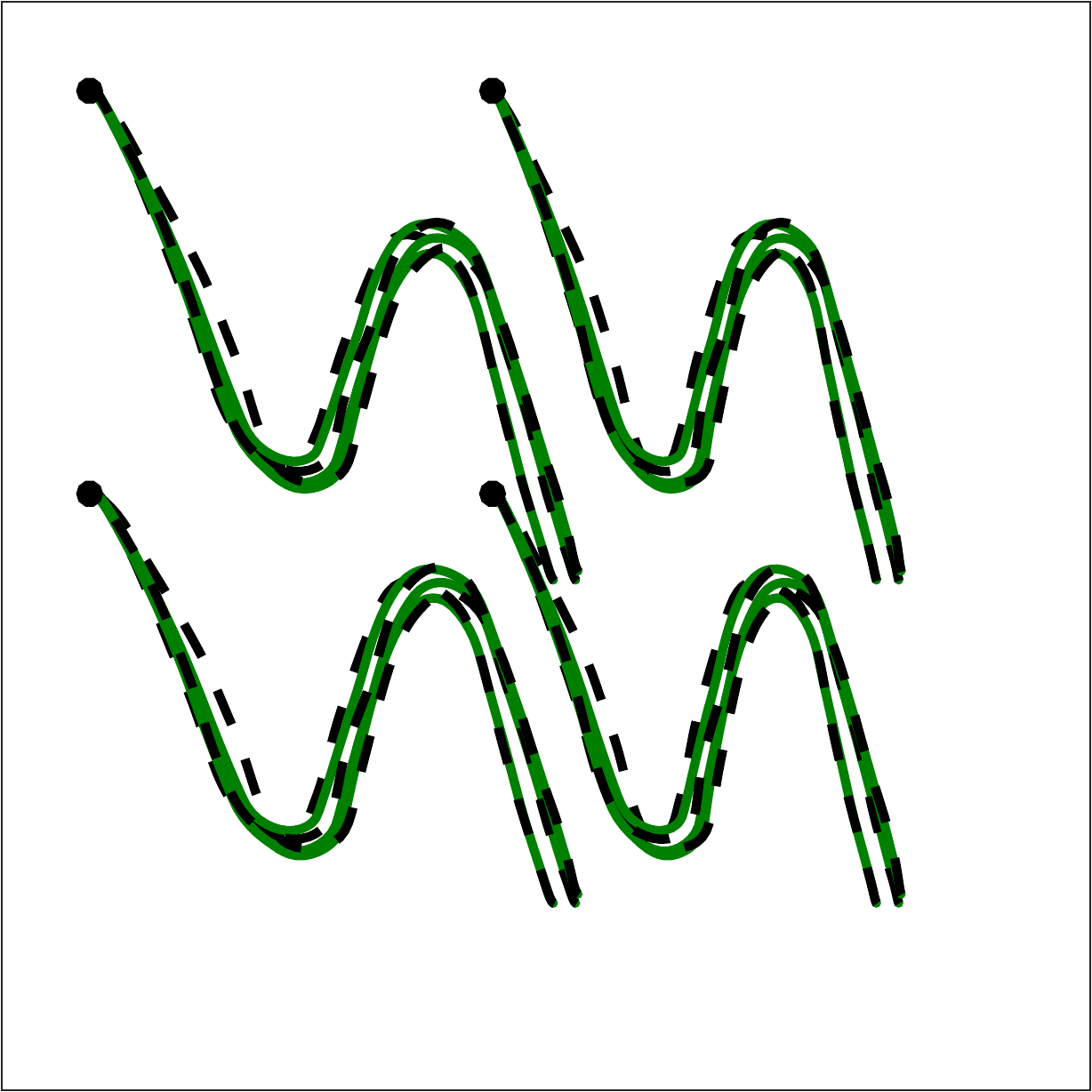}
    \hfill
    \includegraphics[width=\motionClassSize\textwidth]{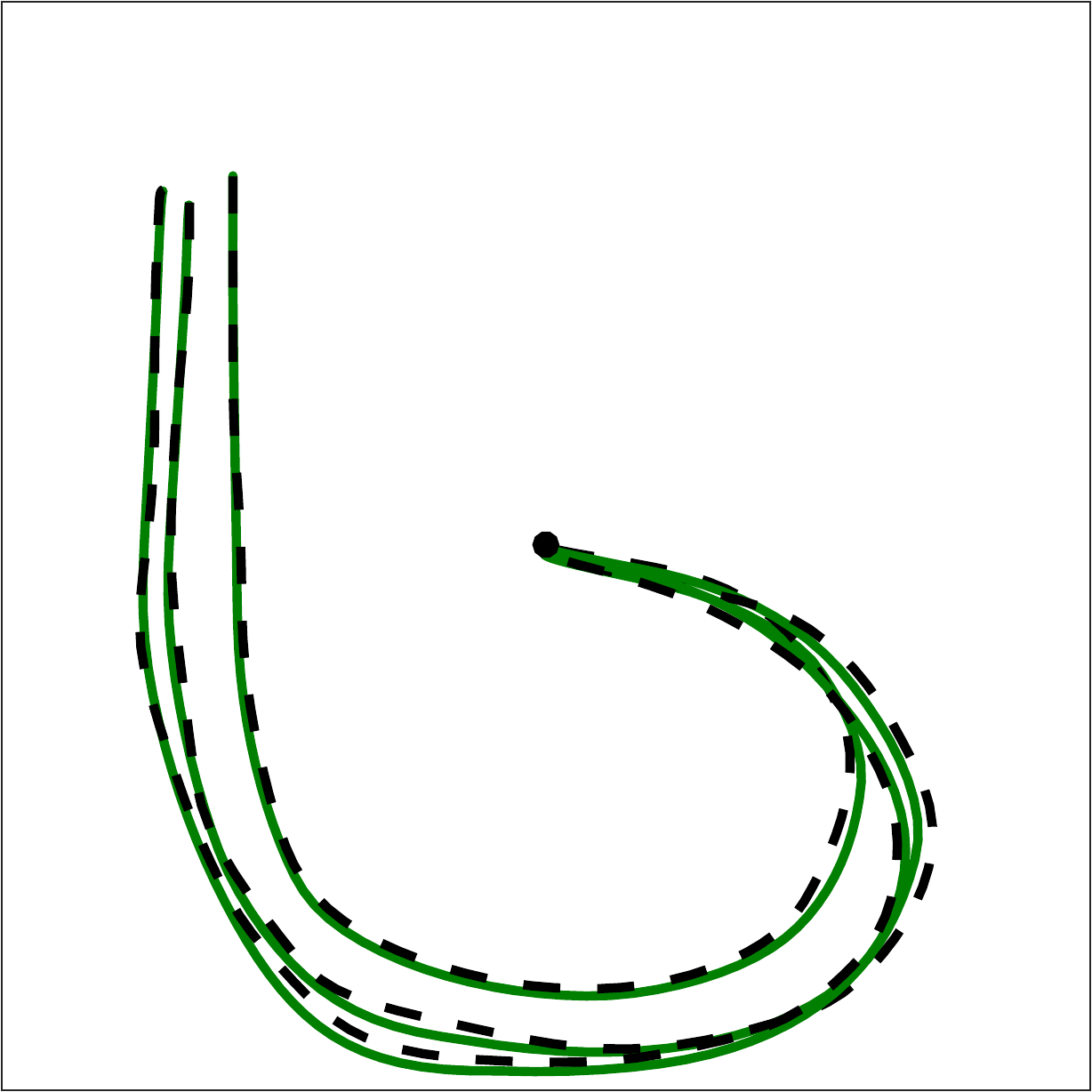}
    \hspace{-0.5mm}
    \includegraphics[width=\motionClassSize\textwidth]{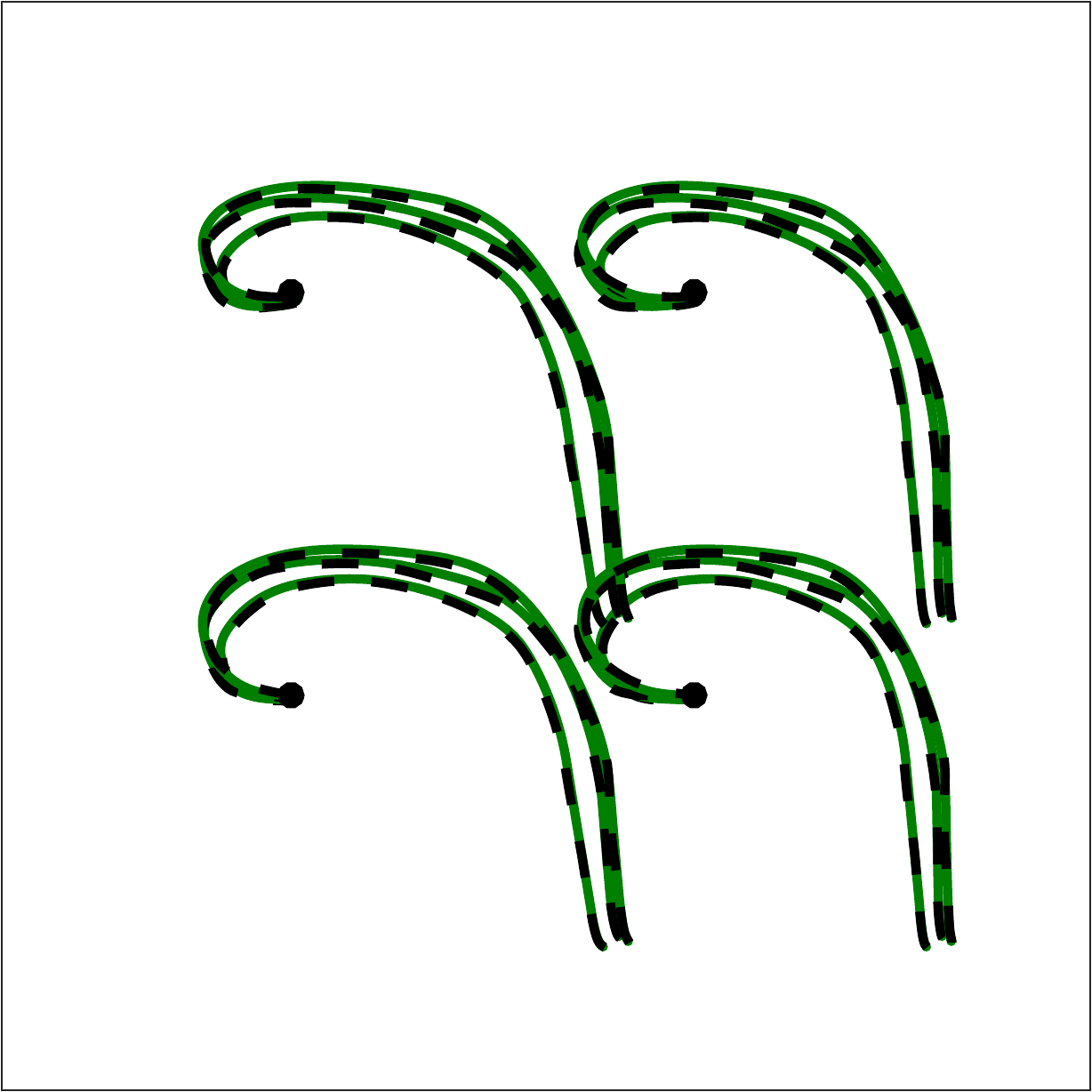}
    \hfill
    \includegraphics[width=\motionClassSize\textwidth]{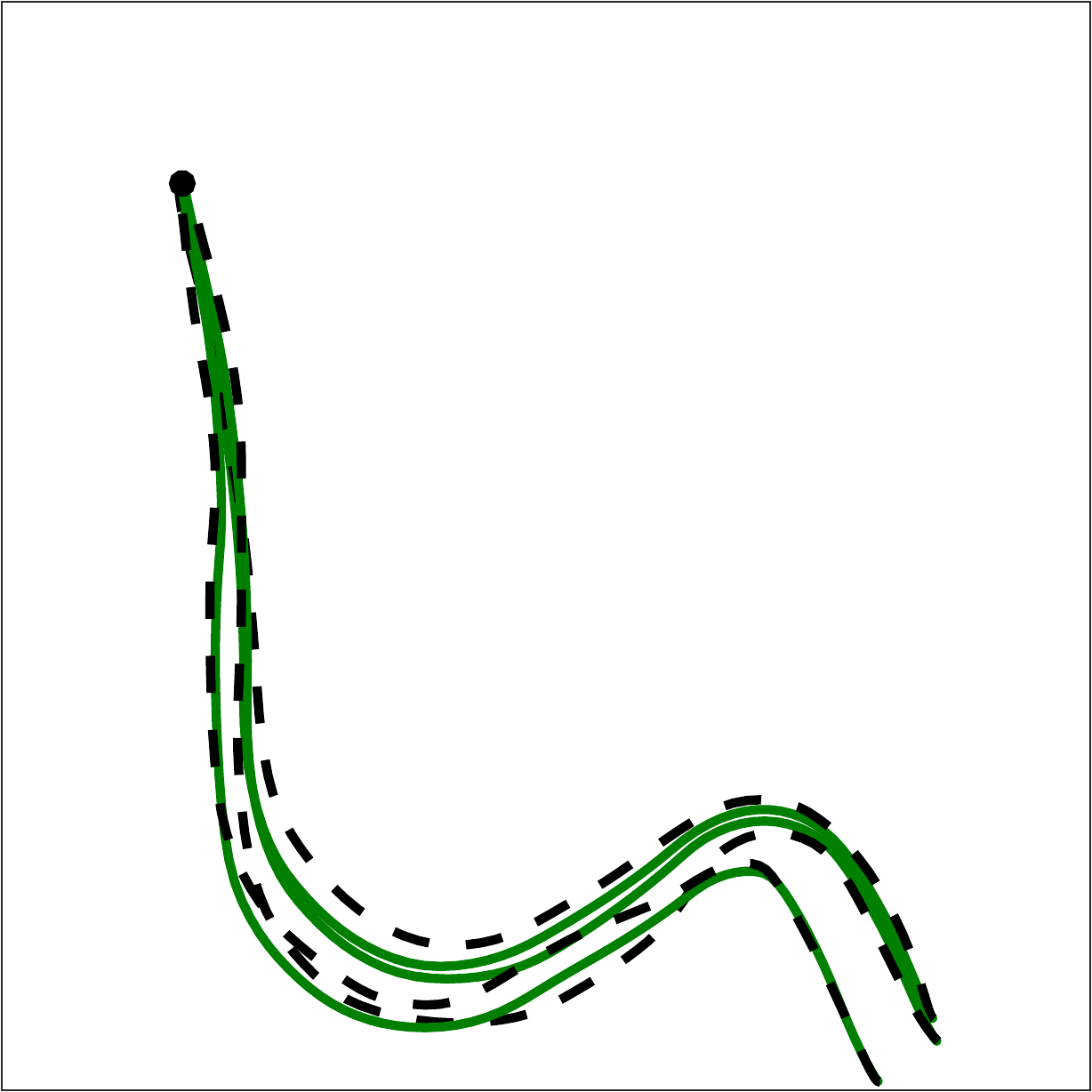}
    \hspace{-0.5mm}
    \includegraphics[width=\motionClassSize\textwidth]{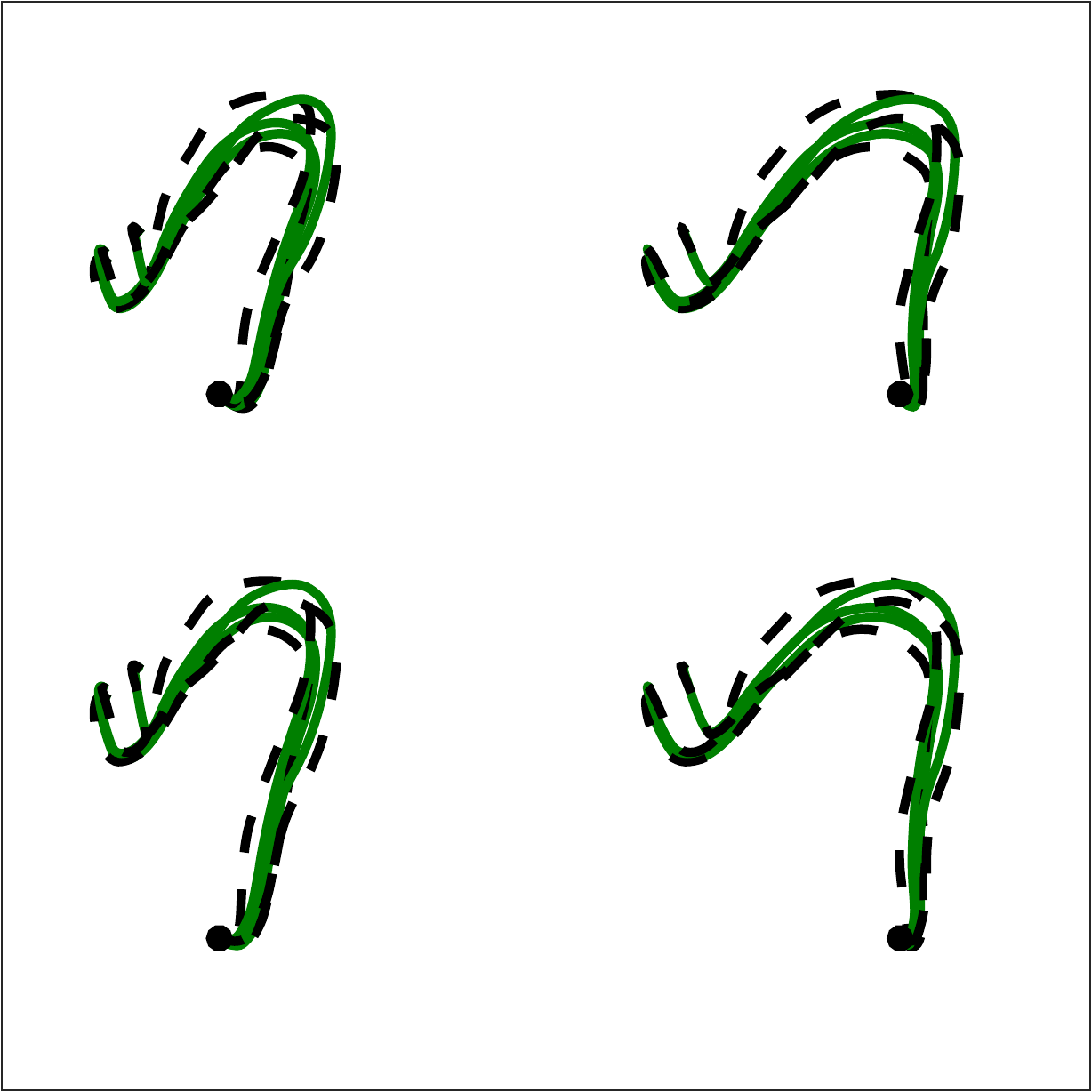}
    \hspace{5mm}~
    \\%
    \vspace{2mm}%
    \hspace{5mm}
    \includegraphics[width=\motionClassSize\textwidth]{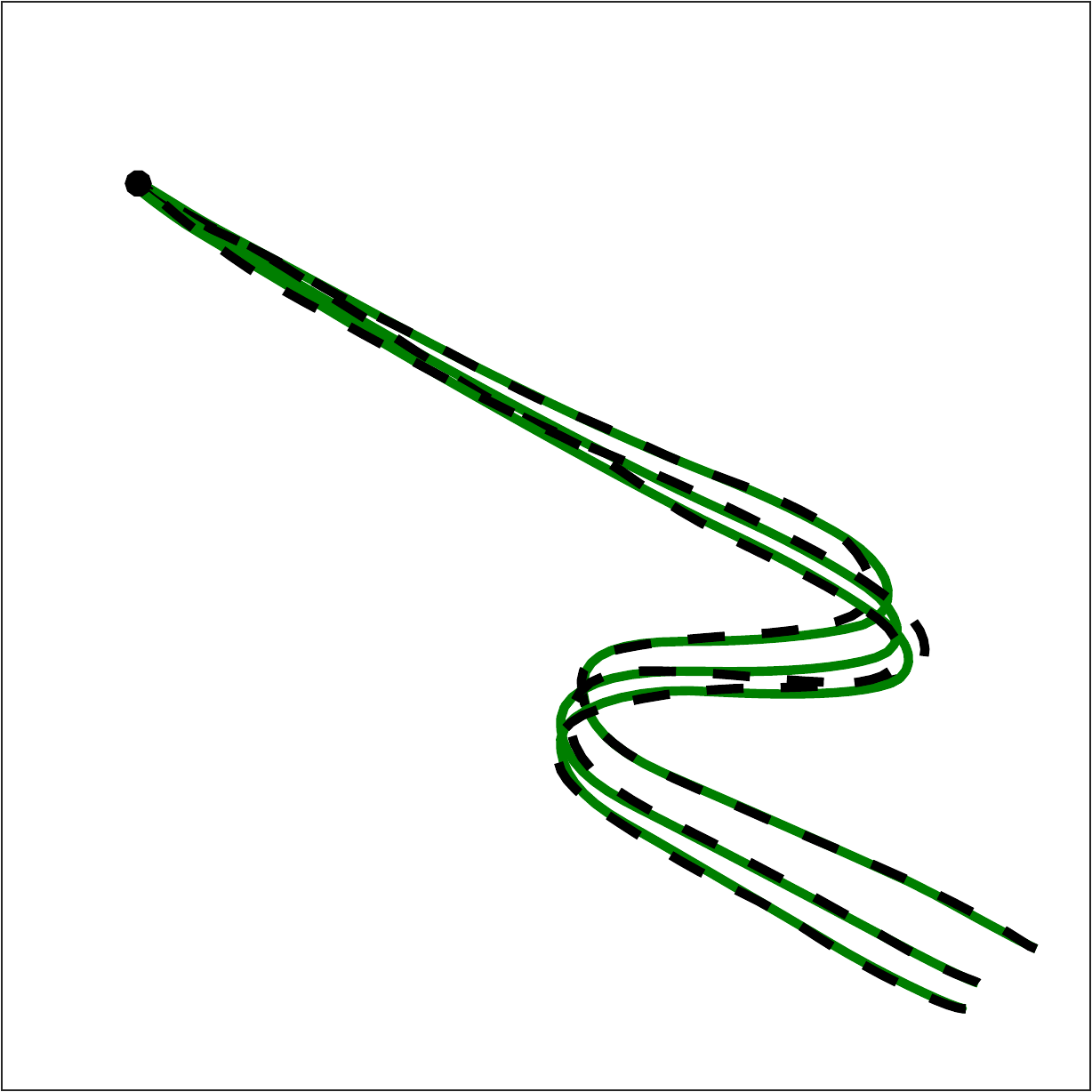}
    \hspace{-0.5mm}
    \includegraphics[width=\motionClassSize\textwidth]{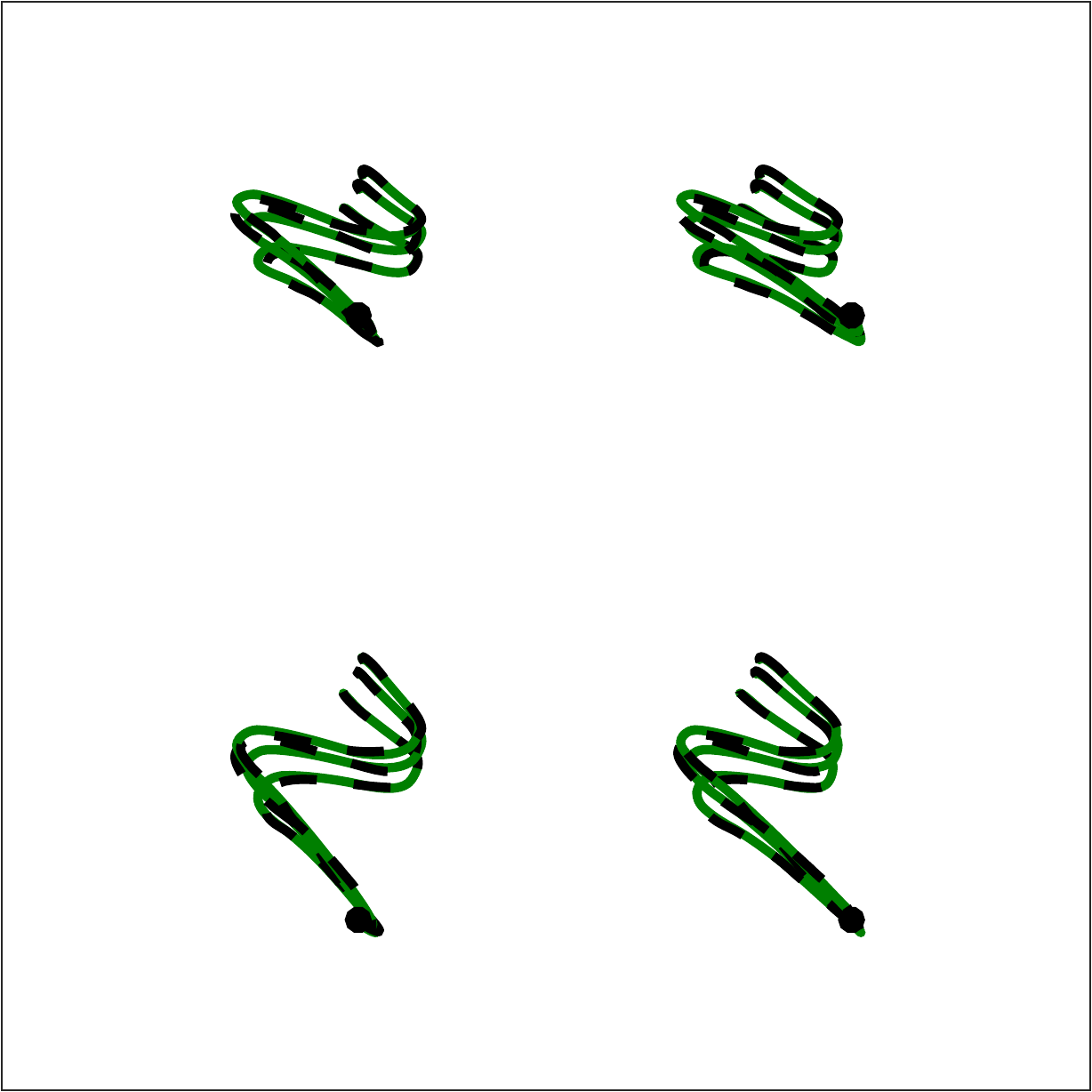}
    \hfill
    \includegraphics[width=\motionClassSize\textwidth]{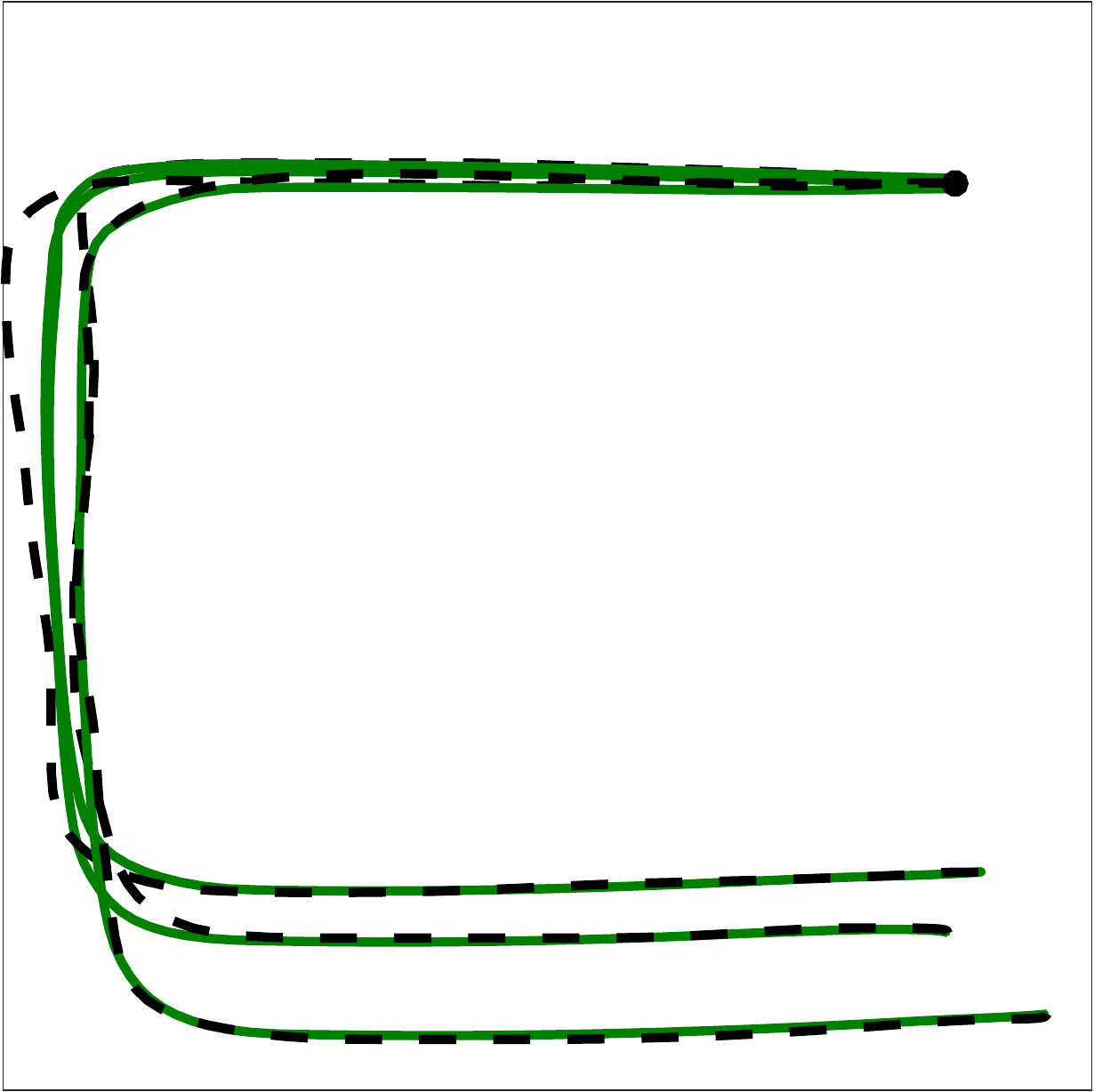}
    \hspace{-0.5mm}
    \includegraphics[width=\motionClassSize\textwidth]{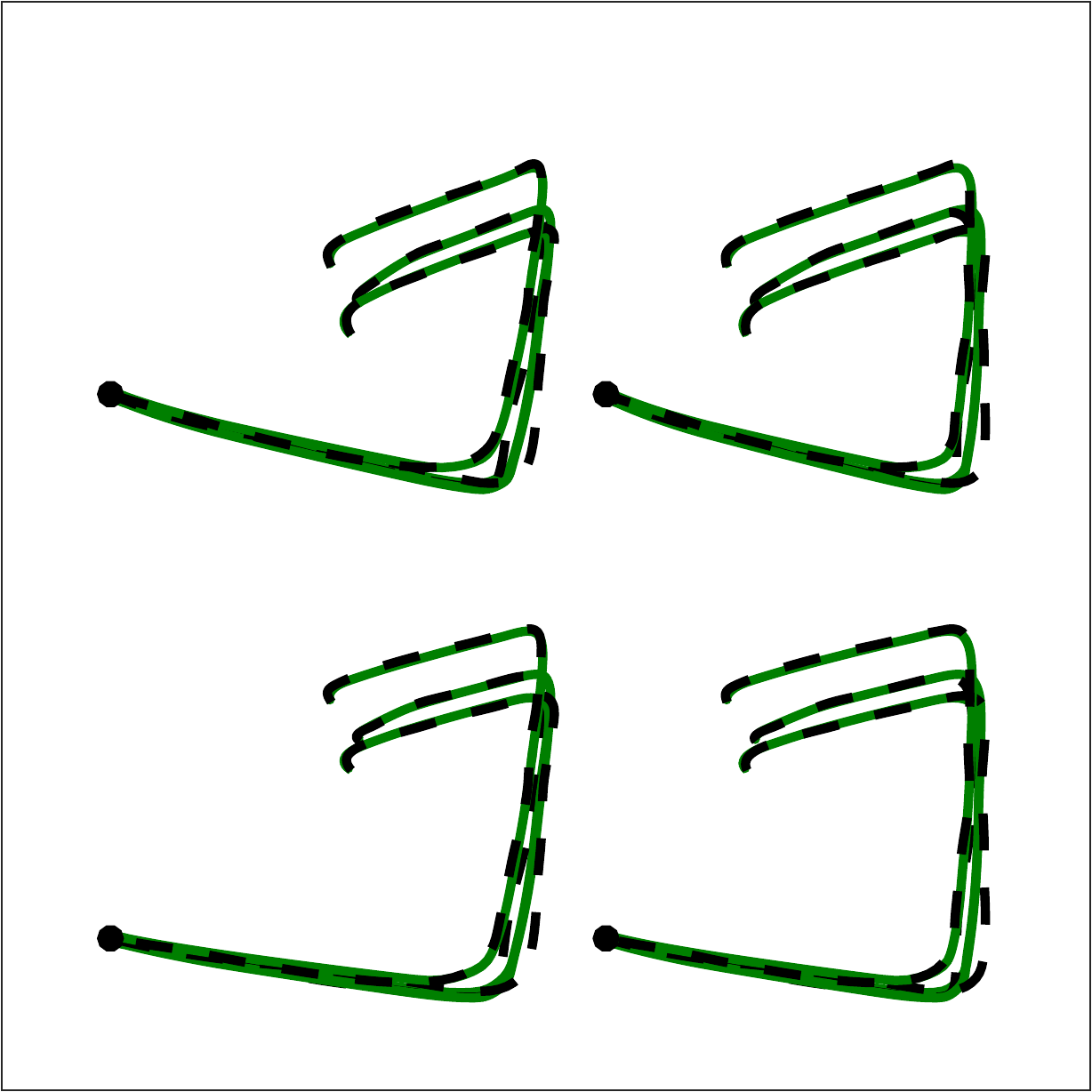}
    \hfill
    \includegraphics[width=\motionClassSize\textwidth]{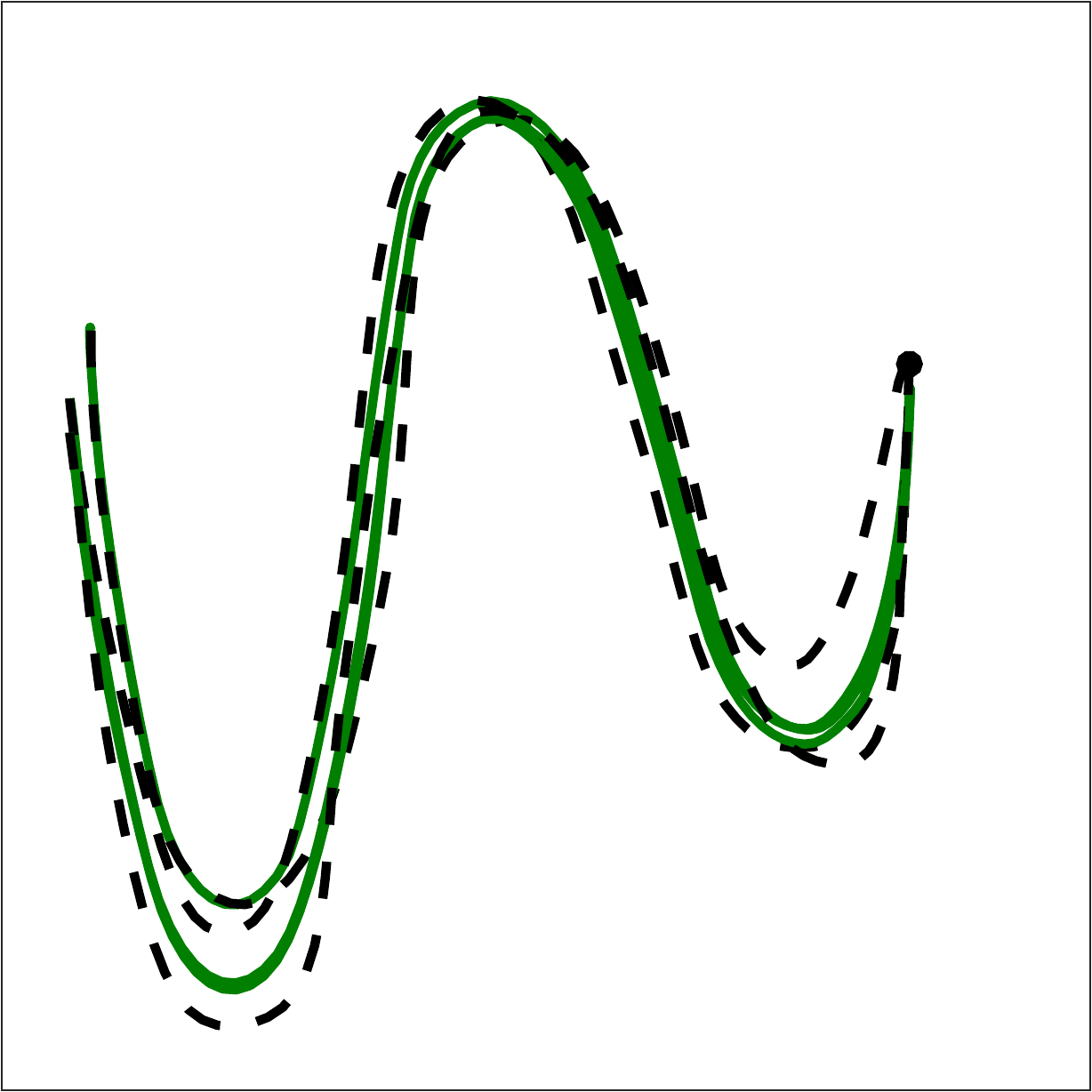}
    \hspace{-0.5mm}
    \includegraphics[width=\motionClassSize\textwidth]{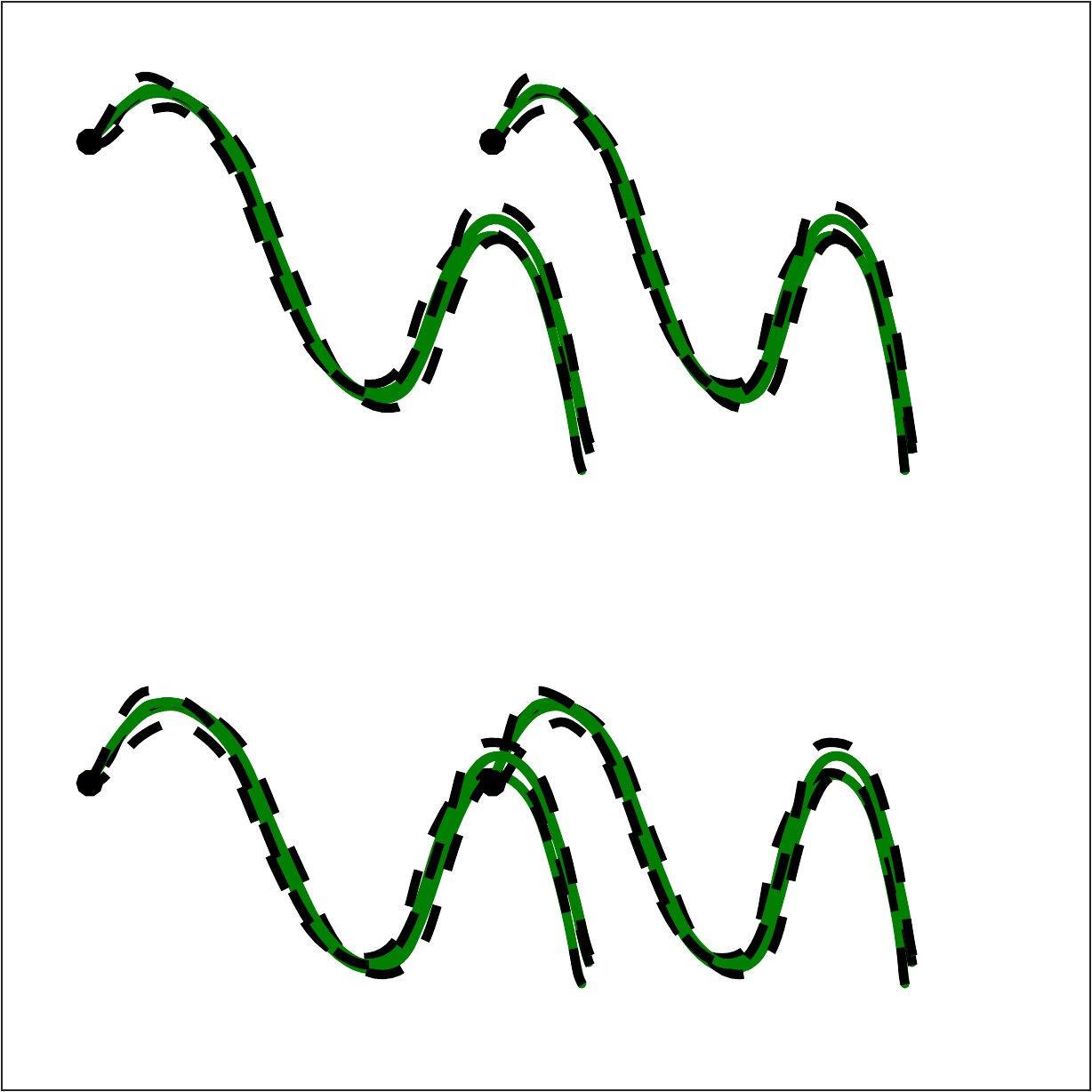}
    \hfill
    \includegraphics[width=\motionClassSize\textwidth]{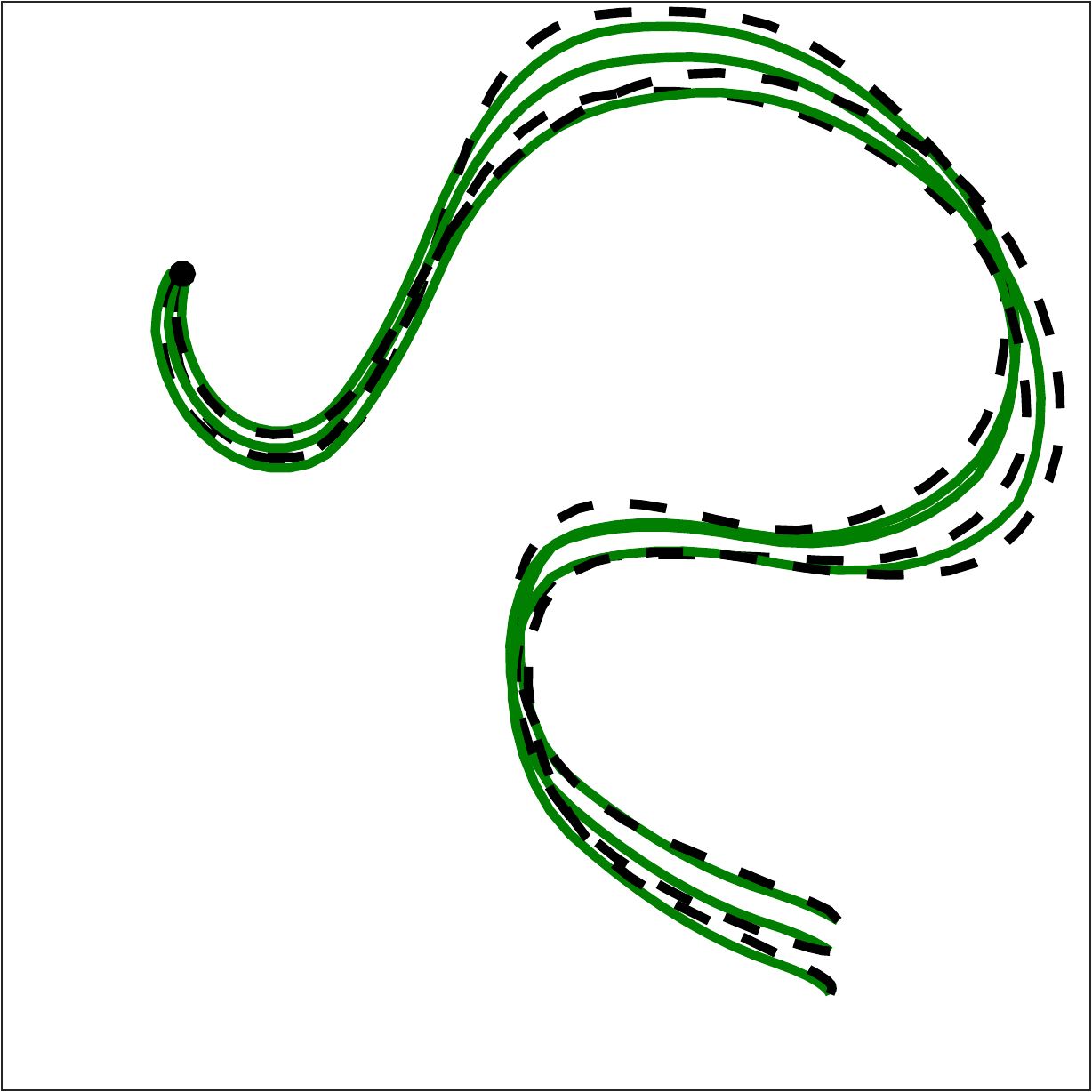}
    \hspace{-0.5mm}
    \includegraphics[width=\motionClassSize\textwidth]{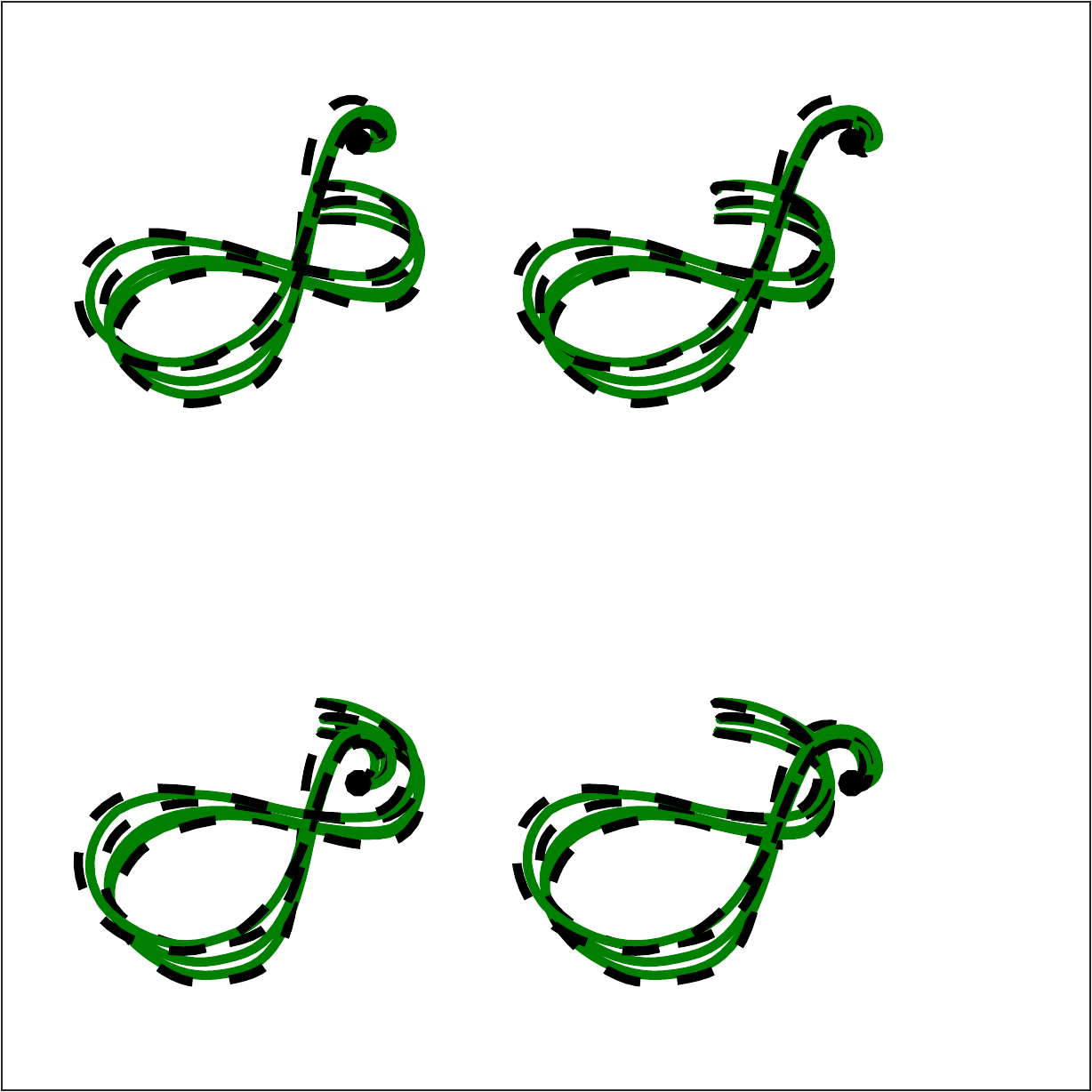}
    \hfill
    \includegraphics[width=\motionClassSize\textwidth]{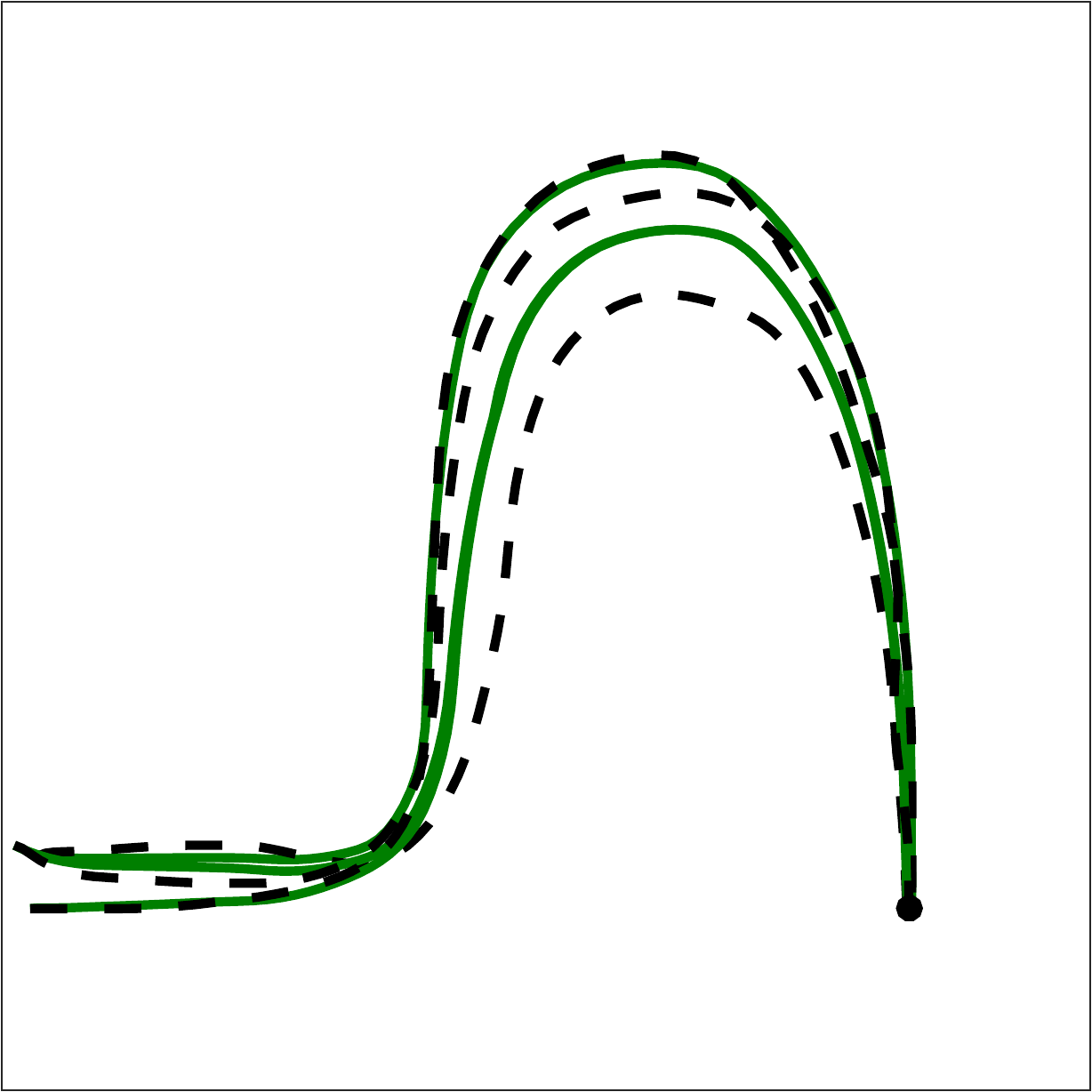}
    \hspace{-0.5mm}
    \includegraphics[width=\motionClassSize\textwidth]{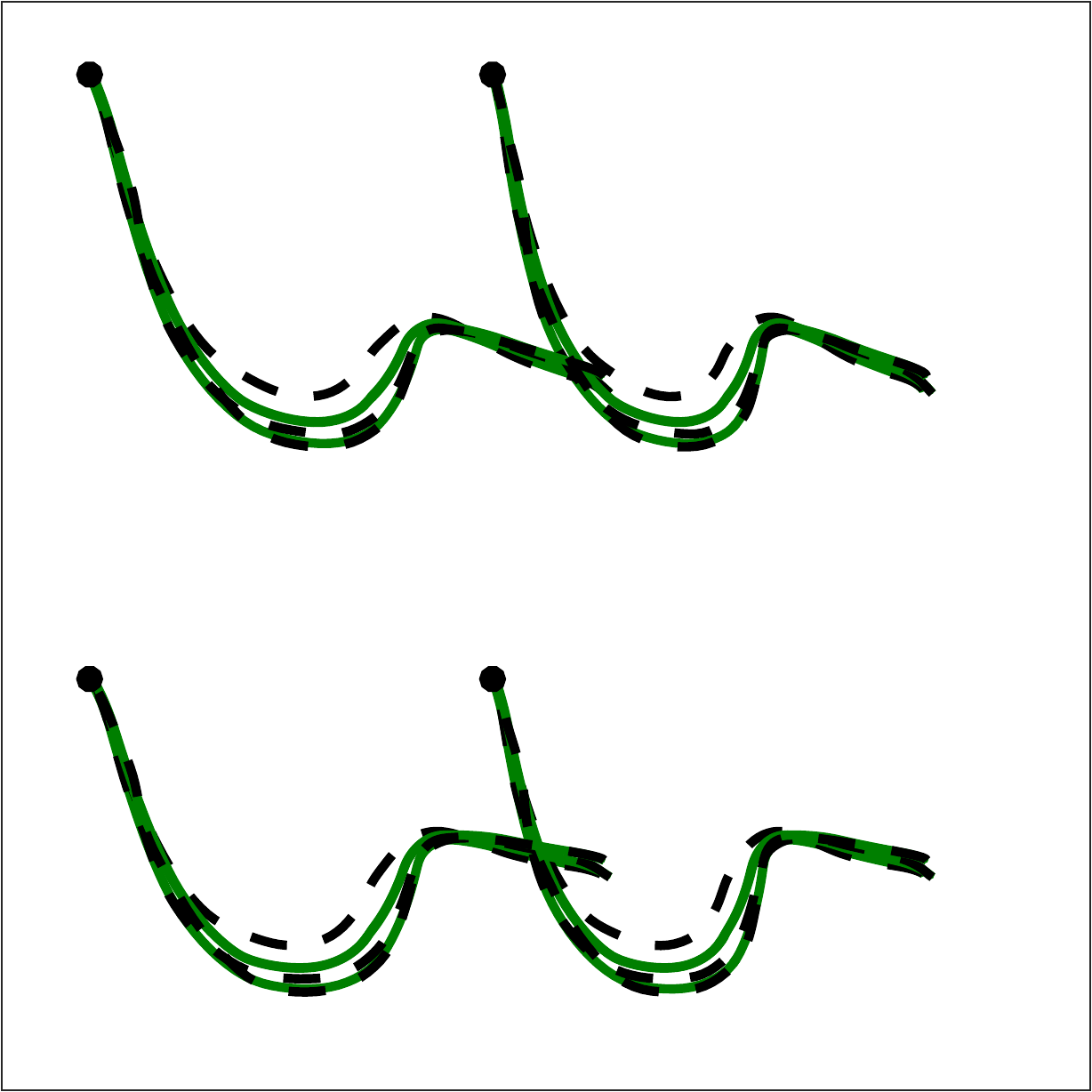}
    \hspace{5mm}~
    \\%
    \vspace{2mm}%
    \hspace{5mm}
    \includegraphics[width=\motionClassSize\textwidth]{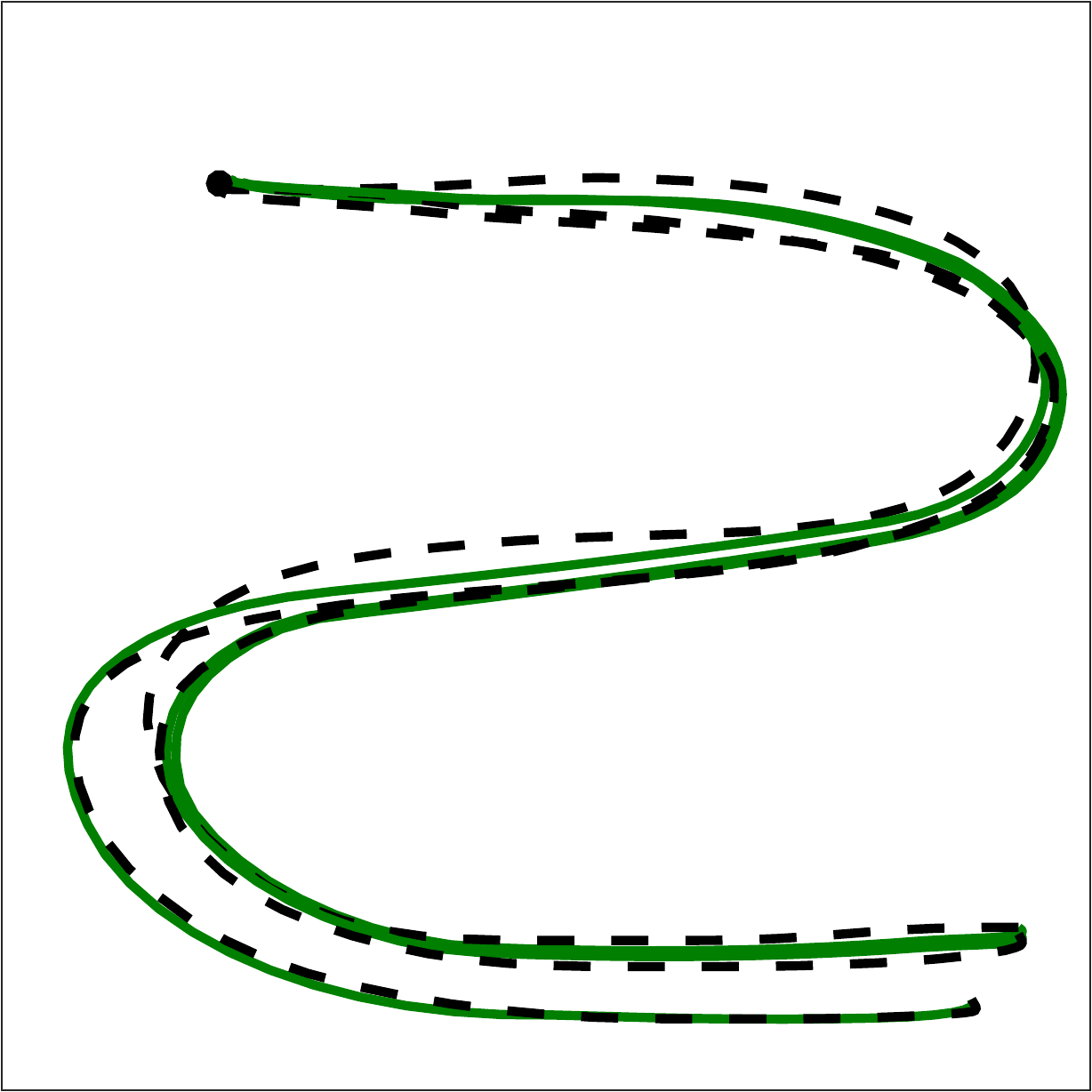}
    \hspace{-0.5mm}
    \includegraphics[width=\motionClassSize\textwidth]{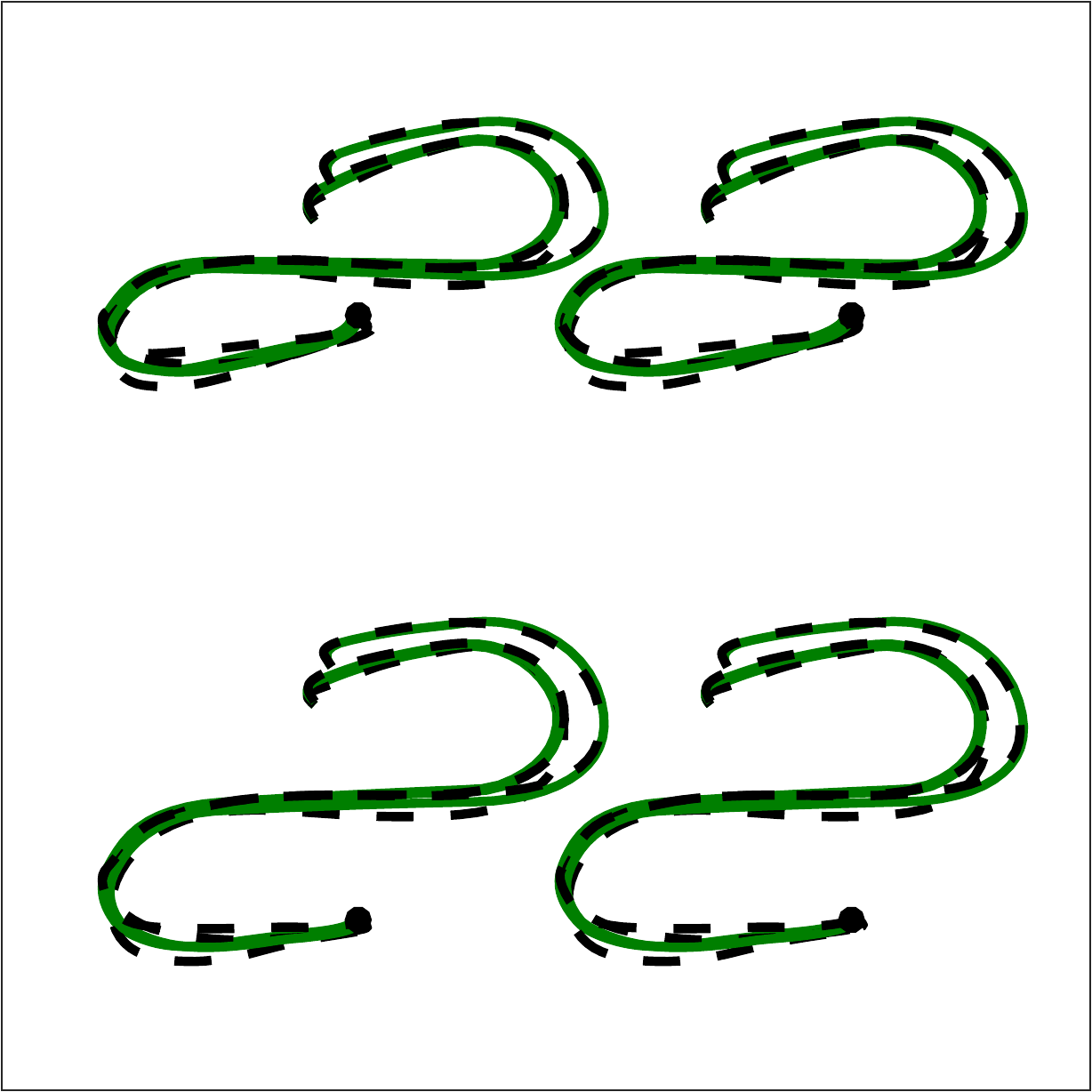}
    \hfill
    \includegraphics[width=\motionClassSize\textwidth]{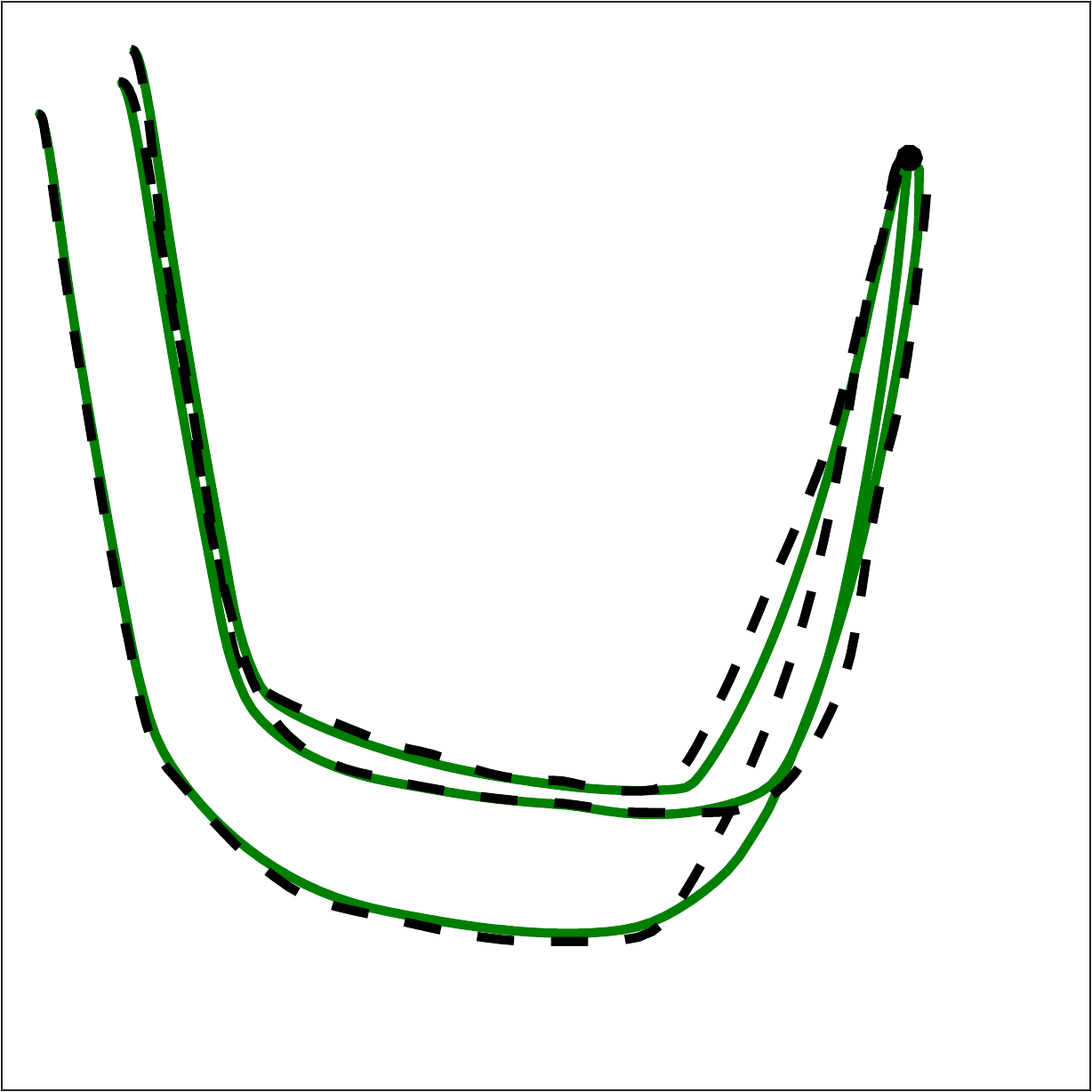}
    \hspace{-0.5mm}
    \includegraphics[width=\motionClassSize\textwidth]{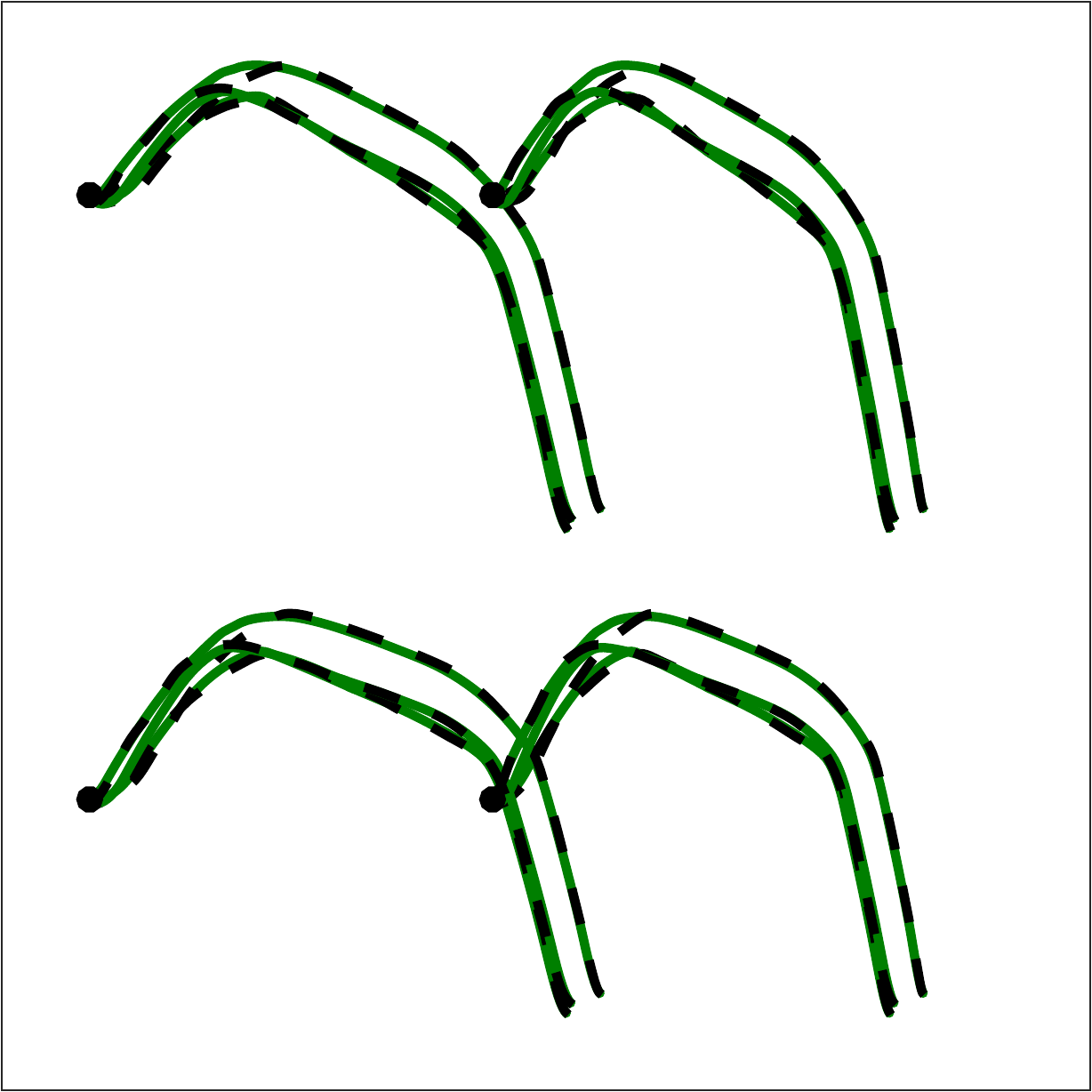}
    \hfill
    \includegraphics[width=\motionClassSize\textwidth]{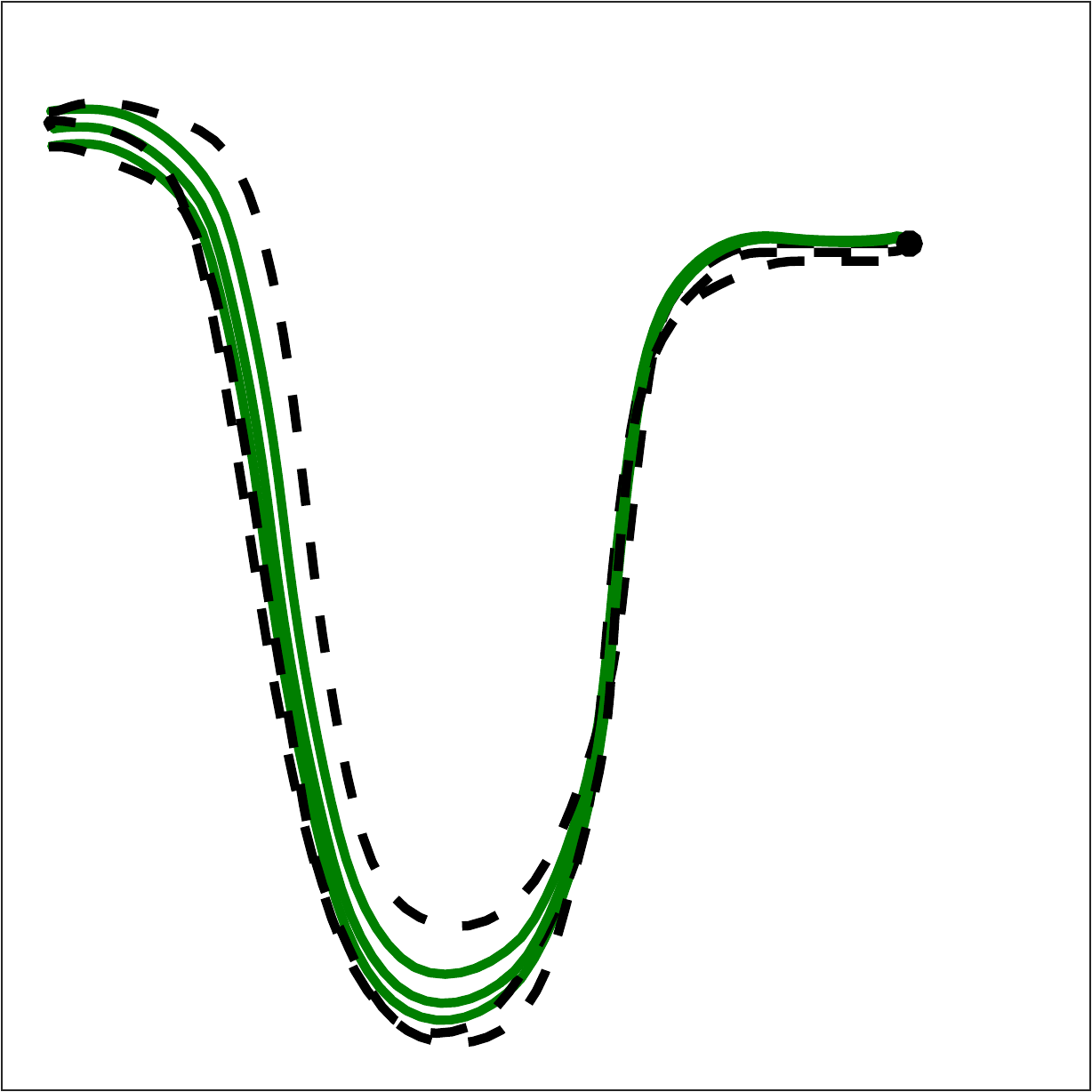}
    \hspace{-0.5mm}
    \includegraphics[width=\motionClassSize\textwidth]{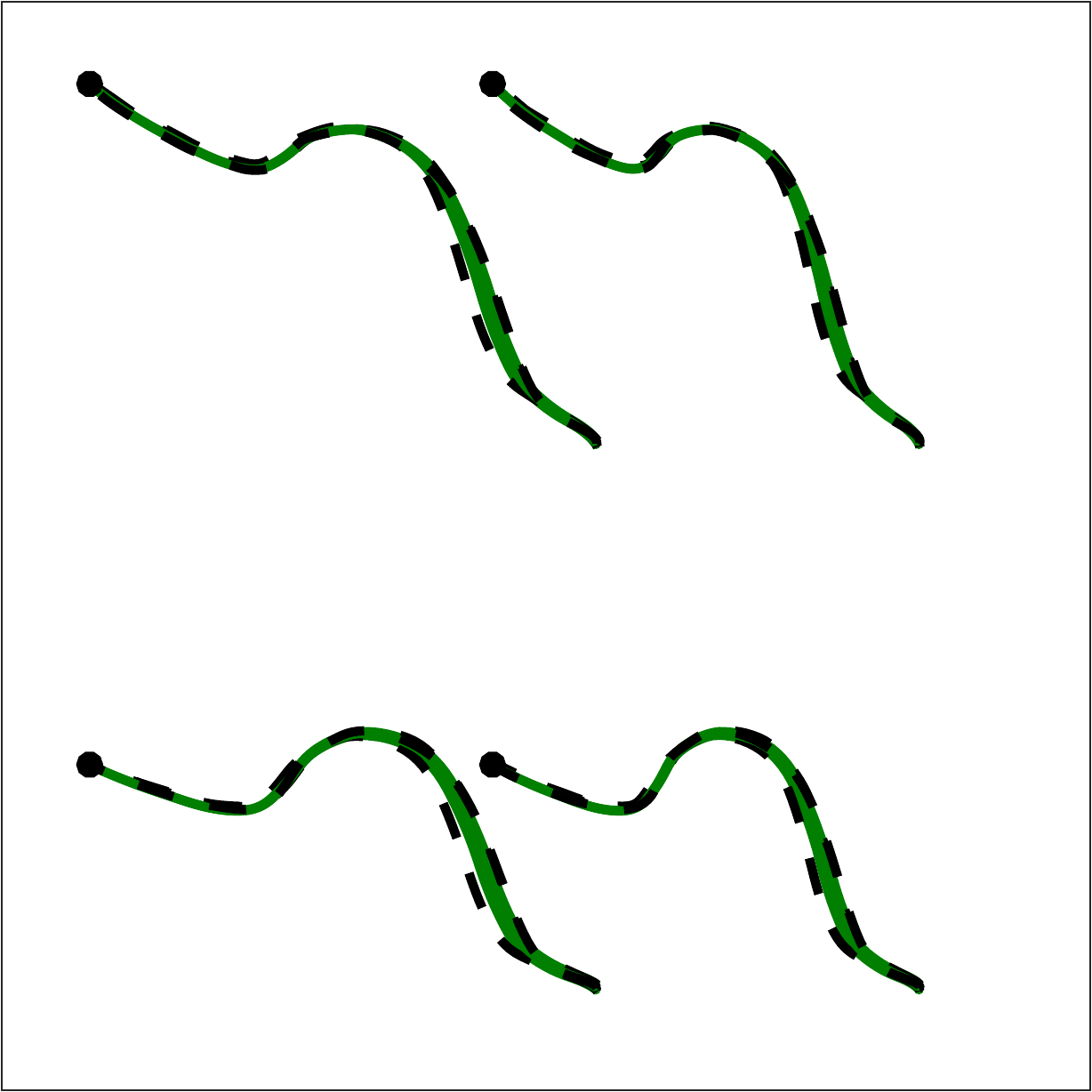}
    \hfill
    \includegraphics[width=\motionClassSize\textwidth]{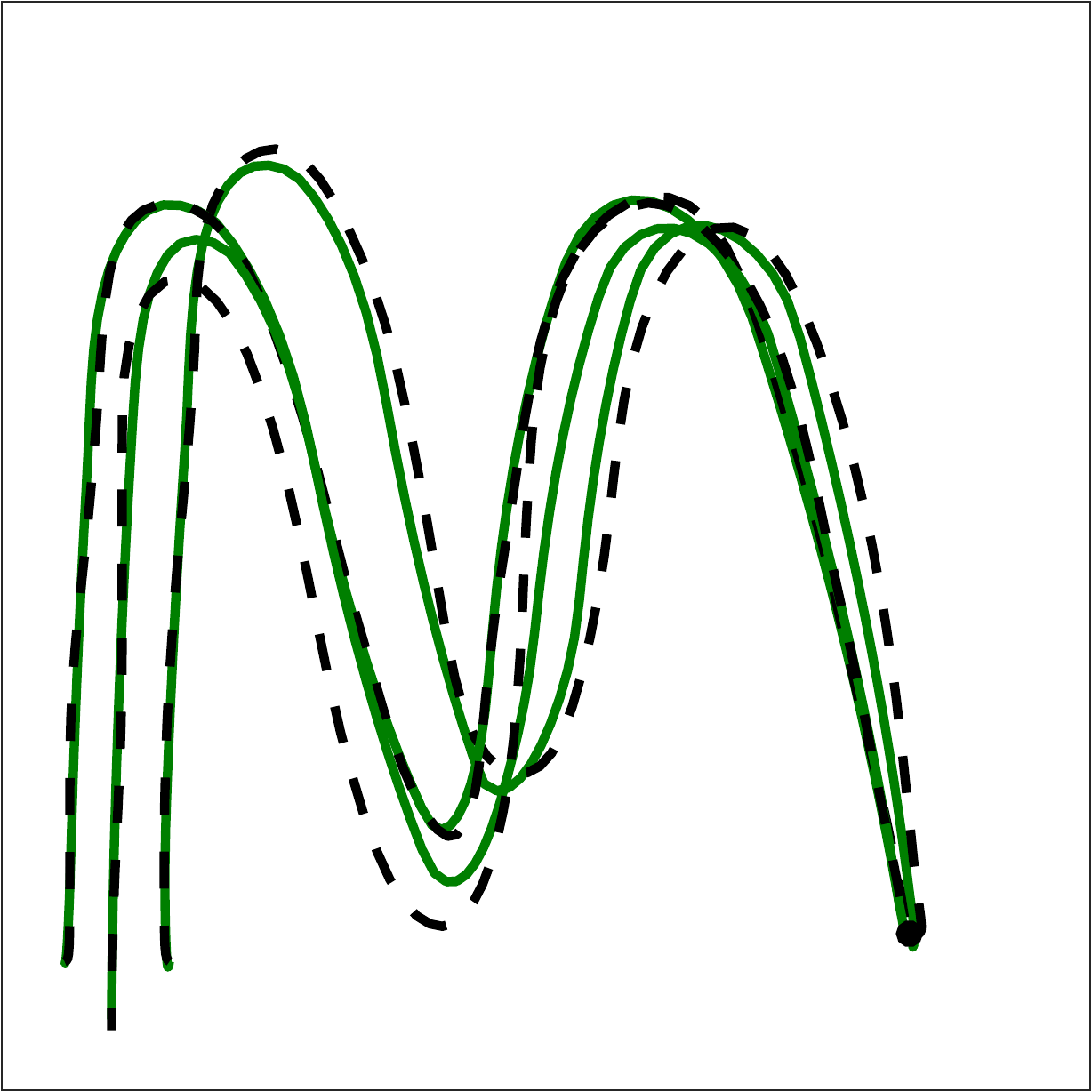}
    \hspace{-0.5mm}
    \includegraphics[width=\motionClassSize\textwidth]{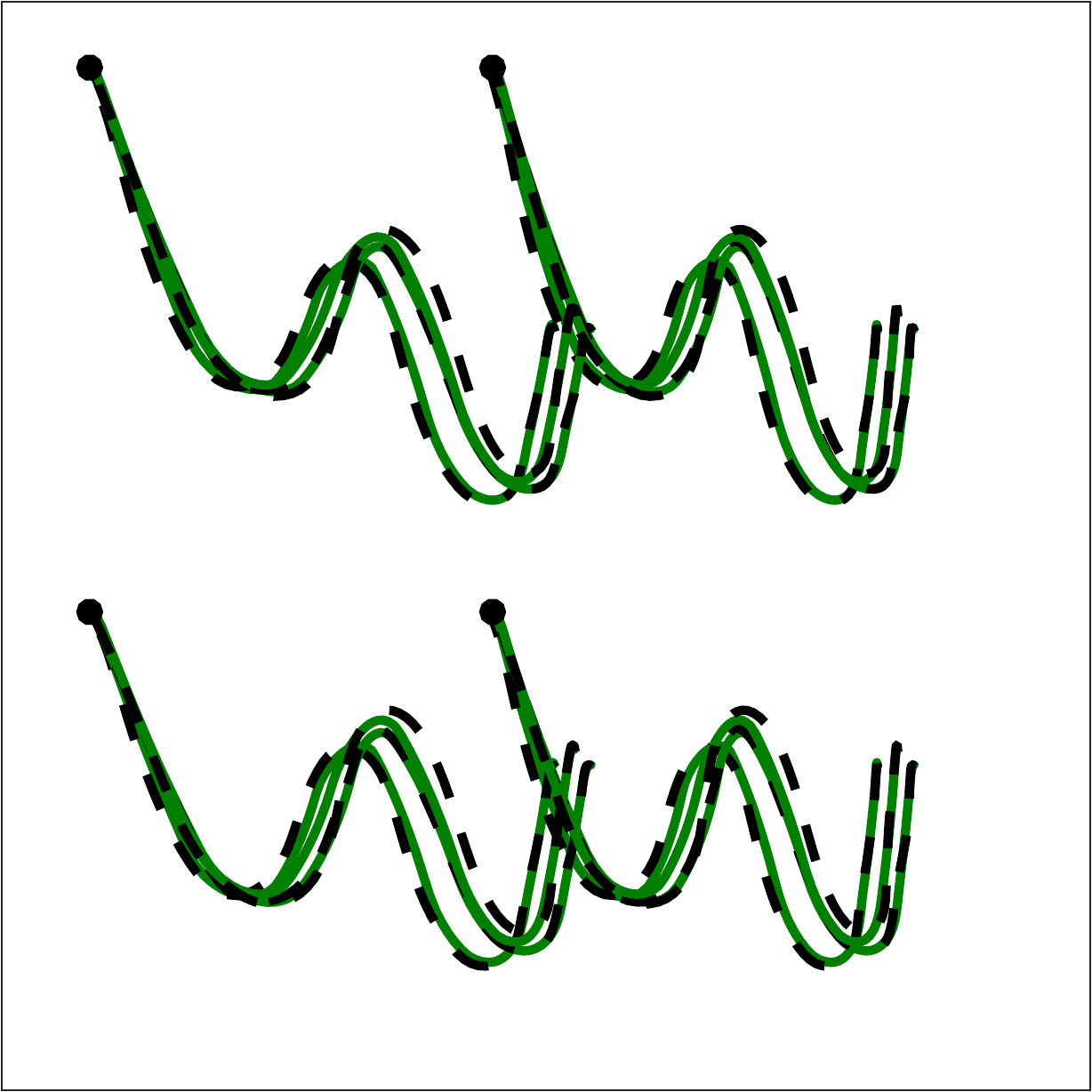}
    \hfill
    \includegraphics[width=\motionClassSize\textwidth]{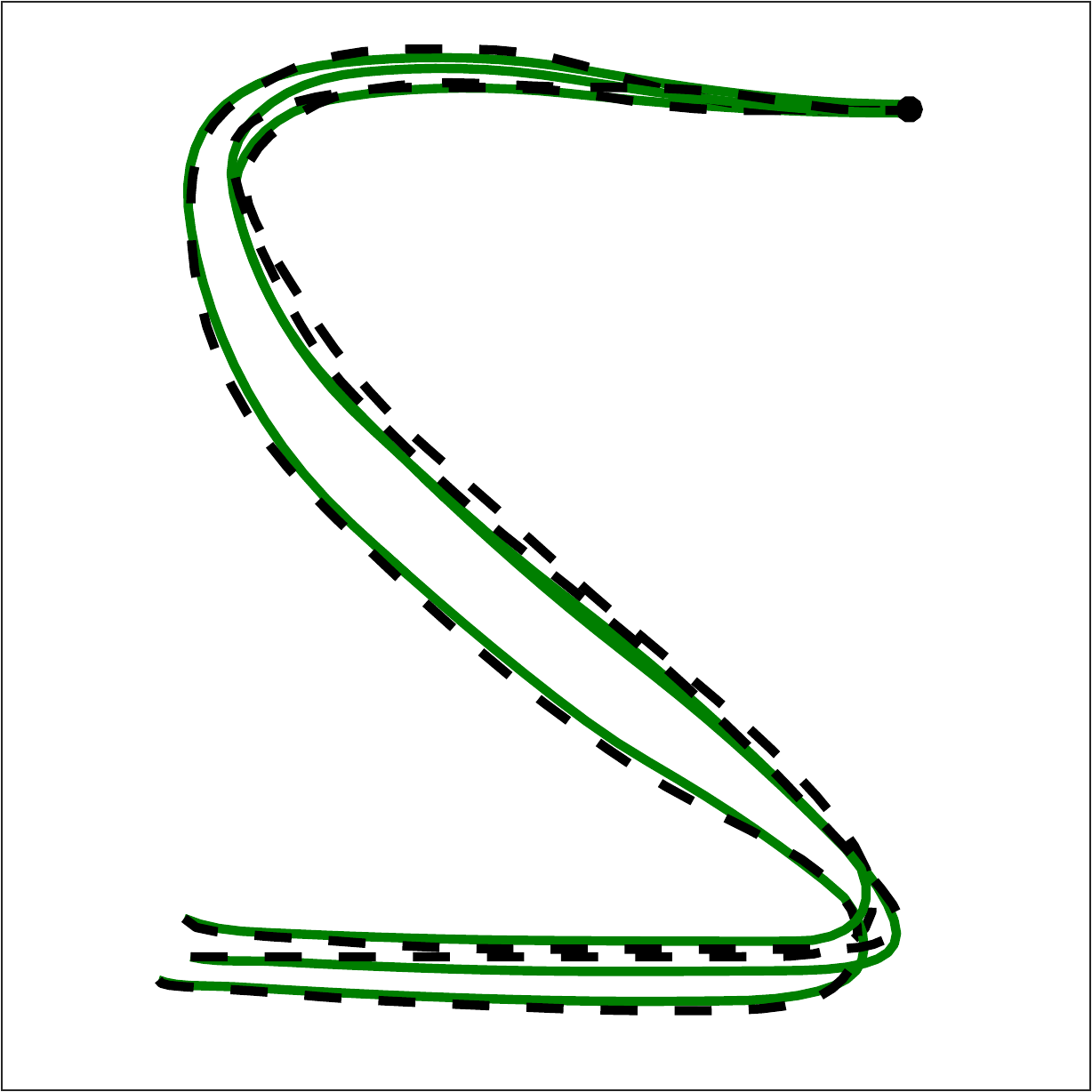}
    \hspace{-0.5mm}
    \includegraphics[width=\motionClassSize\textwidth]{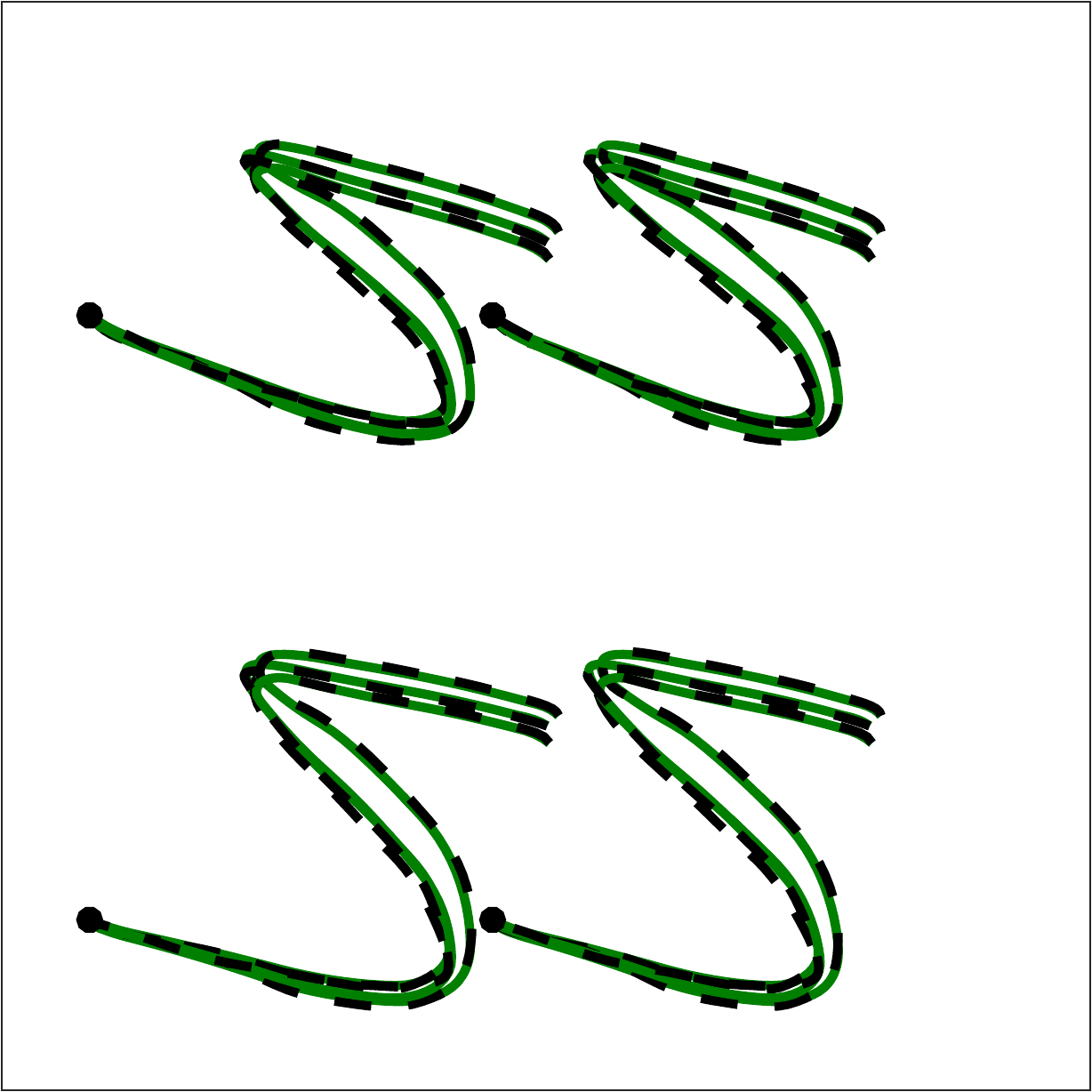}
    \hspace{5mm}~
    \\%
    \vspace{2mm}%
    \hspace{5mm}
    \includegraphics[width=\motionClassSize\textwidth]{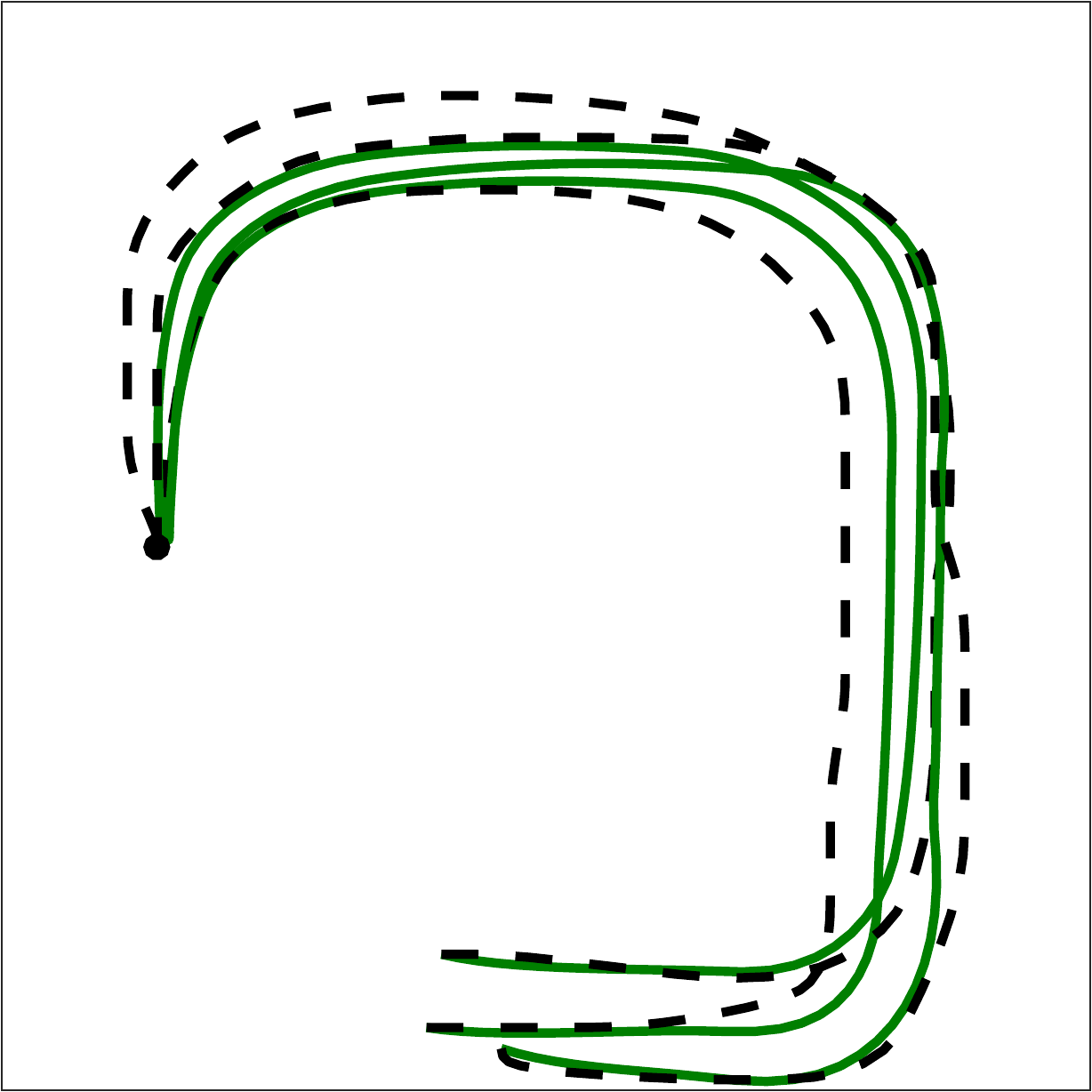}
    \hspace{-0.5mm}
    \includegraphics[width=\motionClassSize\textwidth]{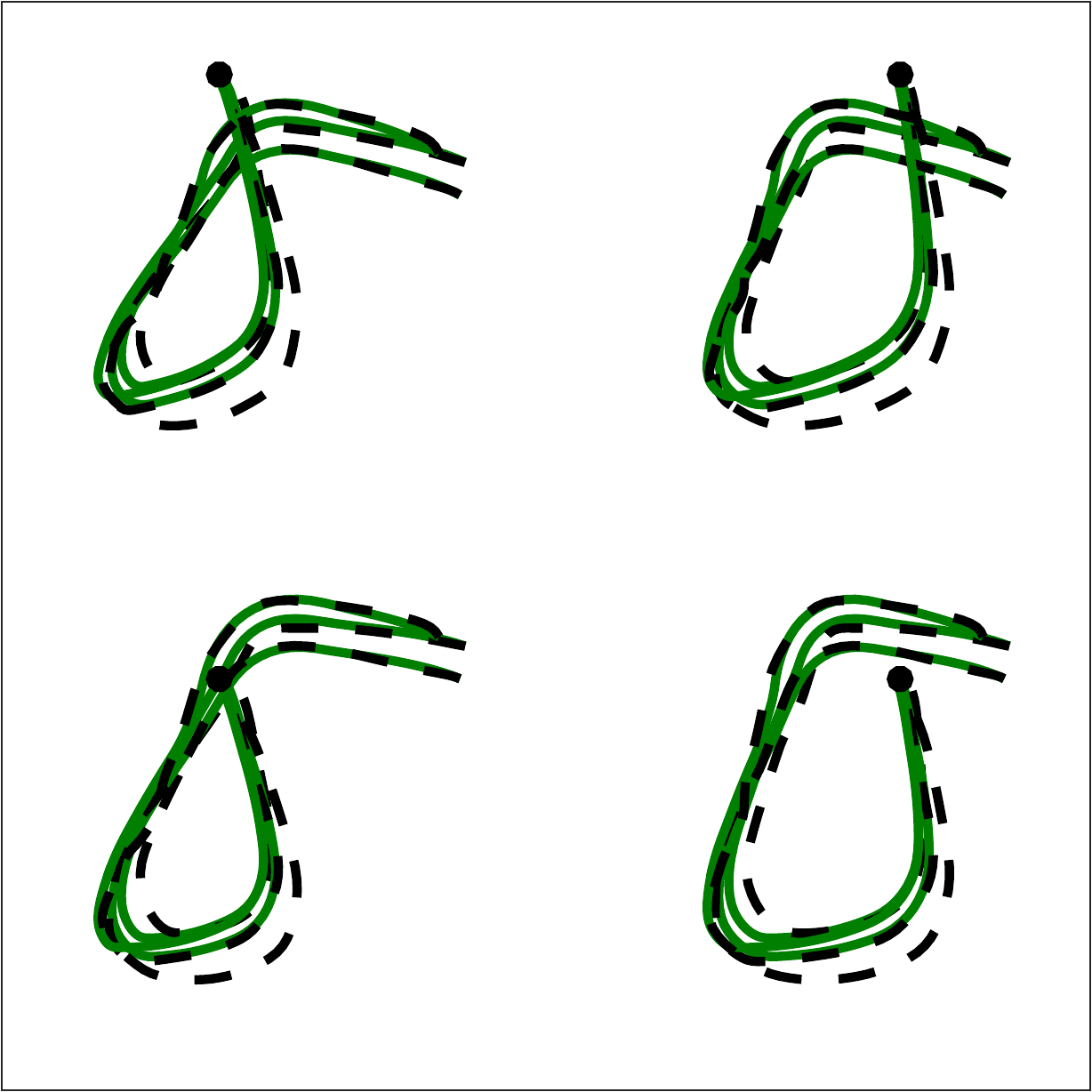}
    \hfill
    \includegraphics[width=\motionClassSize\textwidth]{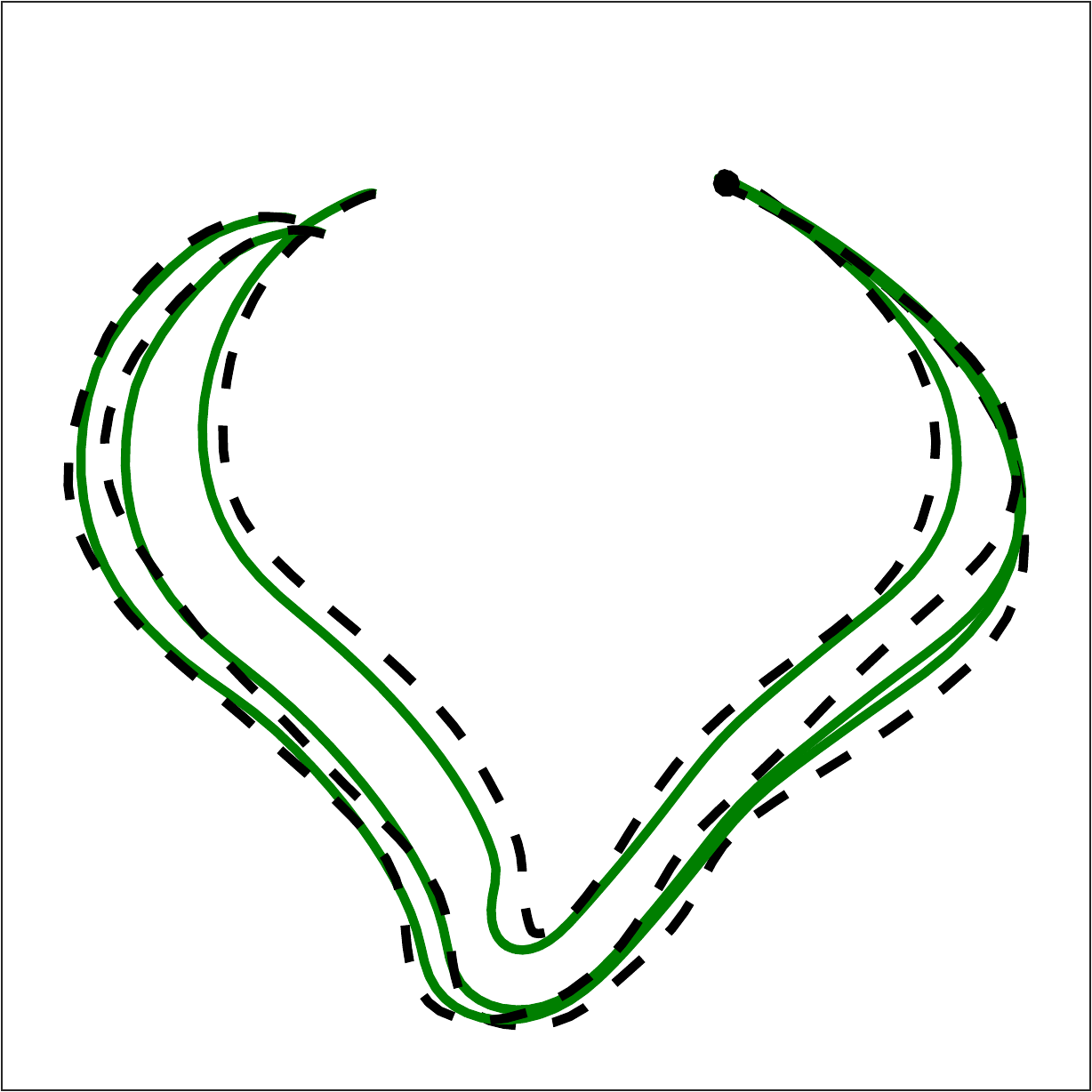}
    \hspace{-0.5mm}
    \includegraphics[width=\motionClassSize\textwidth]{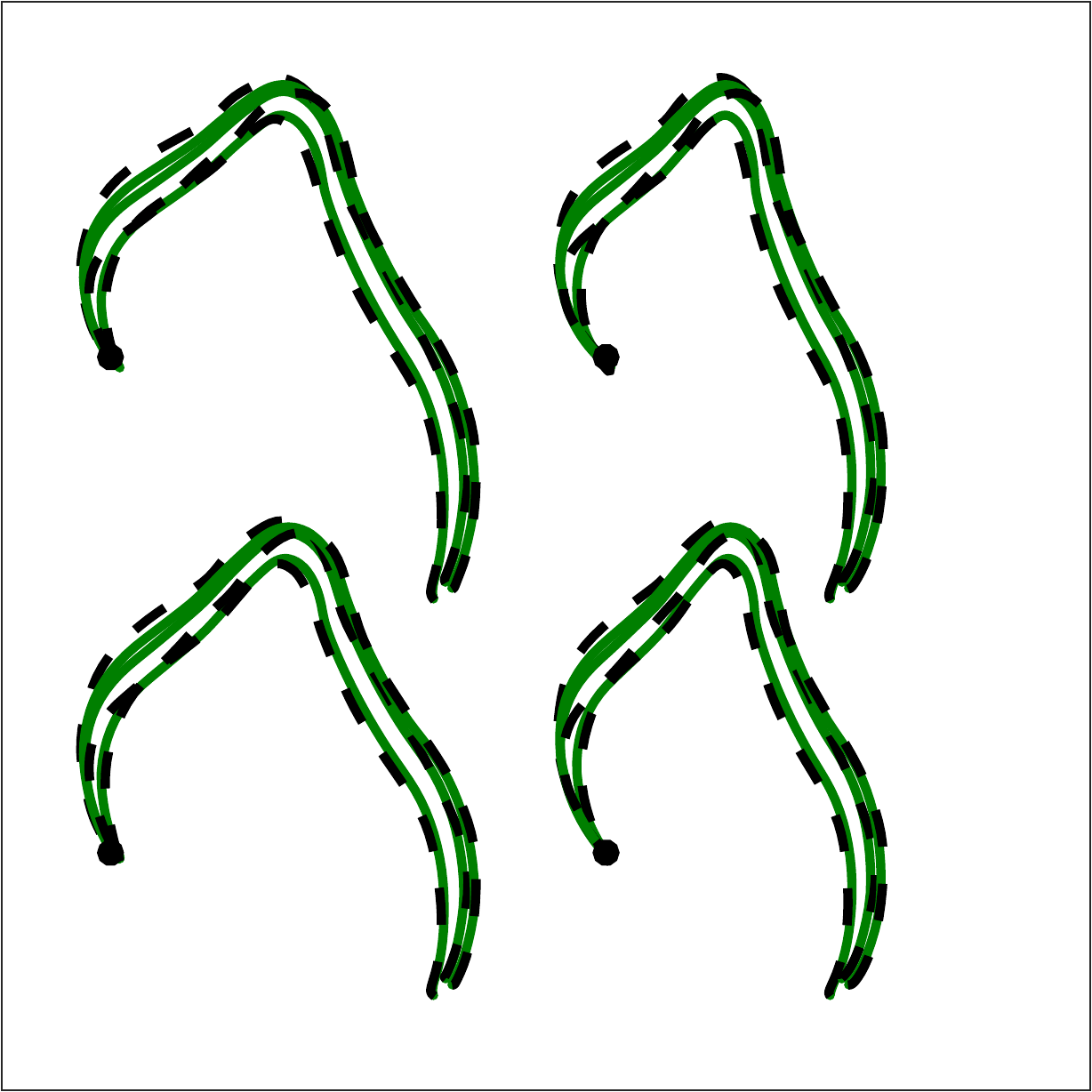}
    \hfill
    \includegraphics[width=\motionClassSize\textwidth]{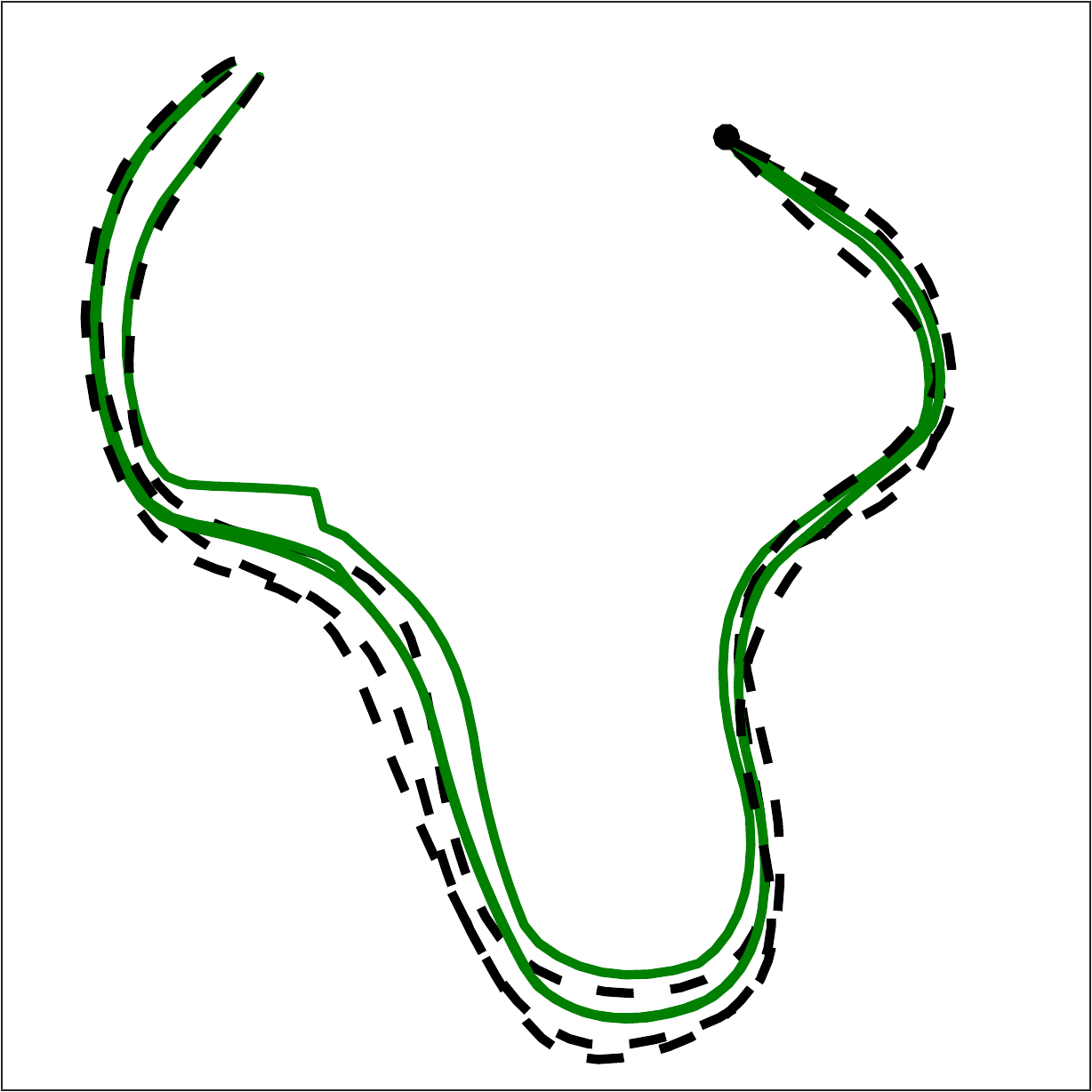}
    \hspace{-0.5mm}
    \includegraphics[width=\motionClassSize\textwidth]{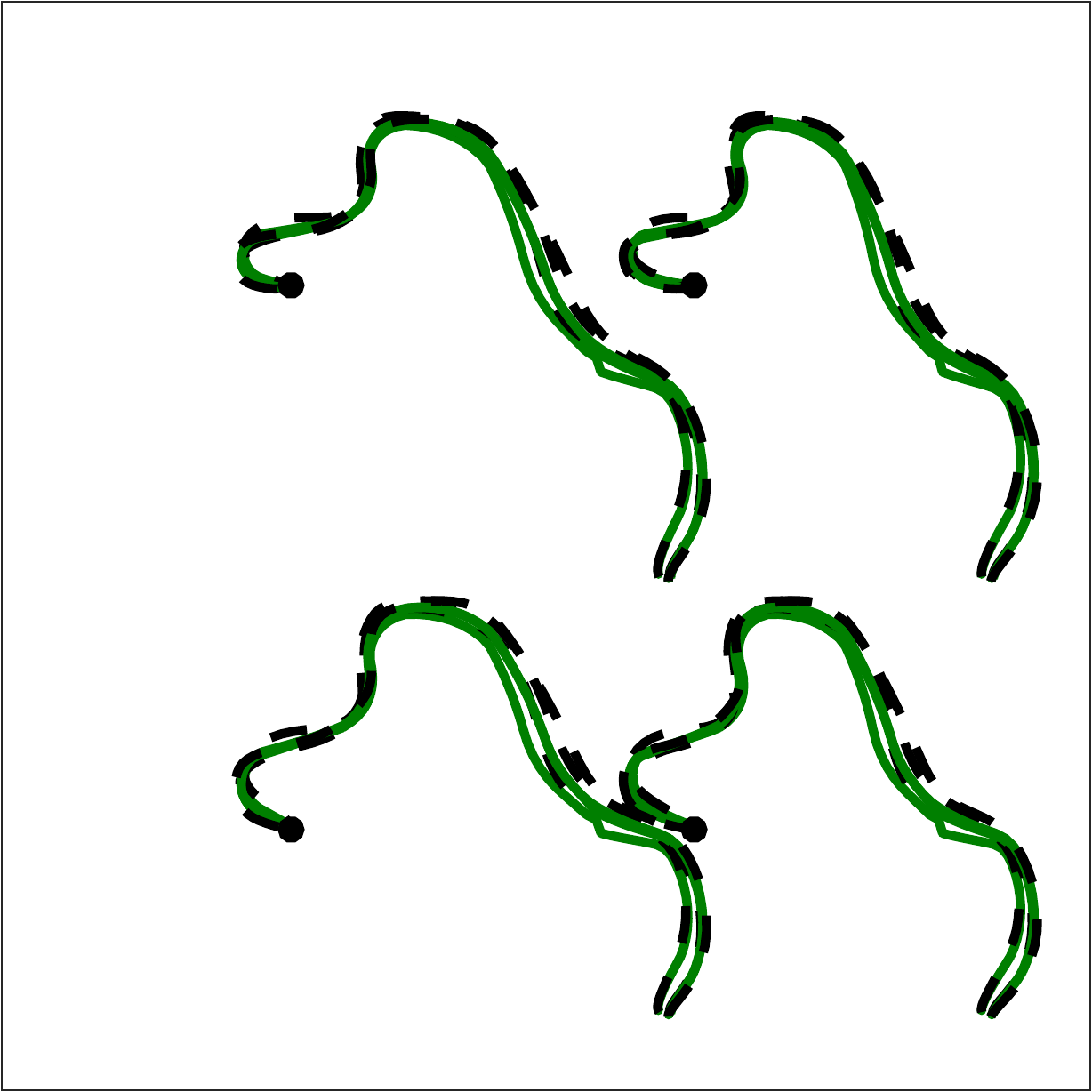}
    \hfill
    \includegraphics[width=\motionClassSize\textwidth]{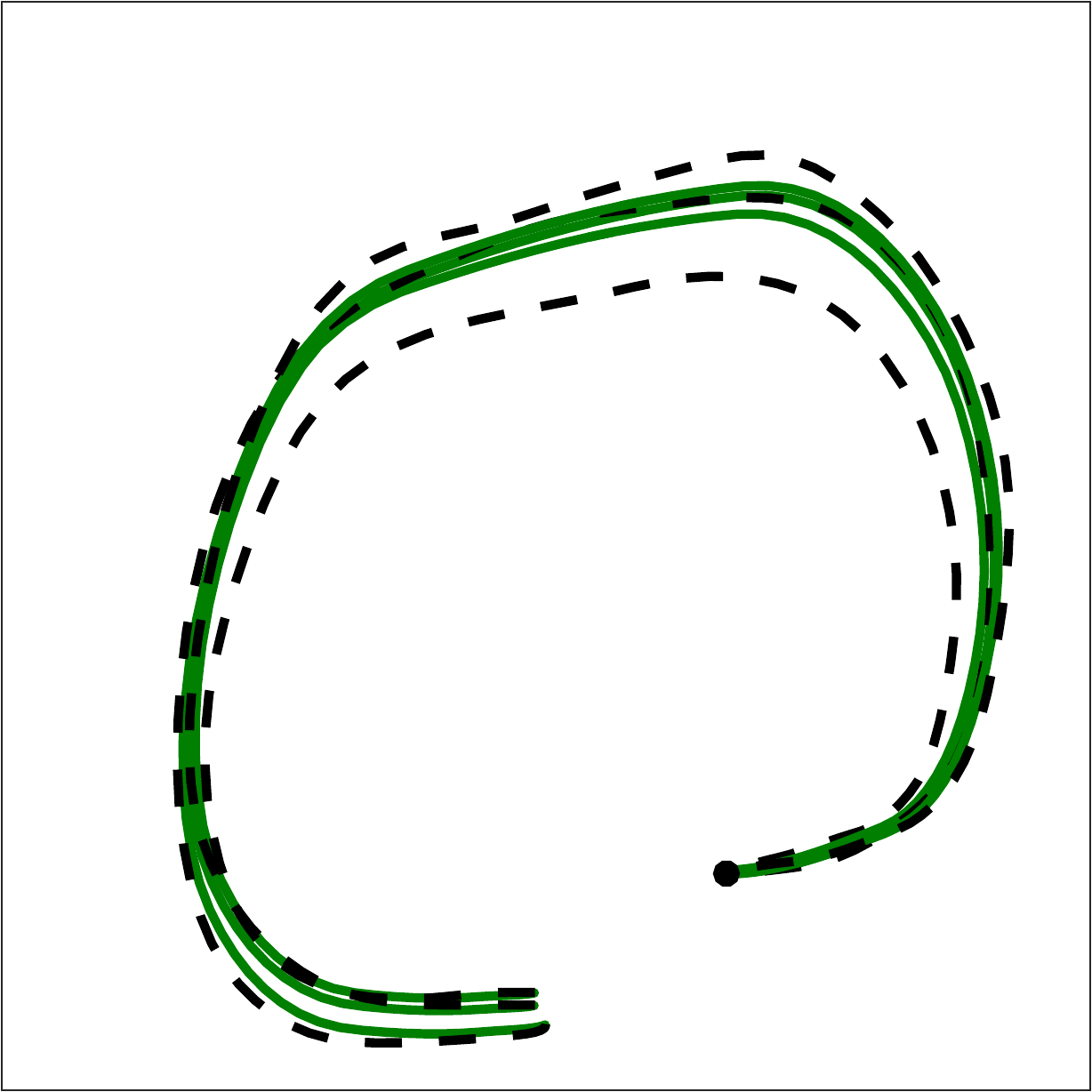}
    \hspace{-0.5mm}
    \includegraphics[width=\motionClassSize\textwidth]{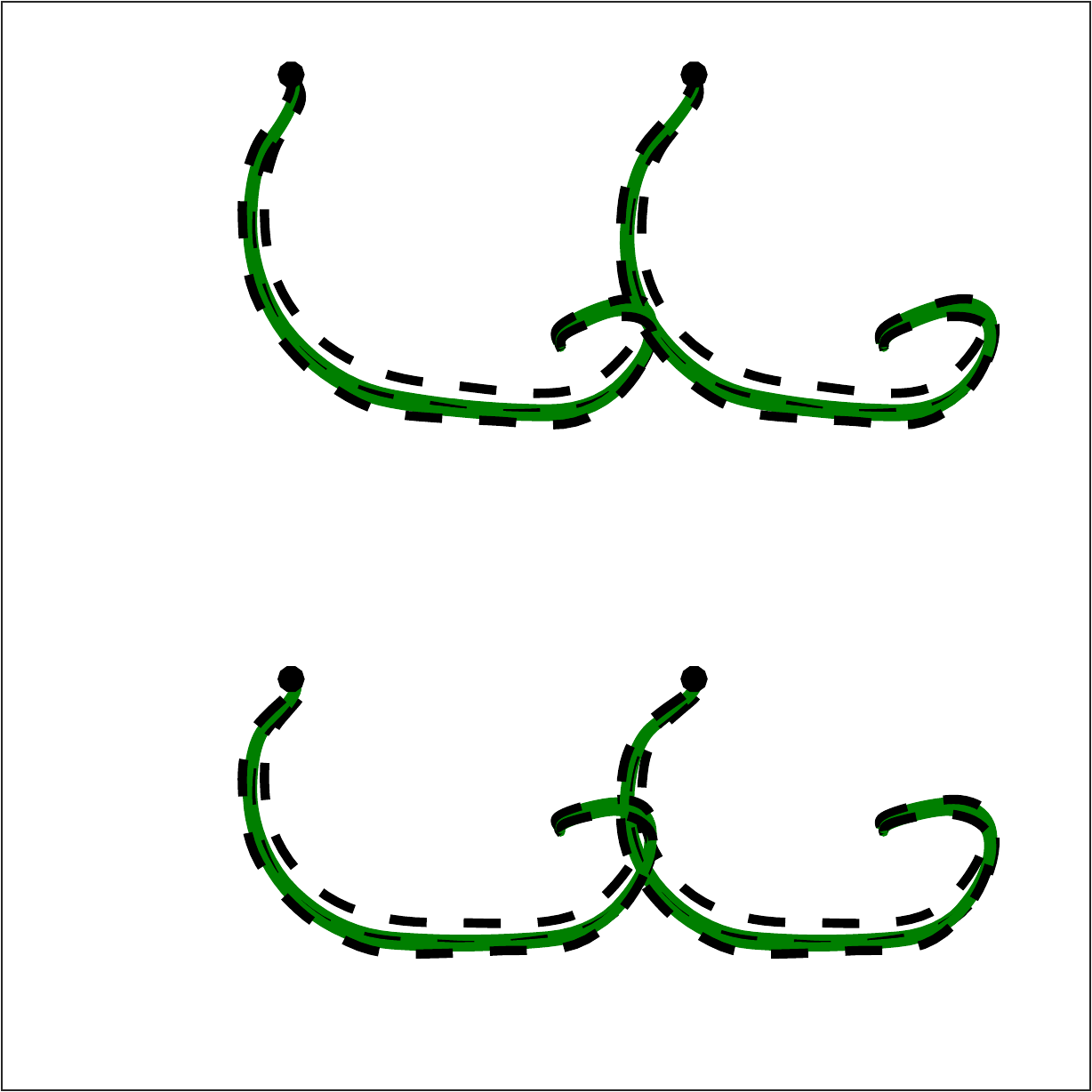}
    \hfill
    \includegraphics[width=\motionClassSize\textwidth]{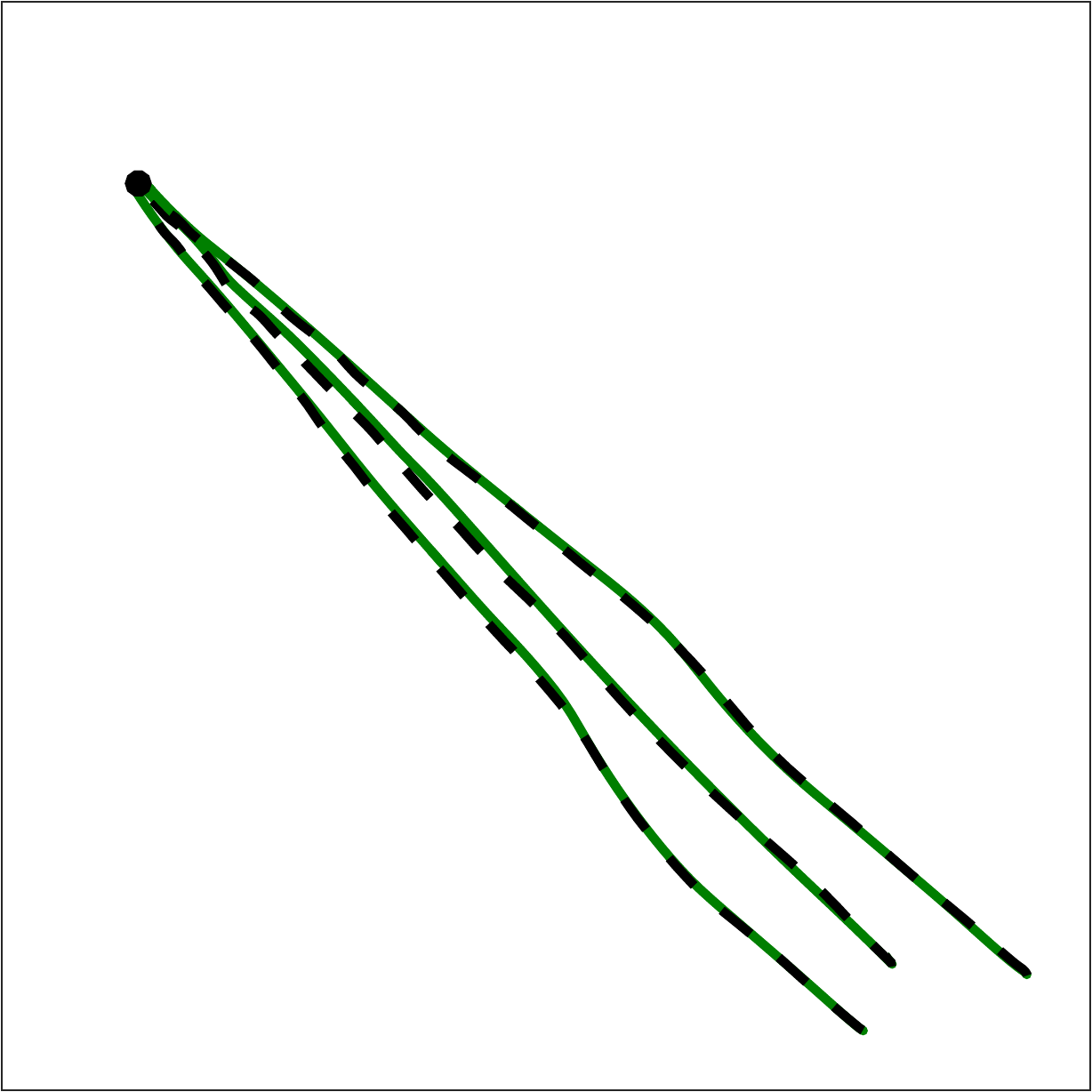}
    \hspace{-0.5mm}
    \includegraphics[width=\motionClassSize\textwidth]{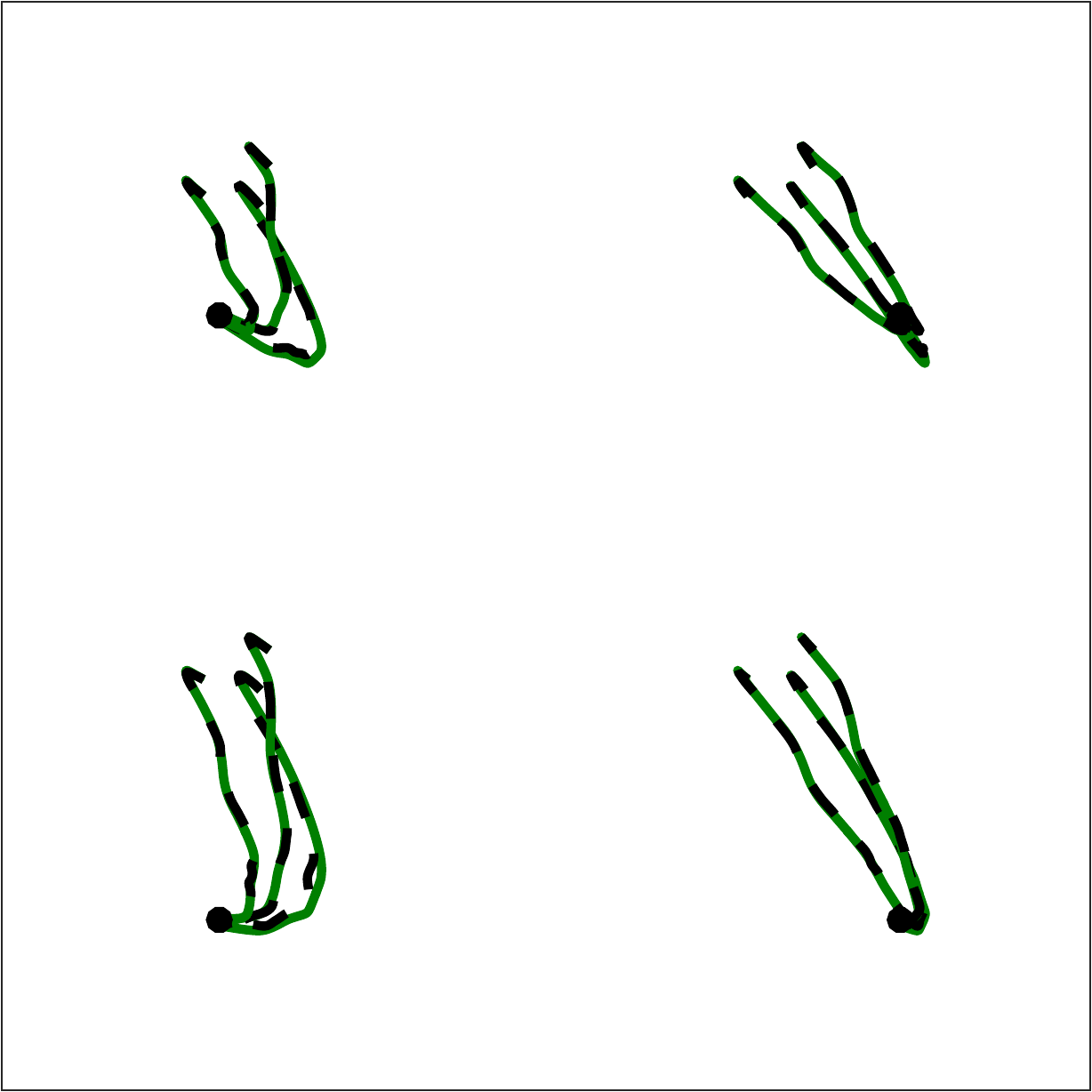}
    \hspace{5mm} %
    \caption{Motion classes contained in the augmented LASA handwriting dataset. Each sample is composed of camera trajectories in the Cartesian space (whose $x$-$y$ view is shown on the left side for each class) and the corresponding visual features trajectories (on a portion of the image plane, right side).}%
     \label{fig:dataset_all}%
\end{figure*}

This section presents the results carried out with the \ac{ds} methods described in Sec.~\ref{sec:approach} to achieve complex VS tasks.
We considered the case of tracking $4$ point features, to be solved along with other tasks or constraints (such as collision avoidance). 
To this end, we leveraged a set of demonstrations that moved the camera on complex trajectories, realizing the full desired behavior.
We first performed a simulation analysis to evaluate and compare the DS-based approaches in imitating the demonstrations while preserving the stability performance of the original VS. 
Experimental results show the effectiveness of the methods in achieving the VS task with collision avoidance, using a real robot manipulator.
For both simulations and experiments, the camera was controlled only in positions to ease the learning process.
We set $T = 0.03\,$s, being $30\,$Hz the camera nominal frame-rate.

\subsection{Dataset}\label{sec:dataset}

%

For our validation, we considered the LASA handwriting dataset~\cite{SEDS, LASA_Dataset}, a widely used benchmark for the evaluation of \ac{ds} methods~\cite{SEDS, Clf, saveriano2018incremental, Perrin16}. 
The latest version 
contains $30$ classes of $2$D motions, each provided with $7$ demonstrations. 
The motion classes consist of planar trajectories (including position, velocity and acceleration profile) starting from different initial points and converging to the same goal. 
Among the $30$ classes, $26$ are individual motions 
and $4$ are multi-modal, obtained combining the individual ones, neglected in our study as we do not focus on multi-modality.

However, to test the methods with the use of a visual feedback, we needed to augment the LASA dataset with corresponding visual features trajectories.
To this end, the original LASA trajectories have been used as the planar $x$-$y$  motion of our simulated camera, while a simple linear trajectory was added as motion along the $z$-axis, i.e. the forward direction, see Fig.~\ref{fig:camera_and_traj}.
The orientation, instead, has been kept constant.
Furthermore, we assumed a known fixed visual pattern of four points at a given pose in the Cartesian space and simulated a free floating camera 
using the Matlab Machine Vision Toolbox~\cite{corke:book:2011}.
Thus, for each camera pose, we were able to project the points on the $u$-$v$ image plane and, thus, construct the $26$ corresponding visual features trajectories, as schematically depicted in Fig.~\ref{fig:camera_and_traj}.
One single motion sample (with the reproductions performed by the three methods, as described in Sec.~\ref{sec:simulations}) 
is presented in Fig.~\ref{fig:dataset_one}; the remaining dataset samples are shown in Fig.~\ref{fig:dataset_all}.

%
%
In these plots, and in the following, the demonstrations are shown with black dashed lines and 
black dots refer to the final configurations; green solid and dotted lines are the reproductions starting from the same start of the demonstrations and from unseen initial points, respectively.
%

\subsection{Simulations}\label{sec:simulations}
%
%
%
%
For our simulation study, we considered only $3$ demonstrations (out of the $7$ available) for each motion class in the dataset described in Sec.~\ref{sec:dataset}, as done in~\cite{saveriano2018incremental}.
Furthermore, we down-sampled the trajectories to contain exactly $100$ points, to significantly reduce the training time and ease the testing procedures. 
We kept the \ac{rds} reshaping term active along all the task execution, i.e. we set $h=1$ in~\eqref{eq:reshaped_control}. 
%
The number of hyper-parameters $k$ used in the learning strategies was obtained with a grid search.
In the case of \ac{clfdm} and \ac{rds}, $k$ is the number of Gaussian components used to encode $\bm{f}$ and $V$, and $\bm{u}_{\text{RDS}}$ respectively; 
in \ac{fdm} $k$ is the number of kernel functions composing the diffemorphism.

%
%

The performance in reproducing the demonstrations 
was evaluated w.r.t. 
training time $\tau$ and accuracy,
measured as the mean 
over the dataset of the root mean square (RMS) error of the camera position ($p_\text{RMS}$), camera linear velocity ($v_\text{RMS}$) and  visual features ($s_\text{RMS}$). 
%
%
%
The methods were evaluated with different $k$; 
\begin{table}[!b]
    \setlength{\tabcolsep}{3.5pt}
    \centering
    \caption{Analysis of the different DS approaches applied to VS.}
    \label{tab:comparison_lasa}
 	\begin{tabular}{cccccc}
 	\toprule
 	 \multirow{2}{*}{\sc{Approach}}  & \multirow{2}{*}{$k$} & $p_\text{RMS}$ & $v_\text{RMS}$ & $s_\text{RMS}$ & $\tau$ \\
 	 & &  [mm] & [mm/s] & [pixels] & [ms] \\
 	\midrule
 	\ac{rds}  & $7$ &  $25 \pm 39$  &  $105 \pm 72$  & $9\pm 15$  & $546 \pm 1431$  \\
 	\ac{rds}  & $11$ &  $17 \pm 13$  &  $86 \pm 42$ & $6\pm 5$ & $697\pm 1540$ \\
 	\ac{clfdm}  & $7+2{}^\dagger$ &  $36 \pm 55$  &  $121 \pm 102$  & $14\pm 21$  & $9298 \pm 4026$  \\
 	\ac{clfdm}  & $11 + 2{}^\dagger$ &  $19 \pm 14$  &  $82 \pm 52$ & $7\pm 5$ & $9291\pm 3886$ \\
 	 FDM  & $150$ &  $60  \pm 11$  &  $118 \pm 45$ & $14\pm 6$  & $173 \pm 376 $\\
 	 FDM & $50$ &  $59 \pm 11$  &  $120 \pm 45$ & $14\pm 6$  & $53 \pm 105 $\\
 	\midrule
	\multicolumn{6}{l}{\scriptsize $^\dagger$For \ac{clfdm}, the number of hyper-parameters is reported considering}\\ [-1.5pt]
	\multicolumn{6}{l}{\scriptsize \phantom{$^\dagger$}both the $\bm{f}$ and $\bm{u}_\text{CLF}$ components.}\\
	\bottomrule
\end{tabular}
\end{table}
Table~\ref{tab:comparison_lasa} reports the most and less accurate models.
%
%
The performance of each method can be qualitatively evaluated in Fig.~\ref{fig:dataset_one} for one of the $26$ motions of the dataset; 
Fig.~\ref{fig:dataset_all} shows the \ac{rds} performance 
on the other 
$25$ motions.
\ac{rds} 
managed to well reproduce the demonstrations, as presented by the green traces in the plots. %
The quantitative analysis of the three methods reproducing 
the demonstrations starting from the same initial points is reported in Table~\ref{tab:comparison_lasa}. 
As shown, \ac{rds} and \ac{clfdm} have a comparable accuracy 
with the best set of hyper-parameter. \ac{rds} learns significantly faster as it only fits a \ac{gmm}, while \ac{clfdm} fits a \ac{gmm} and then learns 
a CLF by solving a non-convex optimization problem. The drawback of \ac{rds} is that convergence is conditional to the clock signal $h$, which might 
generate overshoots or spurious attractors. It is worth mentioning that we set $h=1$ in this simulation study and all the trajectories converged, 
since they start within the demonstration area.
%
Fig.~\ref{fig:dataset_one} (top) 
shows generalization results where the \ac{rds} reshaping term is deactivated after $100$ steps to ensure convergence. Among the tested approaches, \ac{fdm} is the less accurate but it learns faster. Indeed, \ac{fdm} does not handle multiple demonstrations and we simply averaged 
multiple demonstrations (see Sec.~\ref{sec:fdm}~and~\cite{Perrin16}).
As expected, this strategy resulted in a loss of accuracy, as also visible in Fig.~\ref{fig:dataset_one} (bottom), 
where \ac{fdm} reproductions are ``attracted'' by 
the mean of the trajectories. 
It is important to remark that the estimation of $\gamma(\bm{\varepsilon})$ may not be trivial, 
e.g. 
in the case where the robot starts far from the demonstrations.

Dataset and code for using RDS and CLF-DM in a simulated VS scheme are publicly available\footnote{\texttt{https://github.com/matteosaveriano/ilvs}}.

%
%
\subsection{Experiments}
\begin{figure}[!t]
    \newcommand{\mysnapwidht}{1}%
    \centering%
    %
    \includegraphics[width=\mysnapwidht\columnwidth]{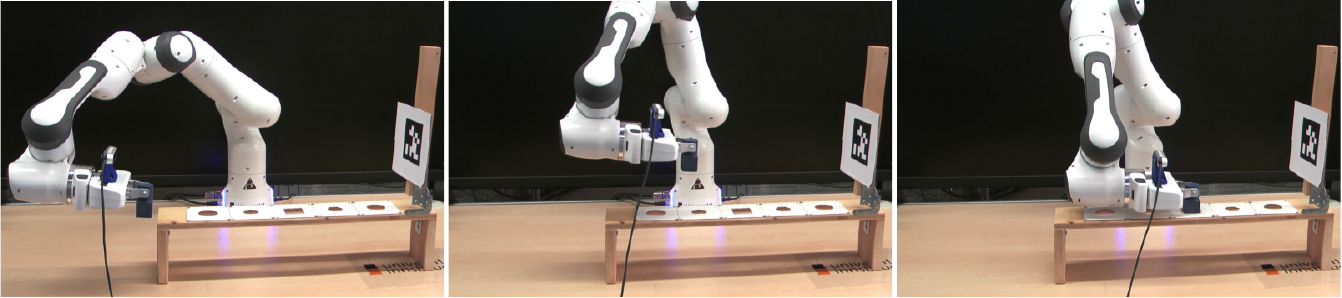}
    \\ [2pt]
    \includegraphics[width=\mysnapwidht\columnwidth]{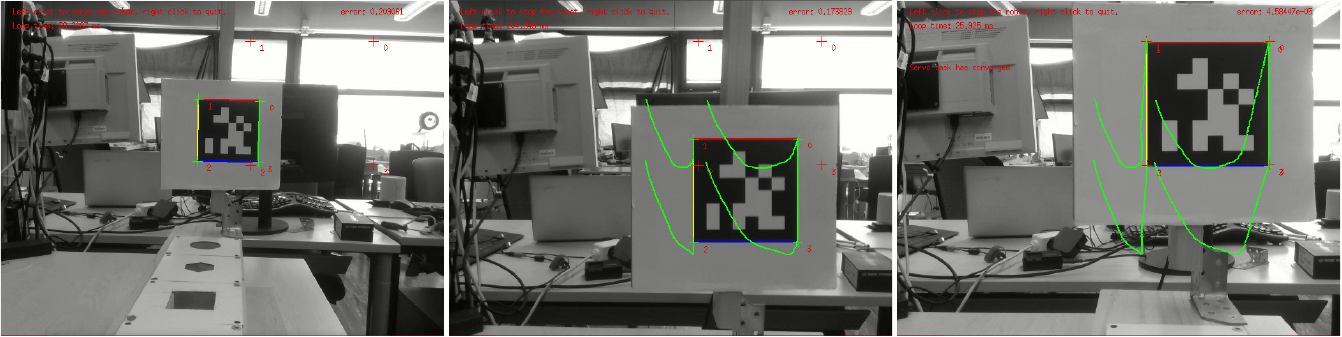}
    \caption{Panda robot performing the peg-in-hole task (top) while avoiding possible collisions using the proposed \ac{rds} approach, 
    and the corresponding images acquired by the on-board camera (bottom).
    }%
    \label{fig:peg_in_hole_snap}%
\end{figure}    
We considered a peg-in-hole task with 
the 7-axes robot manipulator by Franka Emika,
mounting an Intel RealSense D435i sensor 
at its end-effector (Fig.~\ref{fig:peg_in_hole_snap}). 
%
%
%
An AprilTag marker was placed on the support 
of the target hole. 
%
The robot had to move to the target, using vision, 
avoiding the support.
The $4$ AprilTag corners served as visual pattern used to compute the features, see Fig.~\ref{fig:peg_in_hole_snap}.
The image processing and the robot kinematic control, were managed by the ViSP library~\cite{Marchand:ram:2005}.

\begin{figure}[!t]
    \newcommand{\myexpplotwidth}{0.475}%
    \centering%
    %
    \includegraphics[trim={92 295 125 250},clip,width=\myexpplotwidth\columnwidth]{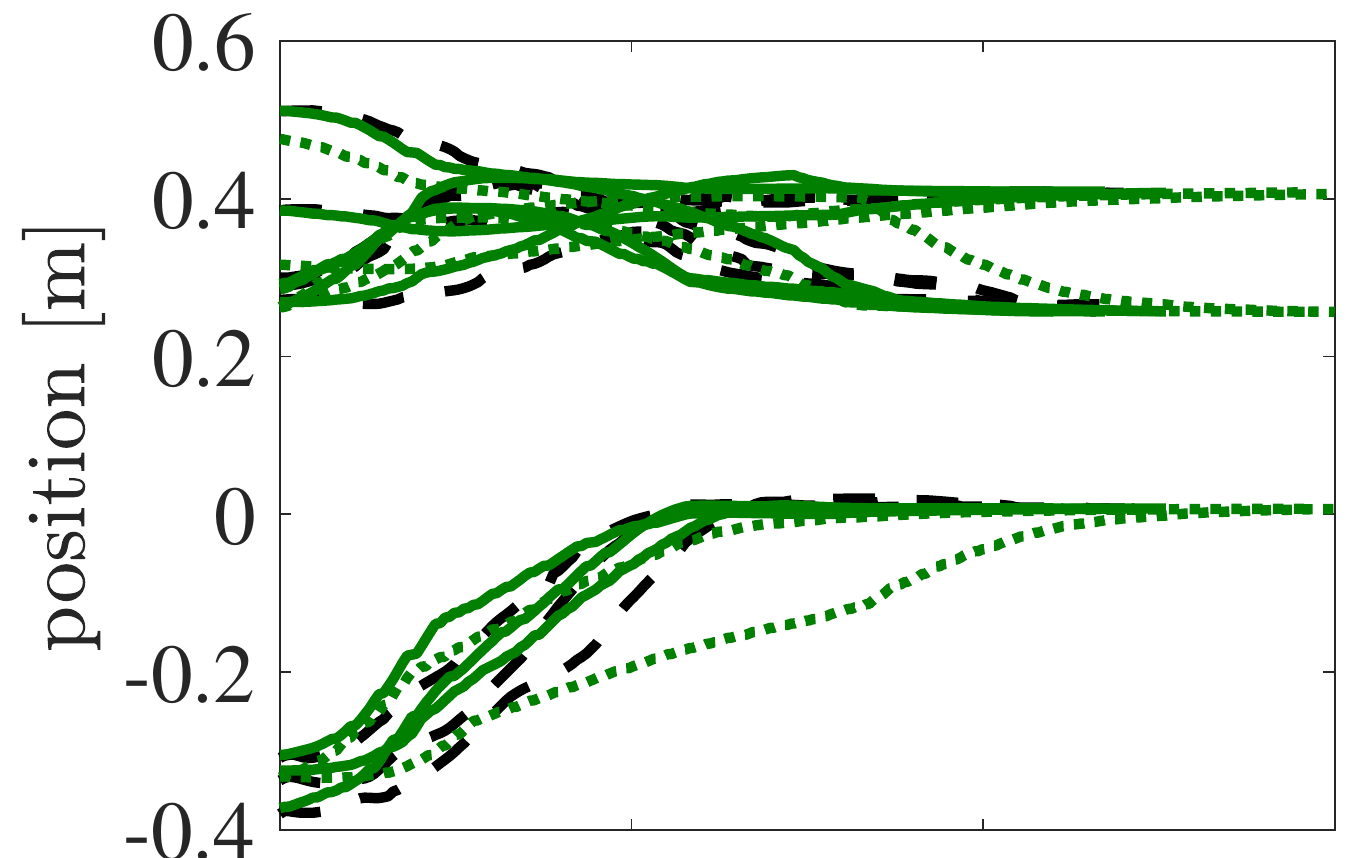}%
    \quad%
    \includegraphics[trim={92 295 125 250},clip,width=\myexpplotwidth\columnwidth]{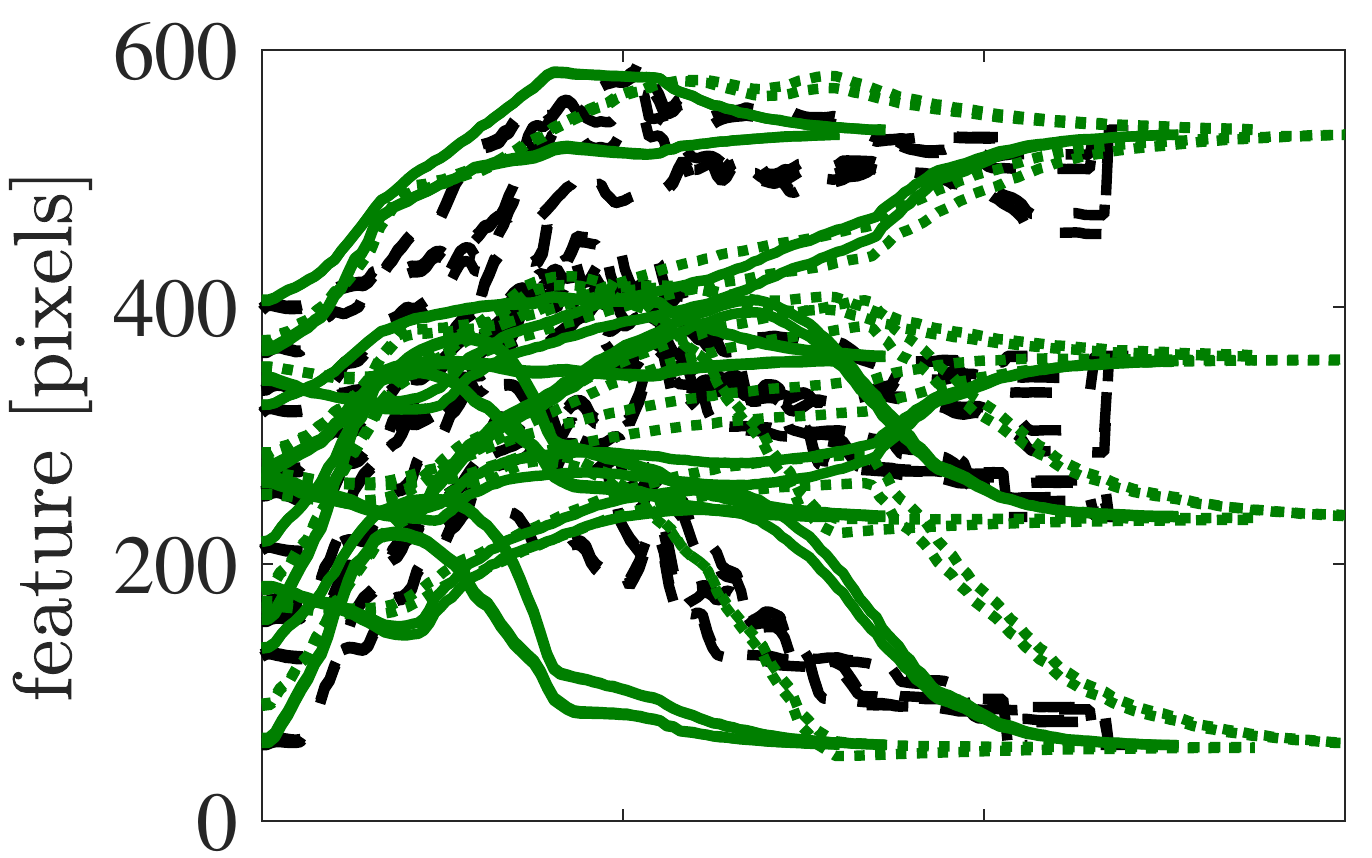}%
    \\[2pt]
    \includegraphics[trim={92 240 123 250},clip,width=\myexpplotwidth\columnwidth]{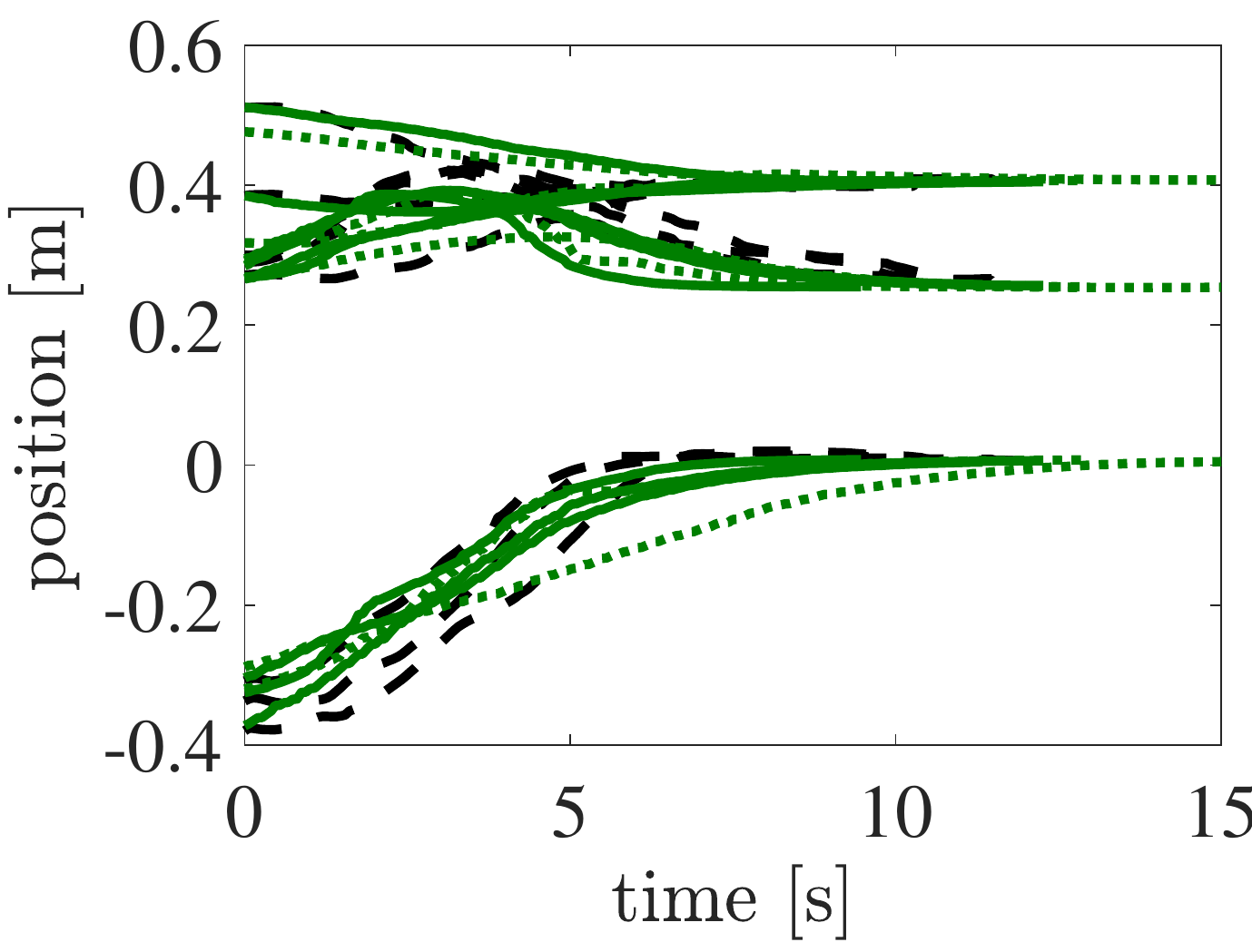}%
    \quad%
    \includegraphics[trim={92 240 125 250},clip,width=\myexpplotwidth\columnwidth]{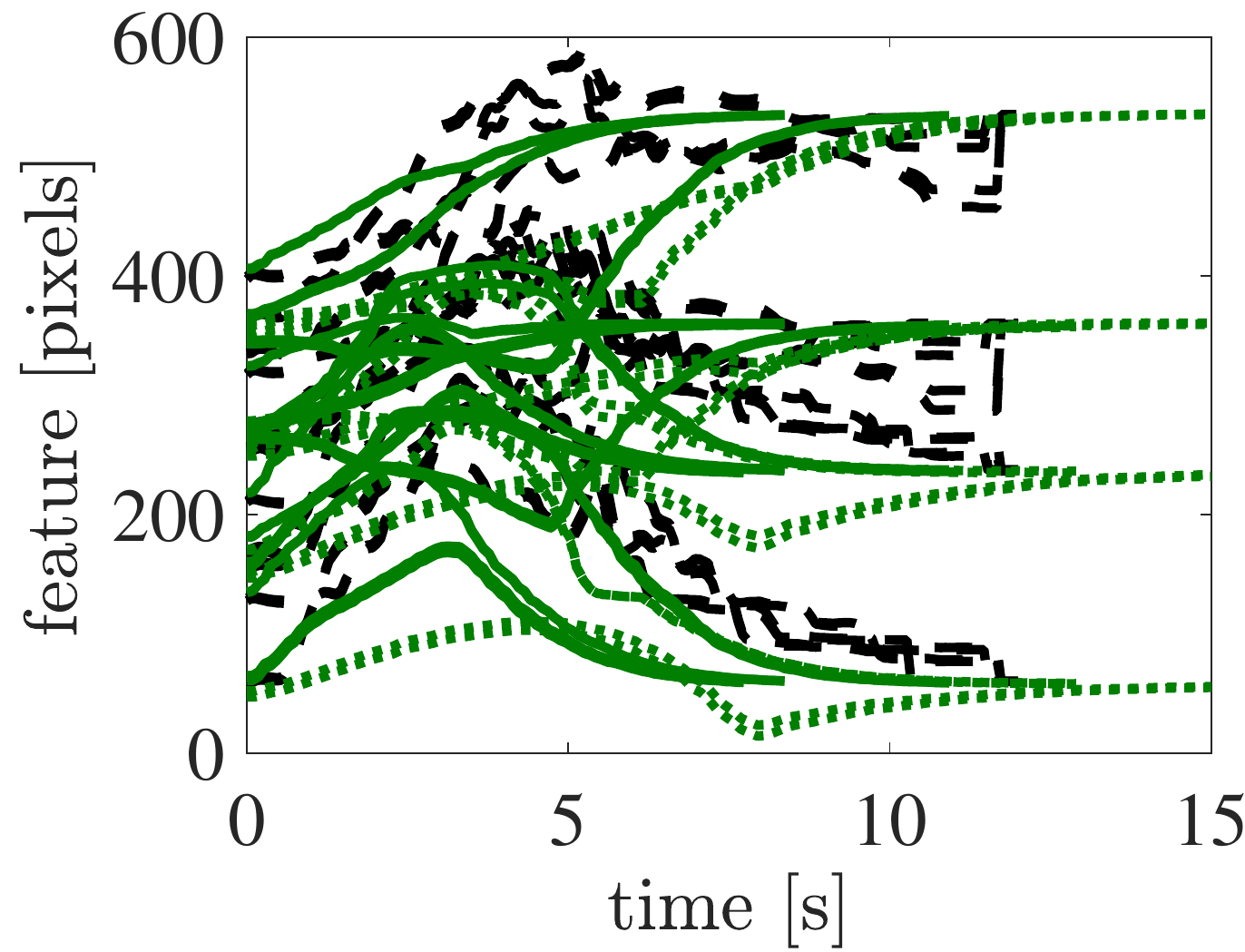}%
    %
    \caption{Peg-in-hole task: camera position (left) and features trajectories (right) performed by \ac{rds}~(top) and \ac{clfdm}~(bottom), shown in green; the demonstrations are represented with black dashed lines.}%
    \label{fig:peg_in_hole_plots}%
\end{figure}

Using the classical \ac{vs} law~\eqref{eq:vs_control} 
the robot hit the support and failed the task. 
%
Thus, we resorted to DS-based IL methods to perform the full desired behaviour, augmenting the basic VS with a collision avoidance skill. 
%
To this end, we selected the two most accurate methods of our simulation study, 
i.e. \ac{rds} and \ac{clfdm}. 
We provided $3$ demonstrations using kinesthetic teaching, showed as black dashed lines in Fig.~\ref{fig:peg_in_hole_plots}. 
The demonstrations were first pre-processed to exactly converge to the same goal in the feature space and then used to learn the \ac{rds} and \ac{clfdm} models. 
The hyper-parameters were selected using the best values in Table~\ref{tab:comparison_lasa}. 
The two approaches were used to generate closed-loop velocity commands for the robot camera, which were directly sent to the kinematic controller.
The results are shown in Fig.~\ref{fig:peg_in_hole_plots}. Both the trajectories starting from the demonstrated initial poses (solid green lines) and the ones generalized to different starts (dashed green lines) realized the desired peg-in-hole task, avoiding the collision with the support. 
%

\begin{figure}[!t]
    \newcommand{\myexpplotwidth}{0.46}%
    \centering%
    \includegraphics[trim={92 295 132 255},clip,width=\myexpplotwidth\columnwidth]{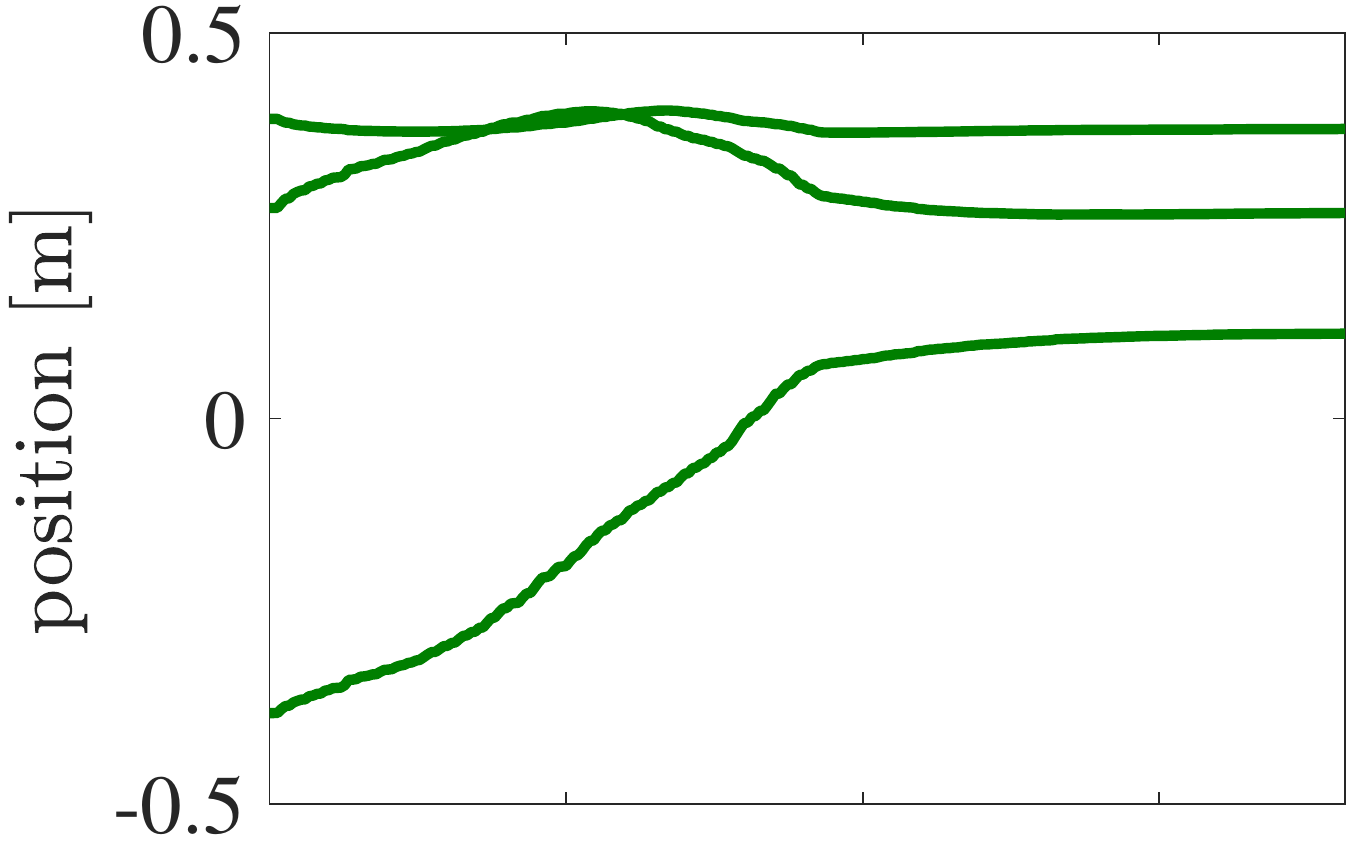}%
    \quad%
    \includegraphics[trim={92 295 132 255},clip,width=\myexpplotwidth\columnwidth]{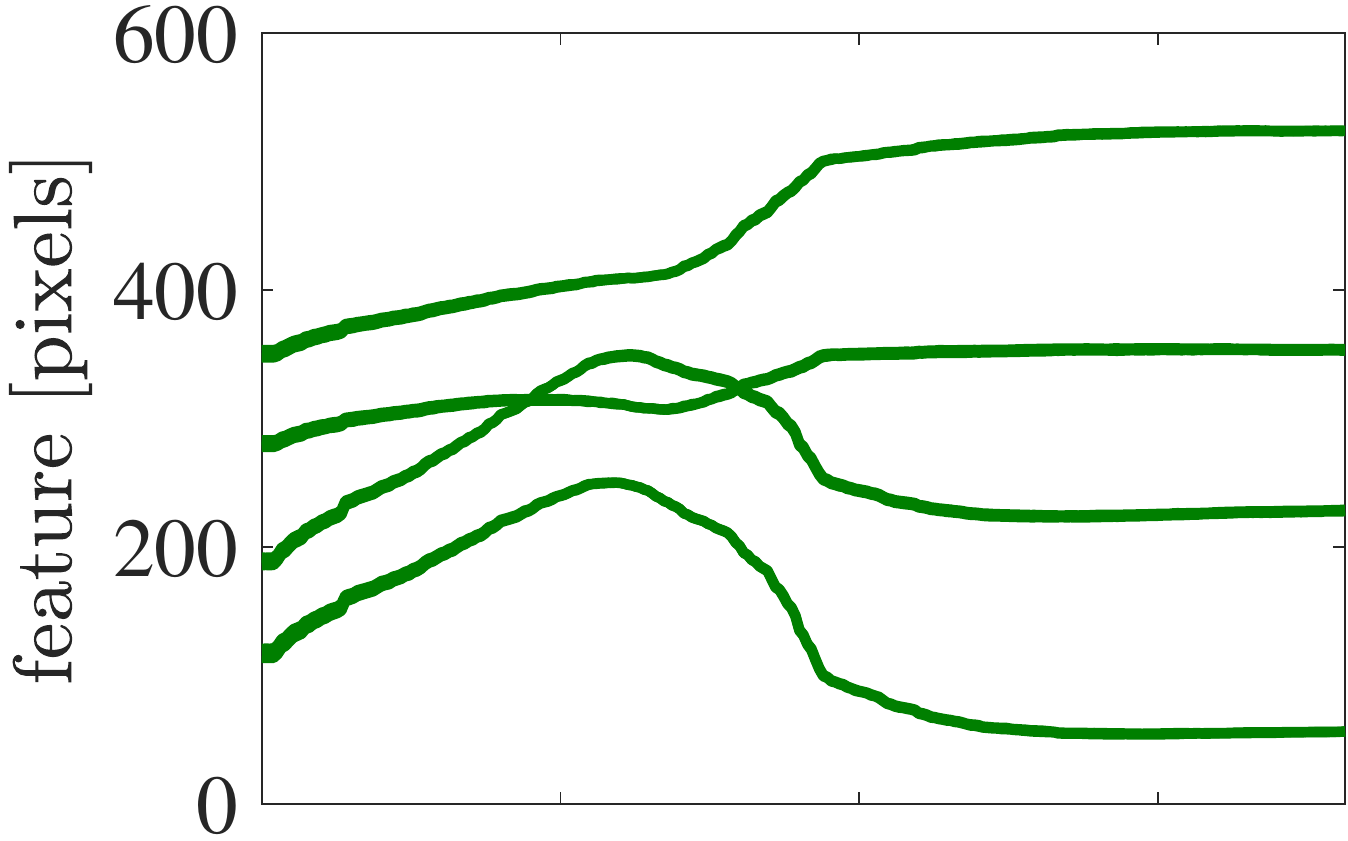}%
    \\[2pt]
    \includegraphics[trim={92 240 132 255},clip,width=\myexpplotwidth\columnwidth]{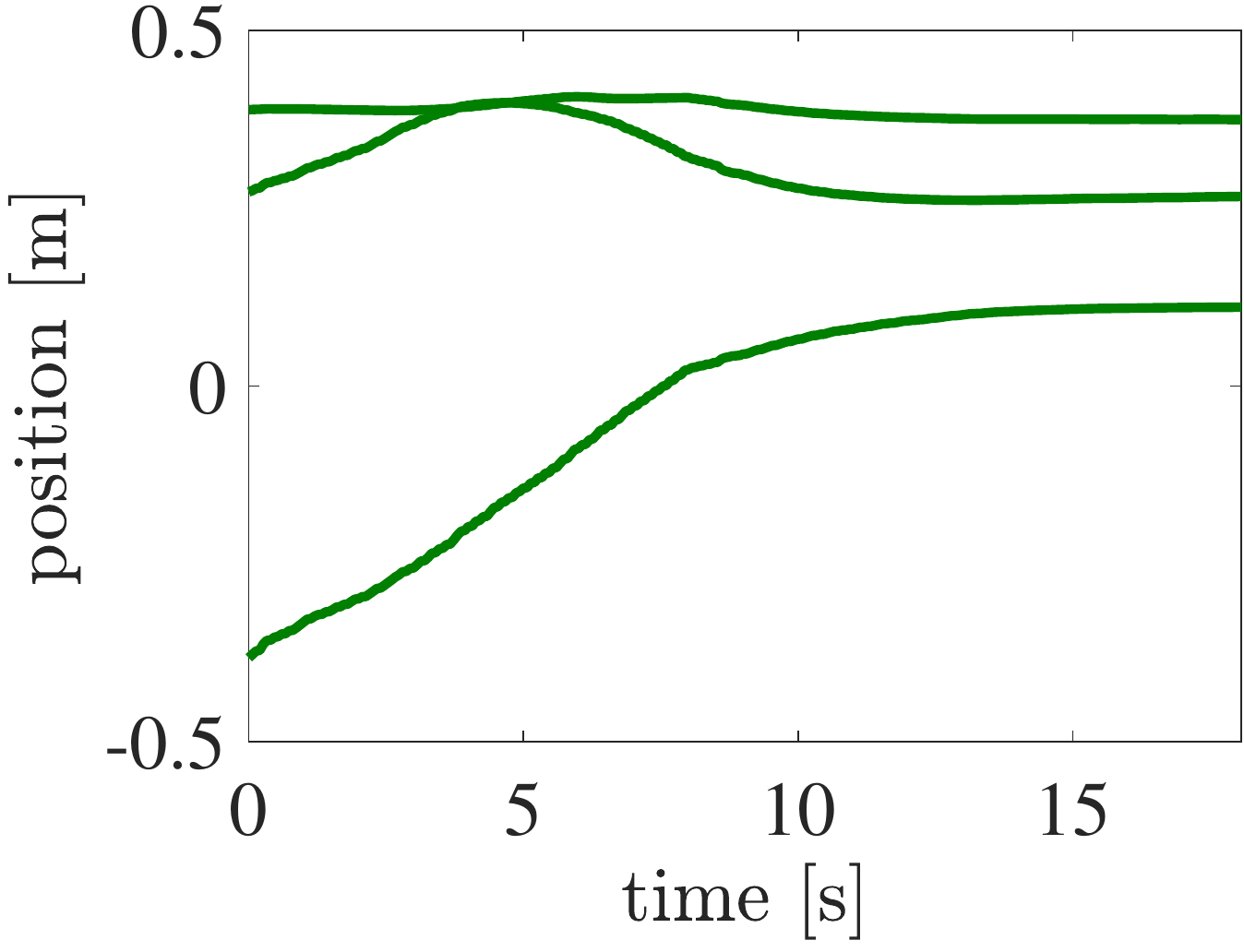}%
    \quad%
    \includegraphics[trim={92 240 132 255},clip,width=\myexpplotwidth\columnwidth]{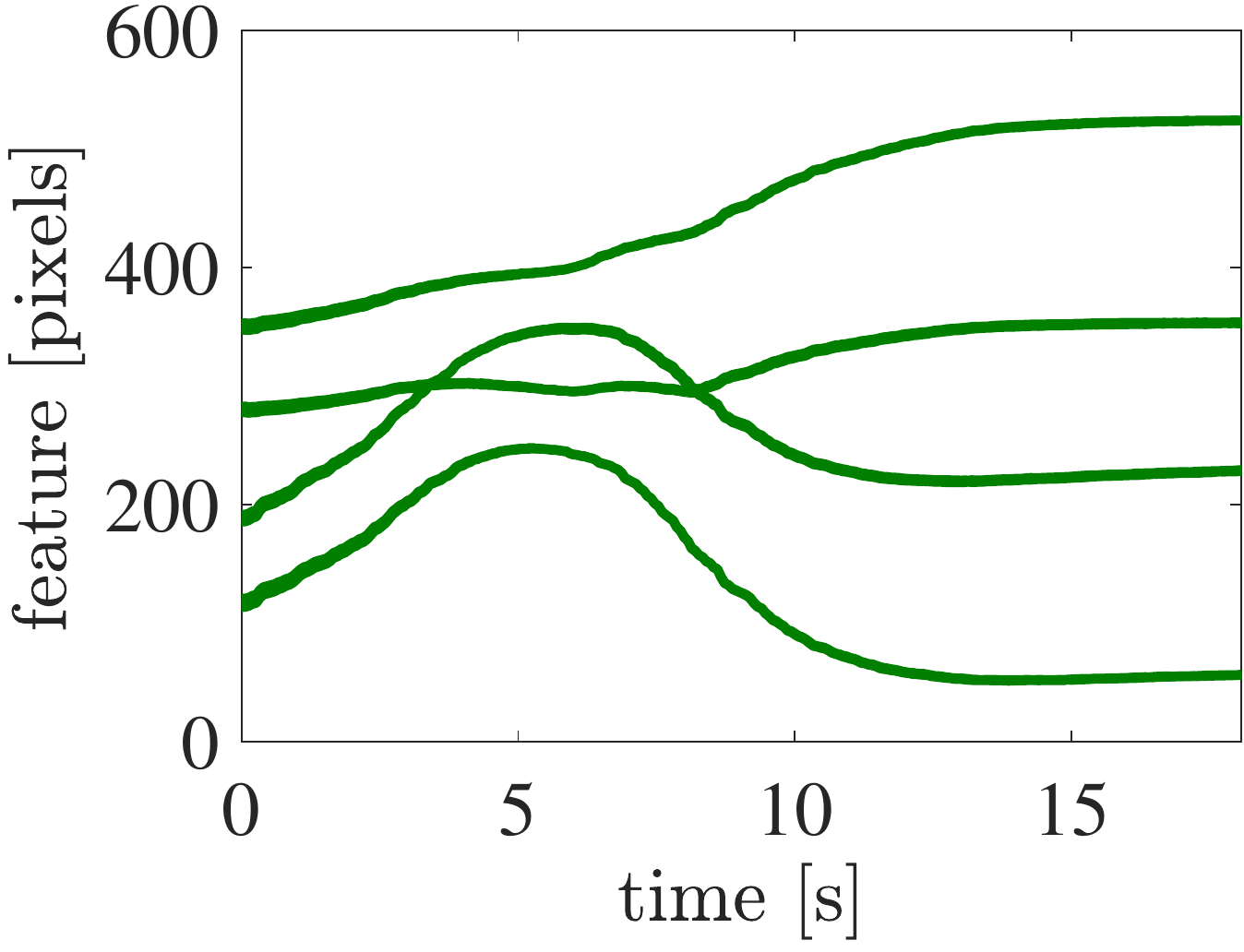}%
    \caption{Peg-in-hole task with shifted target: camera position (left) and features trajectories (right) performed by \ac{rds}~(top) and \ac{clfdm}~(bottom).}%
    \label{fig:peg_in_hole_new_goal}%
\end{figure}

To further show the benefits of using a 
visual feedback into \ac{il} techniques, we repeated the peg-in-hole task by moving the target to a novel position (Fig~\ref{fig:peg_in_hole_new_goal}). 
Also in this case, the robot 
fulfilled the \ac{vs} task 
avoiding collisions, demonstrating enhanced robustness and generalization capabilities.

The robot performing the peg-in-hole task under different conditions is shown in the accompanying video.

\section{Conclusions and Future Work}\label{sec:conclusion}

We presented the application of three DS-based IL methods to the VS case, for the achievement of complex visual tasks without the need of explicit coding. 
By leveraging the information of few demonstrated trajectories, it has been possible to augment the basic VS behavior with additional tasks, such as collision avoidance.
At the same time, using VS within DS techniques has allowed to adapt to environmental changes. 
Simulations and experiments validated the approach, showing the beneficial integration of DS and VS.

Future work will be devoted to reduce even more the user's programming burden. 
For example, direct VS approaches, which avoid to explicitly extract visual features from images, could be integrated in the DS-based paradigm.
Furthermore, the proposed system could be used to ease the deployment of VS strategies on complex platforms, such as humanoids.


\section*{Acknowledgment}
Authors would like to thank N. Perrin and P. Schlehuber-Caissier for sharing the code of the FDM method, and A. Giusti for supporting and proofreading the paper.


\bibliographystyle{IEEEtran}
\bibliography{bibliography.bib}

\begin{thebibliography}{10}
\providecommand{\url}[1]{#1}
\csname url@rmstyle\endcsname
\providecommand{\newblock}{\relax}
\providecommand{\bibinfo}[2]{#2}
\providecommand\BIBentrySTDinterwordspacing{\spaceskip=0pt\relax}
\providecommand\BIBentryALTinterwordstretchfactor{4}
\providecommand\BIBentryALTinterwordspacing{\spaceskip=\fontdimen2\font plus
\BIBentryALTinterwordstretchfactor\fontdimen3\font minus
  \fontdimen4\font\relax}
\providecommand\BIBforeignlanguage[2]{{%
\expandafter\ifx\csname l@#1\endcsname\relax
\typeout{** WARNING: IEEEtran.bst: No hyphenation pattern has been}%
\typeout{** loaded for the language `#1'. Using the pattern for}%
\typeout{** the default language instead.}%
\else
\language=\csname l@#1\endcsname
\fi
#2}}

\bibitem{lee2011incremental}
D.~Lee and C.~Ott, ``Incremental kinesthetic teaching of motion primitives
  using the refinement tub,'' \emph{Autonomous Robots}, vol.~31, no.~2, pp.
  115--131, 2011.

\bibitem{caccavale2019kinesthetic}
R.~Caccavale, M.~Saveriano, A.~Finzi, and D.~Lee, ``Kinesthetic teaching and
  attentional supervision of structured tasks in human--robot interaction,''
  \emph{Autonomous Robots}, vol.~43, no.~6, pp. 1291--1307, 2019.

\bibitem{lee2020gesture}
D.~Lee, ``Gesture, posture, facial interfaces,'' in \emph{Encyclopedia of
  Robotics}, M.~H. Ang, O.~Khatib, and B.~Siciliano, Eds.\hskip 1em plus 0.5em
  minus 0.4em\relax Berlin, Heidelberg: Springer Berlin Heidelberg, 2020, pp.
  1--10.

\bibitem{billard16learning}
A.~Billard, S.~Calinon, and R.~Dillmann, ``Learning from humans,'' in
  \emph{Handbook of Robotics}, B.~Siciliano and O.~Khatib, Eds.\hskip 1em plus
  0.5em minus 0.4em\relax Secaucus, NJ, USA: Springer, 2016, ch.~74, pp.
  1995--2014, 2nd Edition.

\bibitem{Argall2009survey}
B.~D. Argall, S.~Chernova, M.~Veloso, and B.~Browning, ``A survey of robot
  learning from demonstration,'' \emph{Robotics and autonomous systems},
  vol.~57, no.~5, pp. 469--483, 2009.

\bibitem{DMP}
A.~Ijspeert, J.~Nakanishi, P.~Pastor, H.~Hoffmann, and S.~Schaal, ``{D}ynamical
  {M}ovement {P}rimitives: learning attractor models for motor behaviors,''
  \emph{{N}eural {C}omputation}, vol.~25, no.~2, pp. 328--373, 2013.

\bibitem{saveriano2018incremental}
M.~Saveriano and D.~Lee, ``Incremental skill learning of stable dynamical
  systems,'' in \emph{{IEEE/RSJ} Int. Conf. on Intelligent Robots and Systems},
  2018, pp. 6574--6581.

\bibitem{saveriano2020energy}
M.~Saveriano, ``An energy-based approach to ensure the stability of learned
  dynamical systems,'' in \emph{{IEEE} Int. Conf. on Robotics and Automation},
  2020, pp. 4407--4413.

\bibitem{SEDS}
S.~M. Khansari-Zadeh and A.~Billard, ``Learning stable non-linear dynamical
  systems with gaussian mixture models,'' \emph{IEEE Trans. Robot.}, vol.~27,
  no.~5, pp. 943--957, 2011.

\bibitem{Clf}
------, ``Learning control {L}yapunov function to ensure stability of dynamical
  system-based robot reaching motions,'' \emph{Robot. Auton. Syst.}, vol.~62,
  no.~6, pp. 752--765, 2014.

\bibitem{Perrin16}
N.~Perrin and P.~Schlehuber-Caissier, ``Fast diffeomorphic matching to learn
  globally asymptotically stable nonlinear dynamical systems,'' \emph{Systems
  \& Control Letters}, vol.~96, pp. 51--59, 2016.

\bibitem{urain2020imitationflow}
J.~Urain, M.~Ginesi, D.~Tateo, and J.~Peters, ``Imitationflow: Learning deep
  stable stochastic dynamic systems by normalizing flows,'' in \emph{{IEEE/RSJ}
  Int. Conf. on Intelligent Robots and Systems}, 2020, pp. 5231--5237.

\bibitem{Franken11}
M.~Franken, S.~Stramigioli, S.~Misra, C.~Secchi, and A.~Macchelli, ``Bilateral
  telemanipulation with time delays: A two-layer approach combining passivity
  and transparency,'' \emph{Transactions on Robotics}, vol.~27, no.~4, pp.
  741--756, 2011.

\bibitem{Chaumette:ram:2006}
F.~{Chaumette} and S.~{Hutchinson}, ``{Visual servo control. Part I: Basic
  approaches},'' \emph{IEEE Robot. Autom. Mag.}, vol.~13, no.~4, pp. 82--90,
  2006.

\bibitem{Chaumette:ram:2007}
F.~Chaumette and S.~Hutchinson, ``{Visual servo control. Part II: Advanced
  approaches},'' \emph{IEEE Robot. Autom. Mag.}, vol.~14, no.~1, pp. 109--118,
  2007.

\bibitem{Chesi_tro:2007}
G.~{Chesi} and Y.~S. {Hung}, ``Global path-planning for constrained and optimal
  visual servoing,'' \emph{IEEE Trans. Robot.}, vol.~23, no.~5, pp. 1050--1060,
  2007.

\bibitem{Mezouar:tro:2002}
Y.~{Mezouar} and F.~{Chaumette}, ``Path planning for robust image-based
  control,'' \emph{IEEE Trans. Robot.}, vol.~18, no.~4, pp. 534--549, 2002.

\bibitem{Sauvee:cdc:2006}
M.~{Sauvee}, P.~{Poignet}, E.~{Dombre}, and E.~{Courtial}, ``Image based visual
  servoing through nonlinear model predictive control,'' in \emph{IEEE
  Conference on Decision and Control}, 2006, pp. 1776--1781.

\bibitem{Allibert:tro:2010}
G.~Allibert, E.~Courtial, and F.~Chaumette, ``Predictive control for
  constrained image-based visual servoing,'' \emph{IEEE Trans. Robot.},
  vol.~26, no.~5, pp. 933--939, 2010.

\bibitem{Paolillo:icra:2020}
A.~{Paolillo}, T.~S. {Lembono}, and S.~{Calinon}, ``A memory of motion for
  visual predictive control tasks,'' in \emph{{IEEE} Int. Conf. on Robotics and
  Automation}, 2020, pp. 9014--9020.

\bibitem{Agravante:ral:2017}
D.~J. {Agravante}, G.~{Claudio}, F.~{Spindler}, and F.~{Chaumette}, ``Visual
  servoing in an optimization framework for the whole-body control of humanoid
  robots,'' \emph{IEEE Robot. and Autom. Lett.}, vol.~2, no.~2, pp. 608--615,
  2017.

\bibitem{Paolillo:ral:2018}
A.~{Paolillo}, K.~{Chappellet}, A.~{Bolotnikova}, and A.~{Kheddar},
  ``Interlinked visual tracking and robotic manipulation of articulated
  objects,'' \emph{IEEE Robot. and Autom. Lett.}, vol.~3, no.~4, pp.
  2746--2753, 2018.

\bibitem{Mingo:icra:2021}
E.~Mingo~Hoffman and A.~Paolillo, ``Exploiting visual servoing and centroidal
  momentum for whole-body motion control of humanoid robots in absence of
  contacts and gravity,'' in \emph{{IEEE} Int. Conf. on Robotics and
  Automation}, 2021, pp. 2979--2985.

\bibitem{pignat2019bayesian}
E.~Pignat and S.~Calinon, ``{B}ayesian {G}aussian mixture model for robotic
  policy imitation,'' \emph{IEEE Robot. and Autom. Lett.}, vol.~4, no.~4, pp.
  4452--4458, 2019.

\bibitem{castelli2017machine}
F.~Castelli, S.~Michieletto, S.~Ghidoni, and E.~Pagello, ``A machine
  learning-based visual servoing approach for fast robot control in industrial
  setting,'' \emph{International Journal of Advanced Robotic Systems}, vol.~14,
  no.~6, 2017.

\bibitem{jin2019robot}
J.~{Jin}, L.~{Petrich}, M.~{Dehghan}, Z.~{Zhang}, and M.~{Jagersand}, ``Robot
  eye-hand coordination learning by watching human demonstrations: a task
  function approximation approach,'' in \emph{{IEEE} Int. Conf. on Robotics and
  Automation}, 2019, pp. 6624--6630.

\bibitem{Jin:iros:2020}
J.~Jin, L.~Petrich, M.~Dehghan, and M.~Jagersand, ``A geometric perspective on
  visual imitation learning,'' in \emph{{IEEE/RSJ} Int. Conf. on Intelligent
  Robots and Systems}, 2020, pp. 5194--5200.

\bibitem{dometios2018vision}
A.~C. Dometios, Y.~Zhou, X.~S. Papageorgiou, C.~S. Tzafestas, and T.~Asfour,
  ``Vision-based online adaptation of motion primitives to dynamic surfaces:
  application to an interactive robotic wiping task,'' \emph{IEEE Robot. and
  Autom. Lett.}, vol.~3, no.~3, pp. 1410--1417, 2018.

\bibitem{kim:icra:2020}
J.~W. Kim, C.~He, M.~Urias, P.~Gehlbach, G.~D. Hager, I.~Iordachita, and
  M.~Kobilarov, ``Autonomously navigating a surgical tool inside the eye by
  learning from demonstration,'' in \emph{{IEEE} Int. Conf. on Robotics and
  Automation}, 2020, pp. 7351--7357.

\bibitem{Paradis:icra:2021}
S.~Paradis, M.~Hwang, B.~Thananjeyan, J.~Ichnowski, D.~Seita, D.~Fer, T.~Low,
  J.~E. Gonzalez, and K.~Goldberg, ``Intermittent visual servoing: Efficiently
  learning policies robust to instrument changes for high-precision surgical
  manipulation,'' in \emph{{IEEE} Int. Conf. on Robotics and Automation}, 2021.

\bibitem{Jia:ral:2019}
B.~Jia, Z.~Pan, Z.~Hu, J.~Pan, and D.~Manocha, ``Cloth manipulation using
  random-forest-based imitation learning,'' \emph{IEEE Robot. and Autom.
  Lett.}, vol.~4, no.~2, pp. 2086--2093, 2019.

\bibitem{Johns:icra:2021}
E.~Johns, ``Coarse-to-fine imitation learning: Robot manipulation from a single
  demonstration,'' in \emph{{IEEE} Int. Conf. on Robotics and Automation},
  2021.

\bibitem{phung2012tool}
A.~Phung, J.~Malzahn, F.~Hoffmann, and T.~Bertram, ``Tool centered learning
  from demonstration for robotic arms with visual feedback,'' in \emph{IEEE
  Int. Conf. on Robotics and Biomimetics}, 2012, pp. 1117--1122.

\bibitem{vakanski2017imagebased}
A.~Vakanski, F.~Janabi-Sharifi, and I.~Mantegh, ``An image-based trajectory
  planning approach for robust robot programming by demonstration,''
  \emph{Robot. Auton. Syst.}, vol.~98, pp. 241--257, 2017.

\bibitem{Slotine91}
J.~Slotine and W.~Li, \emph{Applied nonlinear control}.\hskip 1em plus 0.5em
  minus 0.4em\relax Prentice-Hall Englewood Cliffs, 1991.

\bibitem{tau-SEDS}
K.~Neumann and J.~J. Steil, ``Learning robot motions with stable dynamical
  systems under diffeomorphic transformations,'' \emph{Robot. Auton. Syst.},
  vol.~70, pp. 1--15, 2015.

\bibitem{shavit2018learning}
Y.~Shavit, N.~Figueroa, S.~S.~M. Salehian, and A.~Billard, ``Learning augmented
  joint-space task-oriented dynamical systems: a linear parameter varying and
  synergetic control approach,'' \emph{IEEE Robotics and Automation Letters},
  vol.~3, no.~3, pp. 2718--2725, 2018.

\bibitem{GMR}
D.~A. Cohn, Z.~Ghahramani, and M.~I. Jordan, ``Active learning with statistical
  models,'' \emph{Journal of Artificial Intelligence Research}, vol.~4, no.~1,
  pp. 129--145, 1996.

\bibitem{LASA_Dataset}
\BIBentryALTinterwordspacing
S.~M. Khansari-Zadeh, ``The {LASA} handwriting dataset.'' [Online]. Available:
  \url{http://bitbucket.org/khansari/lasahandwritingdataset}
\BIBentrySTDinterwordspacing

\bibitem{corke:book:2011}
P.~Corke, \emph{Robotics, Vision \& Control}.\hskip 1em plus 0.5em minus
  0.4em\relax Springer, 2011.

\bibitem{Marchand:ram:2005}
E.~Marchand, F.~Spindler, and F.~Chaumette, ``{ViSP} for visual servoing: a
  generic software platform with a wide class of robot control skills,''
  \emph{IEEE Robot. Autom. Mag.}, vol.~12, no.~4, pp. 40--52, 2005.

\end{thebibliography}

\end{document}